\documentclass[10pt,twocolumn,letterpaper]{article}

\usepackage[pagenumbers]{cvpr} 

\definecolor{customblue}{rgb}{0.2, 0.3, 0.8}
\definecolor{customgreen}{rgb}{0.1, 0.6, 0.3}

\definecolor{lightblue}{RGB}{173, 216, 230}
\definecolor{lightgray}{gray}{0.9}
\definecolor{lightgreen}{RGB}{144, 238, 144}
\definecolor{lightred}{RGB}{255, 182, 193}

\definecolor{wkred}{RGB}{255, 190, 190}
\definecolor{wkblue}{RGB}{210, 230, 250}
\definecolor{wkgreen}{RGB}{226,240,217}

\definecolor{skyblue}{RGB}{0, 102, 204}

\usepackage{listings}
\usepackage{multirow}
\usepackage{tabularx}
\usepackage{booktabs}

\usepackage{float}

\newcommand{\second}[1]{\cellcolor{wkblue}\underline{#1}}
\newcommand{\best}[1]{\cellcolor{wkgreen}\textbf{#1}}

\usepackage{etoolbox, subcaption}
\usepackage[font=footnotesize,labelfont=bf,aboveskip=1pt,belowskip=-10pt]{caption}

\newdimen\abovecrulesep
\newdimen\belowcrulesep
\makeatletter
\patchcmd{\@@@cmidrule}{\aboverulesep}{\abovecrulesep}{}{}
\patchcmd{\@xcmidrule}{\belowrulesep}{\belowcrulesep}{}{}

\definecolor{demphcolor}{RGB}{144, 144, 144}
\definecolor{mygray}{gray}{0.4}
\definecolor{lightgray}{rgb}{0.9, 0.9, 0.9}
\newcommand{\demph}[1]{\textcolor{demphcolor}{#1}}

\newlength\savewidth
\newcommand\shline{\noalign{\global\savewidth\arrayrulewidth\global\arrayrulewidth 1pt}\hline\noalign{\global\arrayrulewidth\savewidth}}

\newcommand{\tablestyle}[2]{\setlength{\tabcolsep}{#1}\renewcommand{\arraystretch}{#2}\centering\footnotesize}
\makeatletter\renewcommand\paragraph{\@startsection{paragraph}{4}{\z@}{.5em\@plus1ex\@minus.2ex}{-.5em}{\normalfont\normalsize\bfseries}}
\makeatother

\newcolumntype{C}[1]{>{\centering\arraybackslash}p{#1}}
\newcolumntype{R}[1]{>{\raggedleft\arraybackslash}p{#1}}
\newcolumntype{L}[1]{>{\raggedright\arraybackslash}p{#1}}

\newcommand{\cmark}{\text{\ding{51}}} 
\newcommand{\xmark}{\text{\ding{55}}}

\preto\align{\small}
\expandafter\preto\csname align*\endcsname{\small}
\preto\equation{\par\nobreak\small\noindent}

\usepackage{graphicx}
\graphicspath{{assets/}} 
\usepackage{amsmath}
\usepackage{amssymb}
\usepackage{booktabs}
\usepackage{enumitem}
\usepackage{algorithm}
\usepackage{algpseudocode}
\usepackage{bm}
\usepackage{multirow}
\usepackage{caption}
\usepackage{subcaption}
\usepackage{tikz}
\usepackage{tabularx}

\usepackage{arydshln}

\usepackage{adjustbox}
\usepackage{natbib}

\usepackage{pifont}

\usepackage{lipsum}
\usepackage{bbm}
\usepackage{dsfont}


\definecolor{cvprblue}{rgb}{0.21,0.49,0.74}
\usepackage[pagebackref,breaklinks,colorlinks,citecolor=cvprblue]{hyperref}

\usepackage[capitalize]{cleveref}
\crefname{section}{Sec.}{Secs.}
\Crefname{section}{Section}{Sections}
\Crefname{table}{Table}{Tables}
\crefname{table}{Tab.}{Tabs.}

\def\paperID{10455} 
\def\confName{CVPR}
\def\confYear{2025}

\begin{document}


\title{ 
Can MLLMs Reason in Multimodality? \\
\raisebox{-0.1cm}{\includegraphics[width=0.13\textwidth]{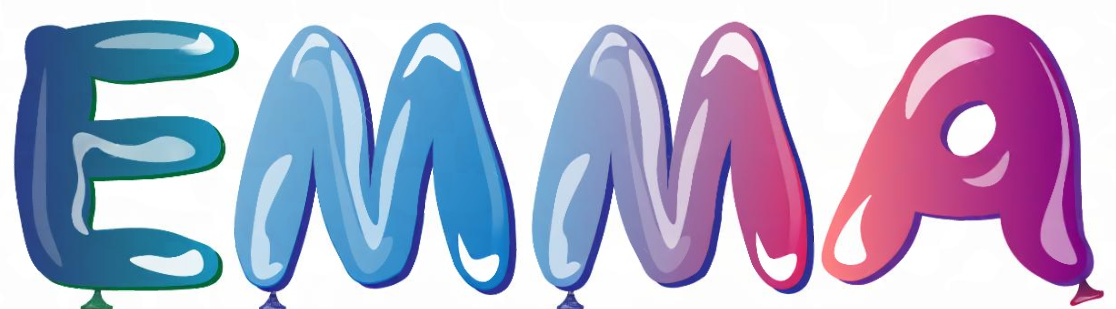}}: An Enhanced MultiModal ReAsoning Benchmark
}

\author{
Yunzhuo Hao$^{1}$*, Jiawei Gu$^{2}$*, Huichen Will Wang$^{3}$*, Linjie Li$^{4}$*, \\
Zhengyuan Yang$^{4}$, Lijuan Wang$^{4}$, Yu Cheng$^{5}$\\
\normalsize $^{1}$University of Electronic Science and Technology of China, $^{2}$Sun Yat-sen University,\\
\normalsize $^{3}$University of Washington, $^{4}$Microsoft, $^{5}$The Chinese University of Hong Kong \\
\href{https://emma-benchmark.github.io}{emma-benchmark.github.io}
}


\twocolumn[{
\maketitle

\begin{center}
    \vskip -0.3in
    \captionsetup{type=figure}
    \includegraphics[width=1.\linewidth]{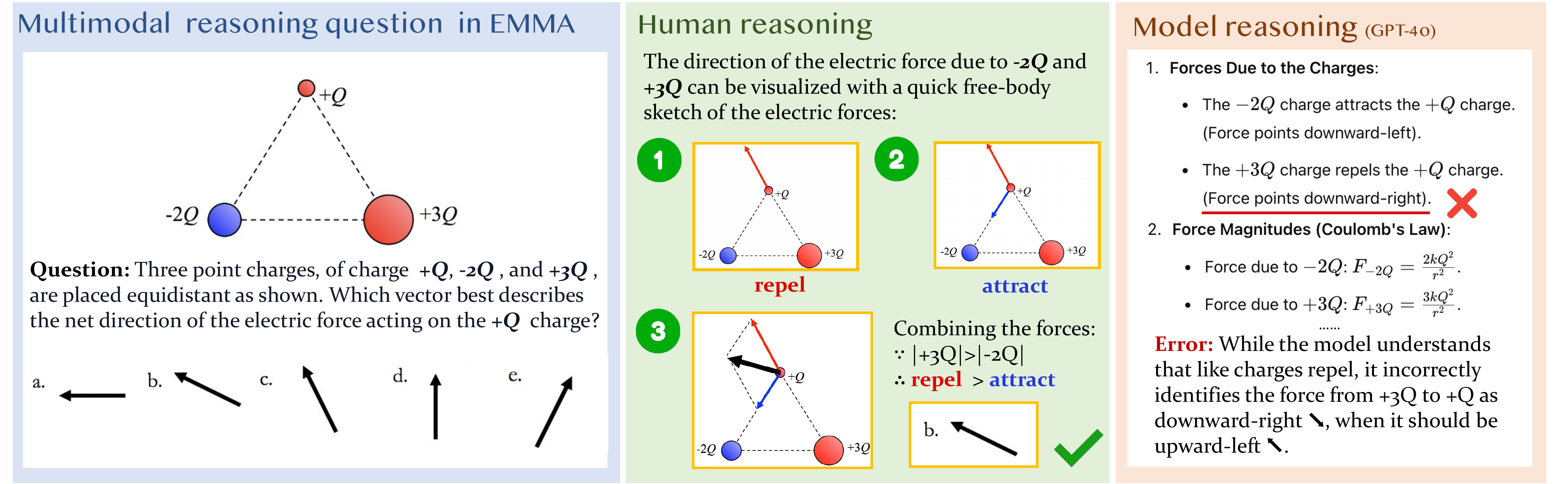}
    \caption{\textbf{A sample multimodal reasoning question in EMMA.} To solve it, humans engage in graphical reasoning (middle panel): guided by the principles of electric force, they draw force vectors with appropriate directions and visually compute their sum. While GPT-4o understands that like charges repel, it mistakes the direction of the repulsive force (right panel), highlighting its limitations in multimodal reasoning.}
    
    \vspace{8pt}
     \label{fig:overview}
\end{center}
}
]


\begin{abstract}
\vskip -0.2in

The ability to organically reason \textbf{over} and \textbf{with} both text and images is a pillar of human intelligence, yet the ability of Multimodal Large Language Models (MLLMs) to perform such multimodal reasoning remains under-explored.
Existing benchmarks often emphasize text-dominant reasoning or rely on shallow visual cues, failing to adequately assess integrated visual and textual reasoning. We introduce EMMA (Enhanced MultiModal reAsoning), a benchmark targeting organic multimodal reasoning across mathematics, physics, chemistry, and coding. 
EMMA tasks demand advanced cross-modal reasoning that cannot be addressed by reasoning independently in each modality, offering an enhanced test suite for MLLMs' reasoning capabilities. 
Our evaluation of state-of-the-art MLLMs on EMMA reveals significant limitations in handling complex multimodal and multi-step reasoning tasks, even with advanced techniques like Chain-of-Thought prompting and test-time compute scaling underperforming. These findings underscore the need for improved multimodal architectures and training paradigms to close the gap between human and model reasoning in multimodality. 

\end{abstract}
\vskip -0.3in

\def\thefootnote{*}\footnotetext{Equal contribution.}\def\thefootnote{\arabic{footnote}}

\section{Introduction}

\begin{figure*}[t]
    \centering 
    \includegraphics[width=\textwidth]{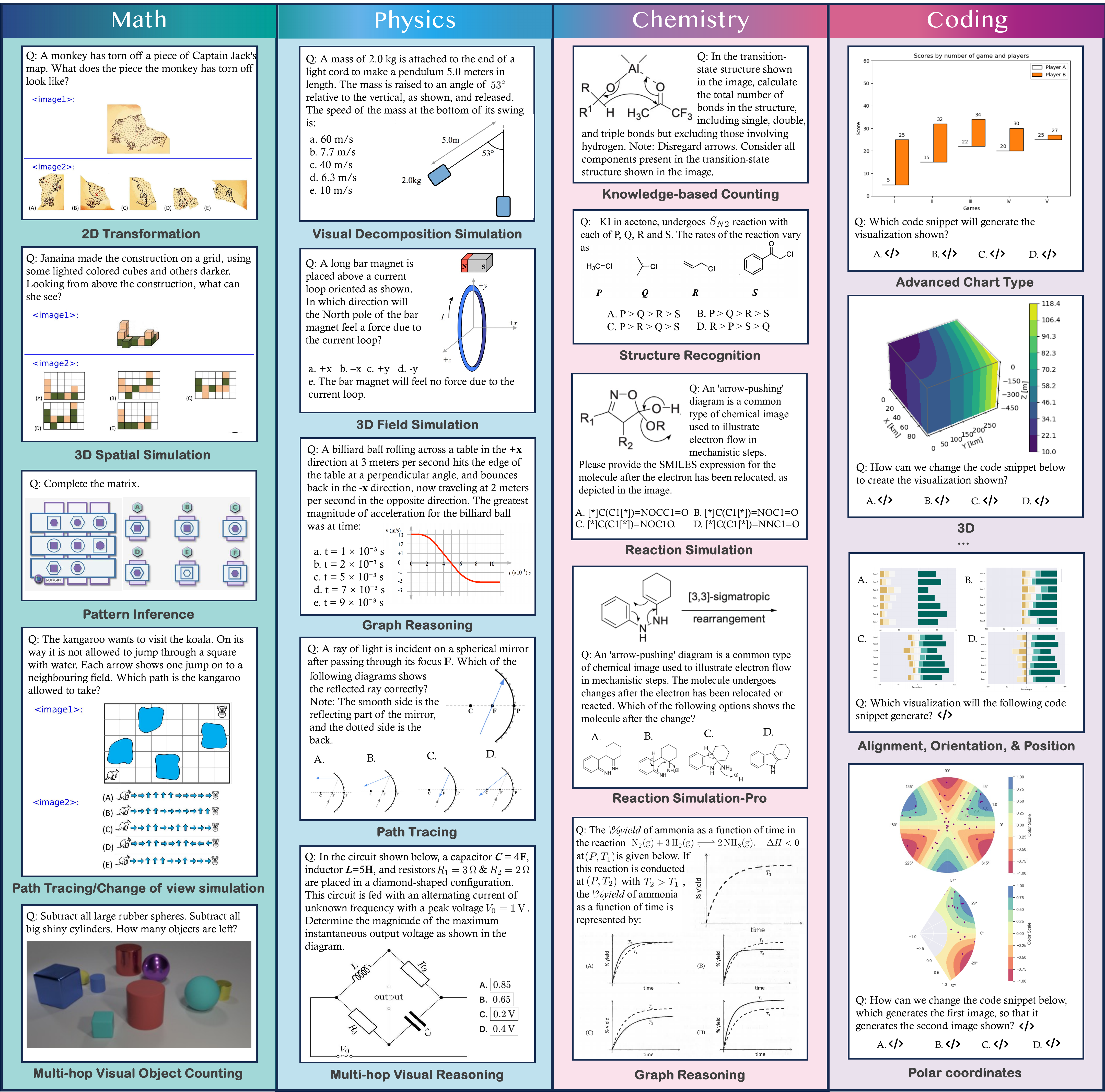}
    \caption{\textbf{Overview of EMMA.} EMMA covers four subjects: math, physics, chemistry, and coding. Questions are categorized based on specific skills. We provide answers to the sample questions in this figure on the GitHub page. }
    \label{fig:categories}
\end{figure*}

\vskip -0.1in
Multimodal reasoning is fundamental to human problem-solving. For example, meteorologists interpret weather maps alongside numerical forecasts, and interior designers combine textual descriptions with mental imagery to optimize room layouts. Text-based reasoning allows us to analyze abstract concepts, while visual reasoning enables us to manipulate and draw insights from complex visual information.
Combining these skills provides a robust framework for solving technical and creative challenges. In math, for example, formal symbolic reasoning and visual aids such as diagrams and spatial thinking work in tandem to explore ideas that neither modality could fully address alone~\cite{Giaquinto_2020}.

Recent advancements in Large Language Models (LLMs) have significantly enhanced their reasoning abilities~\cite{o1, qwq, zhao2024marco, r1}, enabling strong performance on tasks such as formal logic reasoning~\cite{mmlu}, graduate-level academic question answering~\cite{gpqa}, and competitive programming~\cite{codeforces, bai2023qwen, hendrycks2020measuring}. 
Despite these successes, these models primarily focus on \textit{text-only reasoning}, leaving an open question: can Multimodal LLMs (MLLMs) effectively reason across both language and visual inputs?

A major bottleneck in addressing this question is the lack of appropriate benchmarks. Existing multimodal benchmarks largely test surface-level visual understanding~\cite{antol2015vqa, goyal2017making, hudson2019gqa, yu2023mm, akter2024visreas} or textual knowledge recall with multimodal inputs~\cite{yue2024mmmu,yue2023mmmu,okvqa}. While some benchmarks~\cite{lu2024mathvista,wang2024measuring} include math questions with images, studies~\cite{zhang2024mathverse} have shown that many of these tasks reduce to language-only reasoning, as the visual content is often fully described in text.

To address this gap, we introduce EMMA: an Enhanced MultiModal reAsoning benchmark, specifically designed to evaluate the ability to solve problems that require both visual- and language-based problem-solving. EMMA features questions that \textit{are difficult to solve by relying solely on text-based reasoning or a single visual pass}. Instead, solving these problems necessitates a back-and-forth process between interpreting visual inputs and applying multimodal reasoning steps, where visual aids are often integral or more efficient for arriving at the solution. For instance, Figure~\ref{fig:overview} illustrates a sample physics problem that asks for the direction of the net electric force. While GPT-4o understands that like charges repel, it
mistakes the direction of the repulsive force, highlighting its limitations in multimodal reasoning.

Unlike recent benchmarks~\cite{spatialcog, chollet2019measure}, which focus on spatial cognition, or visual puzzles that can be perfectly represented in text, EMMA introduces domain-specific challenges where reasoning can often be strengthened by visual aids. These include tasks like 3D spatial transformations, chemical structure recognition, multi-step physical simulations, and program output visualization (Figure~\ref{fig:categories}). EMMA consists of 992 multimodal reasoning questions gathered from existing benchmarks through a rigorous filtering pipeline, and 1,796 newly constructed questions created manually in collaboration with domain experts. Our evaluation of nine state-of-the-art (SoTA) MLLMs on EMMA reveals three key findings:

\begin{itemize}
    \item \textbf{MLLMs struggle with multimodal reasoning}: All models perform suboptimally on EMMA, regardless of the usage of Chain-of-Thought (CoT) prompting~\cite{wei2022chain}. On the balanced subset of EMMA, the best-performing model, o1, scores only 45.75\%, which is 8.5\% higher than the best non-reasoning MLLM, Qwen2-VL~\cite{wang2024qwen2}, but still trails human experts by 32\%. These results suggest a limitation of current MLLMs to perform in-depth multimodal reasoning. 
    \item \textbf{Test-time compute scaling methods with textual CoTs are insufficient}: 
    We explore test-time compute scaling of SoTA MLLMs with 
    different methods
    (e.g., majority voting, best-of-N, and tournament) up to 16 times, yet they still fail to address the challenges posed by multimodal reasoning questions in EMMA.
    Our analysis reveals that simply increasing the number of candidate responses with textual CoTs does little to compensate for the models' inability to produce valid visual reasoning steps, particularly for tasks requiring fine-grained spatial understanding or multi-step reasoning. In addition, current MLLMs and specialized reward models struggle with complex multimodal reasoning themselves, which can make their reward signals unreliable and limit the utility of test-time compute scaling. 
    \item \textbf{Visual reasoning remains a fundamental bottleneck}: Through thorough error analysis, we find that SoTA MLLMs frequently struggle with tasks requiring precise spatial simulations, multi-hop visual reasoning, and integration of visual and textual information. These shortcomings are particularly pronounced in problems where visual aids offer a simpler or more natural path to the solution. Further, textual CoT negatively impacts model performance on visual-reasoning-heavy tasks, highlighting the need for new paradigms to improve visual reasoning.
\end{itemize}

These insights suggest that the performance gap between text-based and multimodal reasoning arises from MLLMs' limited ability to perform fine-grained visual reasoning. EMMA highlights the need for new architectures and training paradigms that can better integrate and reason over diverse modalities, enabling models to leverage both visual and linguistic information more effectively.

\section{Related Work}
\label{sec:mmbench}

\paragraph{Multimodal Large Language Models}  
Recent years have witnessed rapid progress in MLLM development. Building upon early techniques in vision-language modeling~\cite{tan2019lxmert, lu2019vilbert, chen2020uniter, radford2021learning, li2020oscar, zhang2021vinvl, yu2022coca}, modern MLLMs~\cite{li2024llava, lu2024deepseek, internVL, liu2024improved, qwen2, achiam2023gpt, team5gemini, li2023blip} leverage the success of LLMs and achieve impressive performance in many multimodal tasks. In addition, various visual instruction tuning techniques~\cite{liu2024visual, liu2024improved, zhu2023minigpt} and the increasing availability of open-source and model-generated training data have further contributed to the robustness and zero-shot generalization ability of MLLMs.

\paragraph{LLM and MLLM Reasoning}  
While early LLMs were widely regarded as mere next-token predictors with limited reasoning ability~\cite{huang2023large, huang2022towards}, recent advancements in LLM research~\cite{o1, qwq, zhao2024marco, r1} have begun to challenge this view. State-of-the-art models now achieve strong performance on tasks such as formal logic reasoning~\cite{mmlu}, graduate-level academic question answering~\cite{gpqa}, and competitive programming~\cite{codeforces}. These advancements in text-based reasoning have spurred growing interest in multimodal reasoning, exemplified by visual CoT models such as Visual Chain-of-Thought~\cite{shao2024visual} and Multimodal Chain-of-Thought~\cite{zhang2023multimodal}, as well as visual CoT prompting techniques like Image-of-Thought~\cite{zhou2024image}. 
Although visual CoT prompting techniques have shown promise, their focus is primarily on enhancing perception through methods like cropping images to simulate attention. Hence, these approaches offer limited support for tasks that demand more advanced visual reasoning skills, such as visual manipulation or imagination.




\paragraph{Multimodal Reasoning Benchmarks} 
Most reasoning benchmarks to date are purely text-based (e.g., ~\cite{cobbe2021training, MATH, bigbench, cladder, suzgun2022challenging}). The growing demand for measuring multimodal reasoning has driven the development of multimodal reasoning benchmarks across diverse domains (e.g., ~\cite{lu2024mathvista, wang2024measuring, li2024mmcode, yang2024swebenchmultimodalaisystems, ying2024mmt, chen2024viseval, mmsci, cheng2024comt}). In particular, recent efforts have targeted spatial and relational reasoning~\cite{akter2024visreas, spatialcog} and college-level reasoning that requires domain knowledge~\cite{yue2023mmmu}. Nonetheless, a persistent shortcoming remains: many purported multimodal benchmarks contain redundancy between text and images, allowing models to shortcut through language reasoning alone. For example, Zhang et al.~\cite{zhang2024mathverse} observe that textual elements in math benchmarks often describe the visual content in detail, reducing the need for integrated reasoning across modalities. In light of this, MMMU-Pro~\cite{yue2024mmmu} incorporates a filtering pipeline to better evaluate true multimodal reasoning. In this work, we further refine such approaches by curating a benchmark that focuses explicitly on tasks requiring strong visual reasoning. Unlike existing benchmarks, our test suite emphasizes multimodal reasoning challenges that are difficult to solve with text-based reasoning and a single visual pass.

\section{The EMMA Benchmark}

\subsection{Overview of EMMA}
\begin{figure}[t]
    \centering 
    \includegraphics[width=\columnwidth]{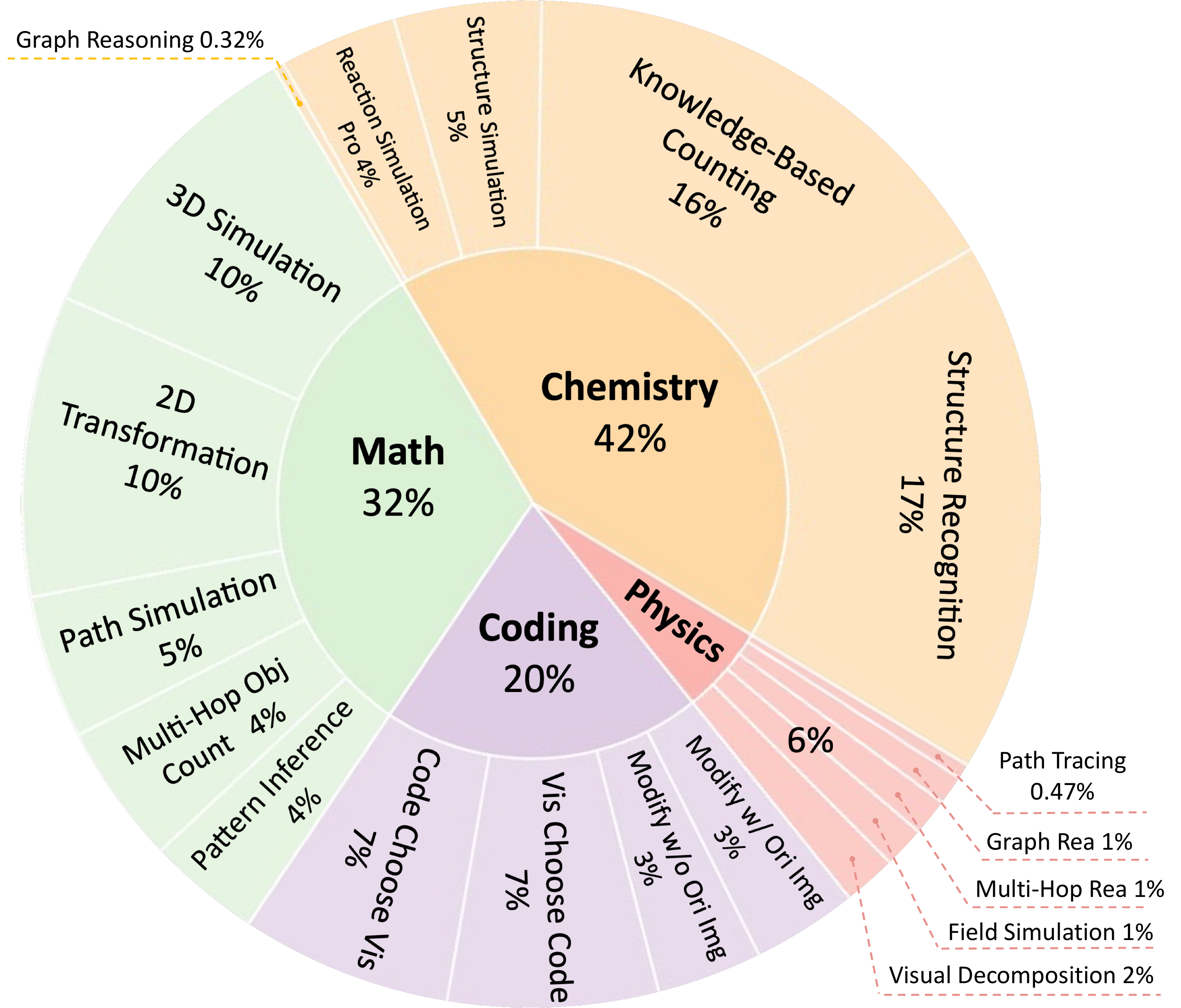}
    \caption{\textbf{Composition of EMMA.} EMMA comprises 2,788 questions across four subjects: math, physics, chemistry, and coding. Within each subject, we further provide fine-grained labels for each question based on the specific skills it measures. 
    }
    \label{fig:composition}
\end{figure}

We introduce EMMA, an Enhanced MultiModal ReAsoning Benchmark. 
EMMA is composed of 2,788 problems, of which 1,796 are newly constructed, across four domains: math, physics, chemistry, and coding. The key statistics of EMMA are summarized in Table~\ref{tab:keystats}, and its composition is presented in Figure~\ref{fig:composition}. 

To provide fine-grained insights into how MLLMs might fail in multimodal reasoning, we assign labels to each problem in our benchmark. These labels are either created by domain experts or assigned by GPT-4o and subsequently verified by experts. 
As shown in Figure~\ref{fig:categories}, questions in EMMA assess a wide array of multimodal reasoning skills. For example, the pattern inference problem in math challenges models to identify and generalize visual patterns; the visual decomposition simulation problem in physics requires graphically decomposing forces to determine resultant effects; the reaction simulation problem in chemistry demands precise interpretation and simulation of electron movement; the 3D visualization problem in coding\footnote{Different from the other subjects in EMMA, coding questions can be assigned more than one category since our visualizations tend to employ multiple advanced techniques.} evaluates spatial imagination by requiring models to associate function calls with their corresponding 3D representations. 

\begin{table}[t]
\tablestyle{15pt}{1.1}
    \centering
    \small
    \resizebox{.95\columnwidth}{!}
{
    \begin{tabular}{l r}
    \shline
    \textbf{Statistic} & \textbf{Number} \\
    \hline
    Total questions &  2,788 \\
    - Multiple-choice questions & 2,002 (72\%) \\
    - Free-form questions & 786 (28\%) \\
    - Questions with answers & 2,788 (100\%) \\
    - Questions newly added & 1,796 (64\%) \\
    \hline
    Image in the question & 2,599 (93\%) \\
    Image in the option(s) & 195 (7\%) \\
    Problems with multiple images & 298 (10\%) \\
    \shline
    \end{tabular}
    }
    \caption{\textbf{Key statistics of EMMA.}}
    \label{tab:keystats}
\end{table}

In addition, to ease the burden of evaluation, all questions in EMMA are in either multiple-choice or open-ended formats with short, easily checkable ground truth answers, obviating the need for using MLLMs as judges~\cite{zhang2024gpt, shi2024chartmimic, wu2024plot2code}. 

\subsection{Data Curation}
\label{subsec: Data Curation}

\begin{figure}[t]
    \center  
    \includegraphics[width=\columnwidth]{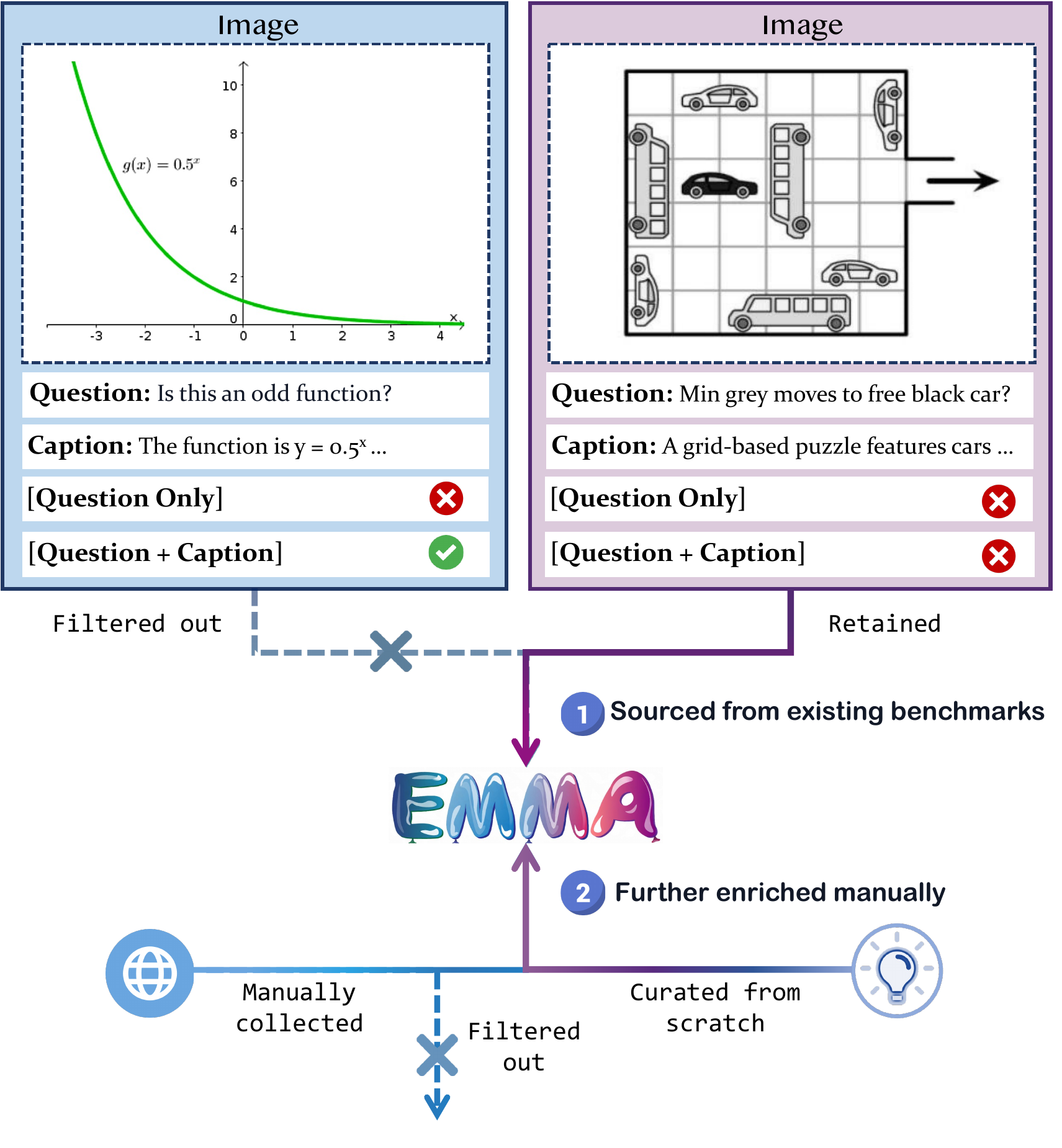}
    \caption{\textbf{Data curation process for EMMA.} We first apply a filtering pipeline to existing benchmarks, discarding problems that MLLMs can solve given only the text and image captions. Next, we categorize the remaining problems based on the skills they assess and manually collect or create additional problems to expand our dataset within these taxonomies.}
    \label{fig:compare_data_curation}
\end{figure}

As discussed in Section~\ref{sec:mmbench}, most existing multimodal reasoning benchmarks likely contain many problems that primarily measure text-based reasoning. To address this, we employ a two-step approach to constructing EMMA (Figure~\ref{fig:compare_data_curation}). First, we source problems from existing multimodal reasoning benchmarks and apply rigorous filtering to exclude those solvable through text-based reasoning and a single visual pass.
Next, we categorize the remaining problems for each subject into fine-grained multimodal reasoning skill taxonomies and manually collect more samples aligned with these taxonomies to expand our dataset.

\paragraph{Filtering Mechanisms} To filter for questions that require multimodal reasoning, Yue et al.~\cite{yue2024mmmu} provide only the text from multimodal reasoning questions to LLMs and discard questions that can be correctly answered this way. Nonetheless, some of the remaining questions may still not \textit{truly} measure visual reasoning, as a single pass of visual perception and language understanding may suffice to answer them. We extend ~\cite{yue2024mmmu} one step further (illustrated in Figure~\ref{fig:compare_data_curation}): we first caption the images in multimodal reasoning questions using GPT-4o and then pass both the original text and our generated captions to MLLMs, filtering out questions that can be answered under this condition. Specifically, for each candidate question, we query Llama-3-70B-Instruct~\cite{llama3}, GPT-4o, and Qwen2-72B-Instruct~\cite{qwen2} ten times; if any model answers a question correctly at least five times, we discard it. This more stringent filtering ensures that the remaining questions require models to engage deeply with visual information. We introduce the data collection process for each project in detail below.

\paragraph{Math}

We first apply the filtering pipeline to Math-Vision~\cite{wang2024measuring} and MathVista~\cite{lu2024mathvista}, and then manually inspect the remaining set and craft a taxonomy consisting of five categories with a strong focus on multimodal reasoning, including 3D Simulation, 2D Transformation, Path Tracing, Multi-hop Object Counting, and Pattern Inference. Next, we use GPT-4o to categorize all questions based on this taxonomy, followed by a manual verification. 
In addition, we supplement our benchmark with additional pattern inference questions from RAVEN~\cite{raven}, which inherently require multi-hop visual reasoning.This process results in a total of 892 math questions. This process ultimately results in

\paragraph{Physics}

We apply the filtering pipeline to multimodal physics problems in OlympiadBench~\cite{olympiadbench}, EXAMS-V~\cite{examsv}, and MMMU~\cite{yue2023mmmu}, which yields only 80 problems. In addition, we manually collect more problems online from Learn AP Physics~\cite{apphysics} and Khan Academy~\cite{khanacademy} and filter them, resulting in 76 more new problems. Through manual labeling, we verify that these problems span a wide range of topics, including 3D Field Simulation, Graph Reasoning, and Path Tracing. We note that despite our best efforts, multimodal physics problems meeting our criteria are difficult to source and construct. 

\paragraph{Chemistry} 
After filtering the chemistry portion of EXAMS-V~\cite{examsv} and MMMU~\cite{yue2023mmmu}, we are left with only 20 problems.
These questions mostly involve reasoning about molecular formulas, which inspires us to manually construct a novel test suite on organic chemistry. Based on molecular images in SMiCRM~\cite{smicrm}, we analyze their chemical properties with RDKit~\cite{landrum2013rdkit}, a computational chemistry toolkit, and develop 746 novel questions over chemical structure recognition, bond counting, and structure simulation. In addition, we draw from the collection of chemical reactions in~\cite{name_reactions} and collaborate with PhD students in chemistry to annotate reaction outcomes, contributing another 210 questions on reaction simulation. 

\paragraph{Coding} 
In addition to STEM questions, we design four types of coding tasks to capture various real-world use cases of MLLMs for visualization creation. For instance, to evaluate the ability to reproduce a visualization, we construct ``Vis Choose Code'' questions where models select the code that generates a target chart. Since previous data visualization benchmarks all rely on using MLLMs as judges, we do not source from existing benchmarks, but manually construct all coding questions from scratch through a three-stage process. First, we identify ``seed visualizations'' that employ advanced visualization techniques from CharXiv~\cite{wang2024charxiv}, the matplotlib example gallery~\cite{mpl} (following Wu et al.~\cite{wu2024plot2code}), and our prior experience. Next, we generate four variations for each seed visualization to form a ``set'' by introducing design variations (e.g., changes in spine configuration, line style, and axis scaling) either manually or through prompting MLLMs, with post-hoc manual verification. We also provide these design variations as labels for each problem. Finally, we construct four types of questions using these visualization sets. In addition to  ``Vis Choose Code'', we have ``Code Choose Vis'' questions that require models to identify the chart produced by a given code snippet. ``Modify'' questions, inspired by visualization debugging scenarios, provide a target chart, a code snippet, and the chart generated by the snippet, requiring models to select the modification needed to transform the code into the target chart. Similarly, ``Modify without Original Image'' presents the same task but excludes the image generated by the initial code snippet. In the end, our curation results in 564 multiple-choice coding questions in total.

\subsection{Comparison with Existing Benchmarks}

\begin{table*}[t]
\tablestyle{5pt}{1.1}
    \centering
    \small
    \resizebox{.98\textwidth}{!}
{
    \begin{tabular}{@{}l c c c c c c c c c c c@{}}
        \toprule
        \multirow{3}{*}{} & \multirow{3}{*}{\textbf{CoT}} & \multicolumn{5}{c}{\textbf{EMMA}} & \multicolumn{5}{c}{\textbf{EMMA-mini}} \\
        \cmidrule(rl){3-7} \cmidrule(rl){8-12}
         & & \textbf{Math} & \textbf{Phys.} & \textbf{Chem.} & \textbf{Coding} & \textbf{Overall} & \textbf{Math} & \textbf{Phys.} & \textbf{Chem.} & \textbf{Coding} & \textbf{Overall} \\
         & & (892) & (156) & (1,176) & (564) & (2,788) & (100) & (100) & (100) & (100) & (400) \\
    

        Random choice & $-$ & 14.01 & 25.64 & 16.50 & 25.71 & 18.08 & 13.00 & 23.00 & 27.00 & 28.00 & 22.75 \\
        Human Expert & $-$ & $-$ & $-$ & $-$ & $-$ & $-$ & 75.00 & 64.50 & 86.00 & 85.50 & 77.75 \\
        
        \midrule
    
        Claude 3.5 Sonnet & $\xmark$ & 25.34 & 33.97 & \second{40.90} & 38.65 & 35.08 & 23.00 & 34.00 & \best{44.00} & 35.00 & 34.00 \\
        Gemini 2.0 Flash & $\xmark$ & 23.88 & 38.46 & 36.31 & \second{42.02} & 33.61 & 20.00 & 40.00 & 36.00 & 41.00 & 34.25 \\
        GPT-4o & $\xmark$ & 27.24 & 38.46 & 31.89 & 40.07 & 32.42 & 30.00 & 38.00 & 33.00 & 40.00 & 35.25 \\
    
        \addlinespace[0.1em]\hdashline\addlinespace[0.1em]
        Qwen2-VL-72B-Instruct & $\xmark$ & \best{33.07} & 42.31 & 32.06 & 34.57 & 33.46 & \second{38.00} & 40.00 & 34.00 & 37.00 & 37.25 \\
        LLaVA-Onevision-72B & $\xmark$ & 27.69 & 35.90 & 25.26 & 28.72 & 27.33 & 25.00 & 32.00 & 24.00 & 28.00 & 27.25 \\
        InternVL2-Llama3-76B & $\xmark$ & 25.11 & 22.44 & 24.06 & 27.84 & 25.07 & 31.00 & 22.00 & 21.00 & 28.00 & 25.50 \\
        InternVL2.5-78B & $\xmark$ & 31.39 & 38.46 & 35.20 & 31.91 & 33.50 & 30.00 & 40.00 & 38.00 & 33.00 & 35.25 \\
    
        \hline
        Claude 3.5 Sonnet & $\cmark$ & 29.37 & 41.03 & \best{41.07} & 40.60 & \second{37.23} (\textcolor{ForestGreen}{$\uparrow 2.15$}) & 30.00 & 38.00 & \second{41.00} & 39.00 & 37.00 (\textcolor{ForestGreen}{$\uparrow 3.00$}) \\
        
        Gemini 2.0 Flash & $\cmark$ & 25.90 & 38.46 & 24.66 & 40.96 & 29.12 (\textcolor{Red}{$\downarrow 4.48$}) & 24.00 & 41.00 & 36.00 & \second{44.00} & 36.25 (\textcolor{ForestGreen}{$\uparrow 2.00$}) \\
        
        GPT-4o & $\cmark$ & 25.56 & \second{43.59} & 33.67 & 39.01 & 32.71 (\textcolor{ForestGreen}{$\uparrow 0.29$}) & 27.00 & 44.00 & 35.00 & 38.00 & 36.00 (\textcolor{ForestGreen}{$\uparrow 0.75$}) \\
        
        \addlinespace[0.1em]\hdashline\addlinespace[0.1em]
        Qwen2-VL-72B-Instruct & $\cmark$ & 27.69 & 34.62 & 24.57 & 29.43 & 27.12 (\textcolor{Red}{$\downarrow 6.35$}) & 35.00 & 34.00 & 32.00 & 23.00 & 31.00 (\textcolor{Red}{$\downarrow 6.25$}) \\
        
        LLaVA-Onevision-72B & $\cmark$ & 22.42 & 15.38 & 22.70 & 30.67 & 23.82 (\textcolor{Red}{$\downarrow 3.52$}) & 23.00 & 26.00 & 23.00 & 29.00 & 25.25 (\textcolor{Red}{$\downarrow 2.00$}) \\
        
        InternVL2-Llama3-76B & $\cmark$ & 22.20 & 32.05 & 19.73 & 30.32 & 23.35 (\textcolor{Red}{$\downarrow 1.72$}) & 27.00 & 33.00 & 21.00 & 32.00 & 28.25 (\textcolor{ForestGreen}{$\uparrow 2.75$}) \\
        
        InternVL2.5-78B & $\cmark$ & 25.56 & 39.74 & 27.47 & 25.18 & 27.08 (\textcolor{Red}{$\downarrow 6.42$}) & 31.00 & 36.00 & 24.00 & 19.00 & 27.50 (\textcolor{Red}{$\downarrow 7.75$}) \\
        
        \addlinespace[0.1em]\hdashline\addlinespace[0.1em]
        Gemini 2.0 Flash Thinking & $-$ & \second{31.61} & \best{56.41} & 37.93 & \best{43.44} & \best{38.06} & 35.00 & \best{57.00} & \second{41.00} & 41.00 & \second{43.50} \\
        o1 & $-$ & $-$ & $-$ & $-$ & $-$ & $-$ & \best{41.00} & \second{49.00} & 40.00 & \best{53.00} & \best{45.75} \\
        \bottomrule
    \end{tabular}
}
    \caption{\textbf{Evaluation results of state-of-the-art MLLMs, which are outperformed by human experts with wide margins.}
    The highest model performance in each column is highlighted in green, and the second-highest is highlighted in blue. Performance improvements from CoT are indicated with upward green arrows, while reductions are marked with downward red arrows. }
    \label{tab:main results}
\end{table*}

Our enhanced data filtering pipeline ensures that EMMA focuses on questions requiring in-depth multimodal reasoning, i.e., those that cannot be solved solely using text-based reasoning or a single visual pass. While MMMU-Pro~\cite{yue2024mmmu} removes questions solvable through their text portion alone, it may still retain problems for which visual reasoning is inessential. In contrast, EMMA applies a stricter filtering criterion, discarding questions solvable with text and image captions. 
For instance, the left example in Figure~\ref{fig:compare_data_curation} (adapted from MathVista) asks whether a depicted function is even or odd. Although unsolvable without the image, the problem can be shortcut by extracting the function's text expression embedded in the image. In this case, the role of vision is more aligned with visual perception than with visual reasoning. By eliminating such problems, which MMMU-Pro's approach would retain, EMMA better evaluates the multimodal reasoning capabilities of models.

We also contribute 1,796 novel multimodal reasoning problems across physics, chemistry, and coding. After filtering physics and chemistry problems from all relevant benchmarks to our knowledge (e.g., \cite{olympiadbench, examsv, yue2023mmmu}), only 100 remain. We expand this to 1,332 in EMMA by manually sourcing additional data and hiring domain experts. For coding, EMMA is the first benchmark to systematically evaluate data visualization skills using a multiple-choice format, enabling a standardized assessment and obviating the need for MLLMs as judges. Moreover, through meticulous manual labeling or verification, we provide fine-grained labels for each question (Figure~\ref{fig:categories}), categorizing them based on the specific skills they assess. These labels enable a detailed analysis of MLLM performance, as we demonstrate in Section~\ref{sec:cot}.

\section{Experiments}
\label{sec: experiment}

\subsection{Evaluation Settings}
\paragraph{Data Split}
To create a more balanced subset of EMMA, we randomly sample 400 questions (100 per subject) from the benchmark, hereafter referred to as EMMA-mini. Within each subject, we aim for equal representation across categories to the extent possible.

\paragraph{Human Performance}
To estimate expert-level performance on EMMA-mini, we hire two human experts per subject and report their average score. This score serves as a baseline contextualizing model performance.

\paragraph{Models}
We evaluate nine state-of-the-art MLLMs under the zero-shot setting, including four open-source models (Qwen2-VL (72B) \cite{wang2024qwen2}, LLaVA-Onevision (72B) \cite{li2024llava}, InternVL2 (76B) \cite{internVL}, and InternVL2.5 (78B) \cite{internvl2.5}) and five proprietary ones (GPT-4o \cite{4o}, Claude~3.5~Sonnet \cite{claude}, Gemini~2.0~Flash \cite{gemini-2-flash}, Gemini~2.0~Flash~Thinking \cite{gemini-2-flash-thinking}, and o1 \cite{o1}). Due to rate limits, we report o1  performance on EMMA-mini only. All other models are evaluated on the entire benchmark.

\paragraph{Prompting Strategies}
For all models except o1 and Gemini~2.0~Flash~Thinking, we test two prompting strategies: (1) \textit{Direct} prompting, which instructs models to output the answers without reasoning steps;  and  (2) Chain-of-Thought (\textit{CoT}) prompting~\cite{wei2022chain}, where we prompt models to ``think step-by-step'' and output responses in a structured format.

\subsection{Main Results}
Table~\ref{tab:main results} compares the performance of different MLLMs and prompting strategies. 

\begin{table*}[t!]
\tablestyle{10pt}{1}
    \centering
    \small
    \resizebox{.98\textwidth}{!}
{
    \begin{tabular}{lllccccc}
    \toprule
    \textbf{Model}  & \textbf{Method} & \textbf{Reward Model} & \textbf{N=1} & \textbf{N=2} & \textbf{N=4} & \textbf{N=8} & \textbf{N=16} \\
    \midrule
    \multirow[c]{5}{*}{\begin{tabular}[c]{@{}c@{}}GPT-4o\end{tabular}} 
    & Majority Voting & $-$ & \multirow{5}{*}{36.00}  & $-$ & 37.25 & 36.25 & 38.25\\
    & BoN & GPT-4o (Self) &  & 35.50 & 35.75 & 36.75 & $-$ \\
    & BoN & Gemini Flash Thinking &  & \second{40.75} & 36.25 & 36.5 & $-$ \\
    & Tournament & Gemini Flash Thinking &  & \second{40.75} & 39.25 & \best{41.25} & 35.25\\
    & \demph{Pass@N} & \demph{$-$} &  & \demph{45.00} & \demph{53.25} & \demph{65.75} & \demph{74.00} \\
    \cmidrule(lr){1-8}
    \multirow[c]{5}{*}{\begin{tabular}[c]{@{}c@{}}Gemini 2.0 Flash\end{tabular}} 
    & Majority Voting & $-$ & \multirow{5}{*}{36.25} & $-$ & 37.75 & 39.25 & 39.75 \\
    & BoN & Gemini Flash (Self) &  & 38.25 & 36.50 & 36.00 & $-$\\
    & BoN & Gemini Flash Thinking &  & 36.75 & 37.00 & \second{40.25} & $-$\\
    & Tournament & Gemini Flash Thinking &  & 36.75 & 37.25 & \best{40.75} & 38.75\\
    & \demph{Pass@N} & \demph{$-$} &  & \demph{45.25} & \demph{56.25} & \demph{64.50} & \demph{75.00} \\
    \midrule
    \multirow[c]{3}{*}{\begin{tabular}[c]{@{}c@{}}Gemini~2.0~Flash~Thinking\end{tabular}} 
    & Majority Voting & $-$ & \multirow{3}{*}{43.50} & $-$ & 48.00 & \second{49.00} & \best{50.75}  \\
    & Tournament & Gemini Flash Thinking (Self) &  & 45.50 & 47.25 & 47.25 & 48.00 \\
    & \demph{Pass@N} & \demph{$-$} &  & \demph{53.75} & \demph{64.50} & \demph{71.50} & \demph{81.50} \\
    \cmidrule(lr){1-8}
    o1 & $-$ & $-$ & \best{45.75} & $-$ & $-$ & $-$ & $-$ \\
    \bottomrule
    \end{tabular}}
    \caption{\textbf{Results of different test-time scaling strategies on EMMA-mini.} We also include Pass@N accuracies as upper bounds to scaling performance. While test-time scaling tends to improve model accuracy, they do not help models achieve near-human-level performance. Overall, Gemini 2.0 Flash Thinking is the best reward model. It also benefits the most from test-time scaling, gaining 7.25\% in accuracy with majority voting at N=16 over N=1.}
    \label{tab:scaling}
\end{table*}

\paragraph{Are MLLMs Multimodal Reasoners?} 
Table~\ref{tab:main results} demonstrates that all models perform suboptimally across the subjects in EMMA. On EMMA-mini, the best-performing model, o1, achieves an accuracy of 45.75\%, trailing human experts by 32\%. At the lower end, LLaVA-OneVision-72B scores only 25.25\%, barely surpassing random choice by 2.5\%. Drilling down into subjects, the best models show the smallest gap with human performance on physics, with Gemini 2.0 Flash Thinking scoring 7.5\% lower than human experts. This smaller gap may reflect the inherent difficulty of physics problems, leading human experts to achieve a score of 64.5\%. For other subjects, however, \textbf{the best-performing models lag significantly behind human experts, with gaps of 34\%, 42\%, and 32.5\% in math, chemistry, and coding, respectively.} These results underscore the limitations of current MLLMs in addressing complex multimodal reasoning tasks.

On the full EMMA benchmark, closed-source models generally outperform open-source ones, particularly with CoT prompting. Across all subjects, Qwen2-VL-72B-Instruct is the only open-source model to place in the top two for any subject. Gemini 2.0 Flash Thinking scores best overall, ranking among the top two models in three out of four subjects, further highlighting the advantages of optimizing models for reasoning by training them to generate thought processes over traditional MLLM paradigms (note that o1 is not evaluated on the full set due to rate limits). In particular, Gemini 2.0 Flash Thinking performs exceptionally well in physics, leading by nearly 13\% over the second-best model, GPT-4o. In contrast, Claude 3.5 Sonnet excels in chemistry, surpassing the next best model, Gemini 2.0 Flash Thinking, by over 2\%. These results suggest interesting comparative strengths between models, which may result from differences in training data.

\paragraph{Does CoT help?} We observe divergent tendencies in the effectiveness of CoT prompting across both closed- and open-source models. 
We exclude o1 and Gemini 2.0 Flash Thinking from this analysis, as these models inherently generate CoT as part of their responses. Under direct prompting, accuracies achieved by the best open-source models are well within 2\% of Claude 3.5 Sonnet (the best closed-source model). However, the gap widens significantly under CoT prompting, with the best open-source model underperforming by almost 10\%. Comparing each model's performance with and without CoT on EMMA and EMMA-mini, \textbf{CoT prompting generally improves performance for closed-source models, while reduces performance for open-source models}. Notably, Qwen2-VL-72B-Instruct and InternVL2.5-78B, the top two open-source models overall under direct prompting, suffer decreases of over 6\% in accuracies on both EMMA and EMMA-mini. While some tasks might not benefit significantly from textual CoT, we hypothesize that this divergence arises because open-source models fail to fully leverage the potential of language to assist in multimodal reasoning tasks where language could be helpful. We elaborate on this hypothesis in detail in Section~\ref{sec:cot}.

\subsection{Results with Test-Time Compute Scaling}

In this section, we test three test-time compute scaling methods~\cite{snell2024scaling, yang2024qwen2, o1} on EMMA-mini: majority voting, Best-of-N selection, and Tournament-Style selection. Both Best-of-N and Tournament-Style selection require a reward model to select the best response among multiple candidates. We use CoT prompting to generate the candidate responses, so that the reward model has enough context to score the responses. For each test-time scaling method, we experiment with N = 1, 2, 4, 8, and 16, as long as the context length of the reward model allows.

\begin{itemize}
\item \textbf{Majority Voting:} Majority voting selects the most frequent response among batches of N candidate responses. When there is a tie, we randomly choose one among the most frequent answers.

\item \textbf{Best-of-N:} Best-of-N selection~\cite{cobbe2021training, lightman2023let} selects the highest-scoring response according to a reward model. We explore two configurations: using the base model itself or a stronger reasoning model (\eg Gemini 2.0 Flash Thinking) as the reward model.

\item \textbf{Tournament-Style Selection:} In Tournament-Style Selection~\cite{tournament, o1-pro}, responses are paired in matches, and the winners progress through successive rounds until a final selection is made.\footnote{Tournament-style selection with N=2 is equivalent to Best-of-2.} 
We use the best-performing reward model identified in the Best-of-N experiments, which is Gemini 2.0 Flash Thinking. 
\end{itemize}

\paragraph{Does test-time compute scaling help?} 
Tables~\ref{tab:scaling} presents results of test-time compute scaling methods on top of GPT-4o, Gemini 2.0 Flash, and Gemini 2.0 Flash Thinking. 

\textbf{Overall, test-time compute scaling improves model performance, but it fails to close the gap to human expert performance.
}  
The highest accuracy improvements are 5.25\% for GPT-4o, 4.5\% for Gemini 2.0 Flash, and 7.5\% for Gemini 2.0 Flash Thinking. Notably, without test-time scaling (N=1), Gemini 2.0 Flash Thinking's accuracy is 2.25\% lower than that of o1, but it overtakes o1 by 5\% with majority voting at N=16. Nonetheless, its best performance still lags human performance by 27\%.

\begin{figure}[t]
    \centering 
    \includegraphics[width=\columnwidth]{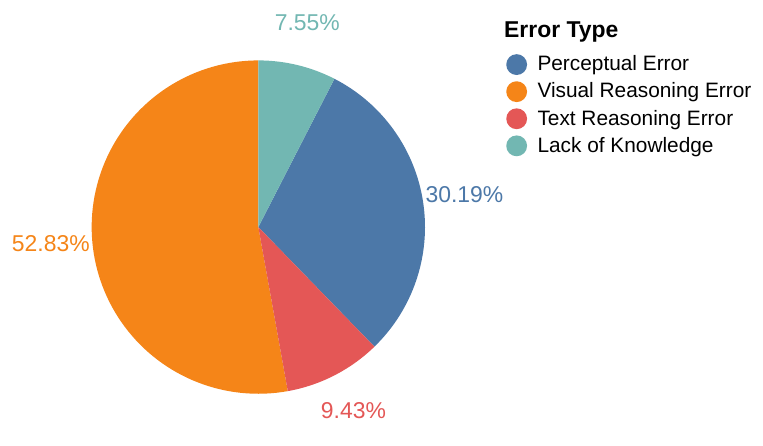}
    \caption{\textbf{Distribution of error types made by o1} on the math and coding portions of EMMA-mini. The majority of errors arise in visual reasoning.}
    \vspace{-3pt}
    \label{fig:error dist}
\end{figure}

We also observe distinct patterns in test-time compute scaling performance across different models. While scaling beyond N=8 for GPT-4o and Gemini 2.0 Flash leads to performance degradation, Gemini 2.0 Flash Thinking continues to benefit incrementally from additional test-time compute, at least up to N=16. In fact, stronger base models also achieve higher Pass@N accuracy: Gemini 2.0 Flash Thinking's Pass@N consistently surpasses those of the other two models by around 7\%, suggesting that a stronger base reasoner is more likely to cover the correct response when given multiple attempts. In sum, these results suggest that using a stronger model as the base model raises the upper bound for test-time scaling. 

Comparing scaling strategies for each model, we find that GPT-4o and Gemini 2.0 Flash achieve their greatest improvements when Gemini 2.0 Flash Thinking is used as the reward model. Additionally, tournament-style selection consistently outperforms Best-of-N (BoN) selection. These results suggest that employing a stronger model as the reward model enables less capable models to achieve better results, particularly when the reward model can make fine-grained decisions involving a couple candidate responses each time. This is intuitive, as evaluating responses also requires reasoning.

On the other hand, we find that self-reward modeling tends to be perform suboptimally. Even using Gemini 2.0 Flash Thinking for self-reward modeling yields performance consistently below that of majority voting. We conjecture that self-reward modeling may be less effective because the model's evaluation criteria may be disrupted by its own generation patterns, making it less sensitive to differences in the reasoning of the generated responses compared to an independent reward model.

\section{Error Analysis}

\begin{figure}[t]
    \centering 
    \includegraphics[width=\columnwidth]{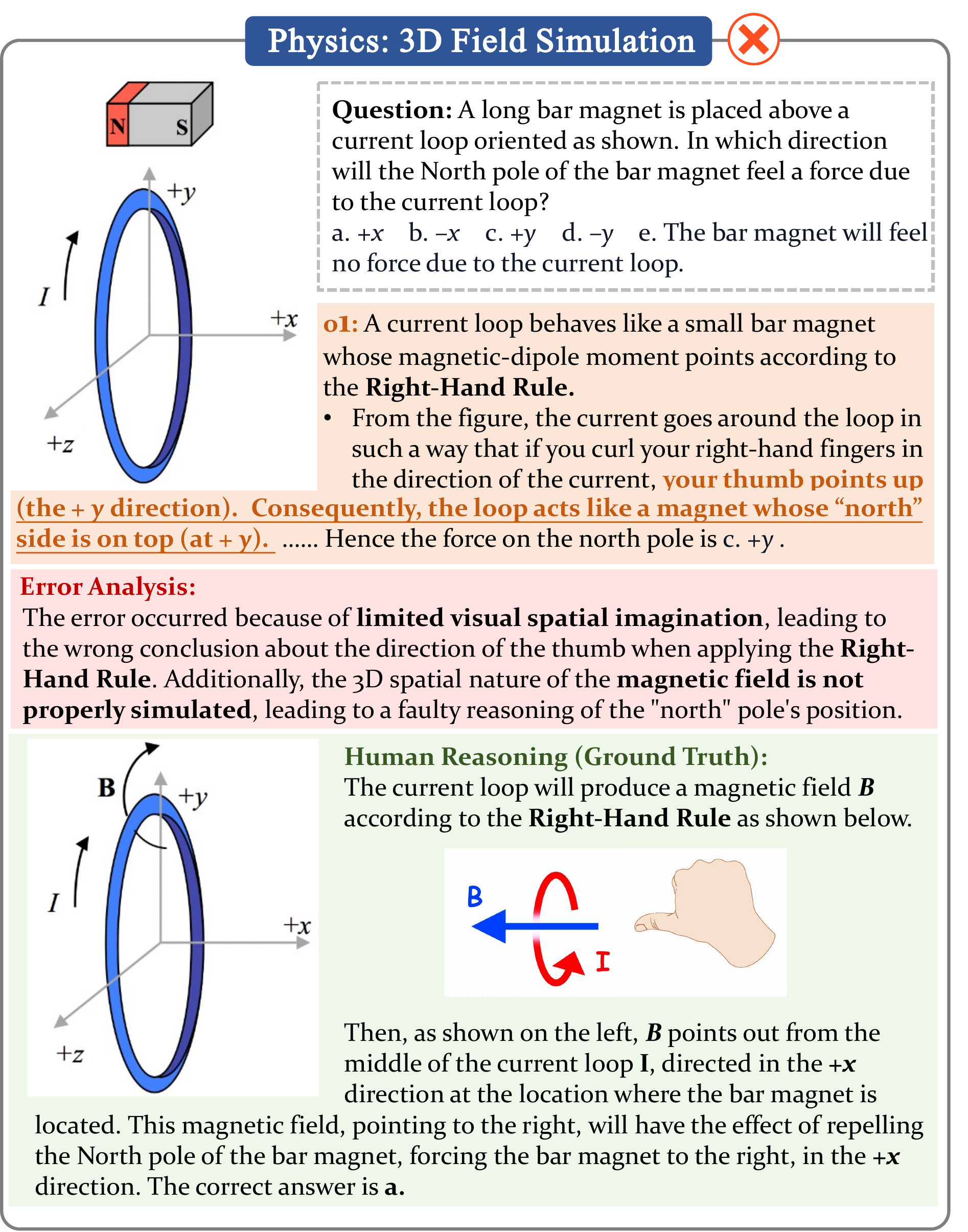}
    \caption{\textbf{A representative example of visual reasoning error.} While o1 recognizes that solving this problem involves the Right-Hand Rule, it misapplies the rule due to limited visual spatial simulation skills, resulting in an incorrect conclusion about the direction of the magnetic field.}
    \label{fig:visual reasoning error}
\end{figure}

\subsection{Error Distribution}
We present an analysis of the errors made by o1 on the math and coding portions of EMMA-mini. In total, o1 incorrectly answers 59 math questions and 47 coding questions. Figure~\ref{fig:error dist} categorizes these errors into four types. \textbf{Perceptual errors}, such as misinterpreting visual information, account for 30.19\% of all errors. \textbf{Lack of knowledge errors}, including mistakes related to API usage, contribute 7.55\%. \textbf{Visual reasoning errors}, such as failures to simulate 3D processes, constitute the largest category at 52.83\%. Finally, \textbf{textual reasoning errors}, including calculation mistakes or logical missteps, represent 9.43\%. The predominance of visual reasoning errors underscores the limitations of current models in addressing complex visual reasoning tasks. Figure~\ref{fig:visual reasoning error} illustrates a representative case: while o1 correctly identifies that the problem calls for the application of the Right-Hand Rule, it fails to simulate where the thumb would point to when the right-hand fingers are curled in the direction of the current. We provide more error cases in the Appendix.

\subsection{The Effects of Textual CoT} 
\label{sec:cot}

\begin{figure}[t]
    \centering 
    \includegraphics[width=\columnwidth]{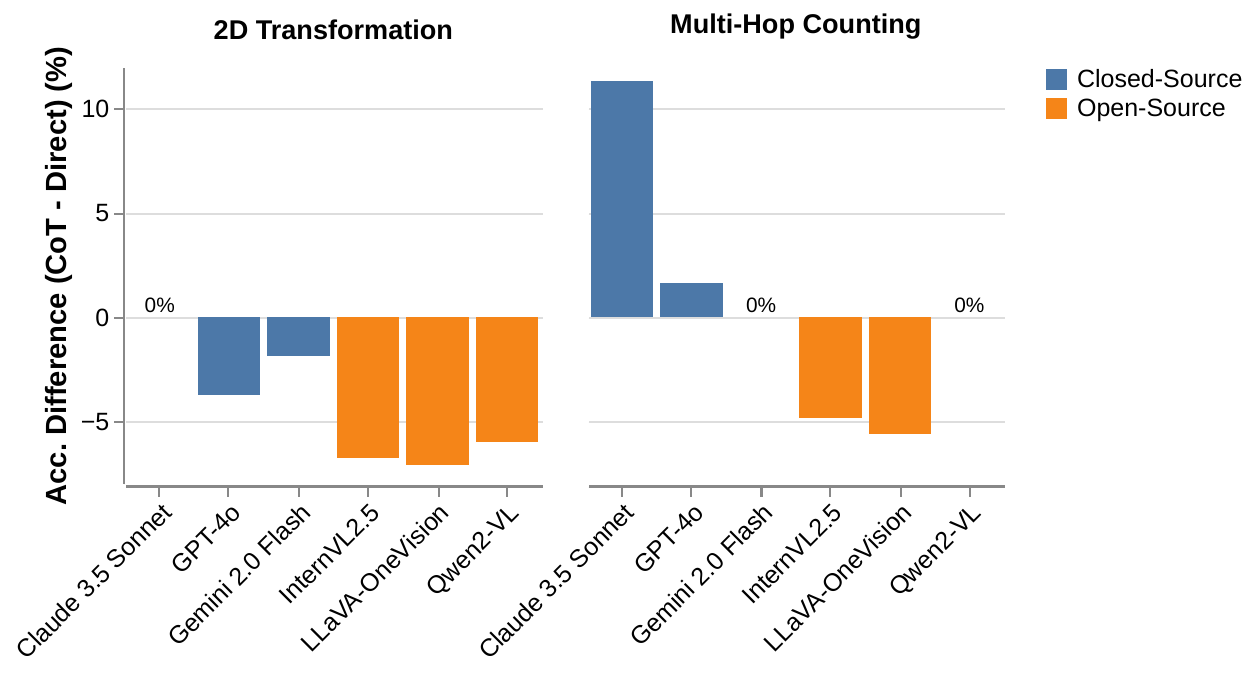}
    \caption{\textbf{Performance differences with and without CoT} for two math tasks: 2D Transformation and Multi-Hop Object Counting. The observed discrepancies in CoT's effectiveness suggest that its impact might depend on the extent of visual reasoning required to solve the task.
    }
    \label{fig:task diff}
\end{figure}

Results in Table~\ref{tab:main results} reveals a notable discrepancy: while CoT prompting improves reasoning in closed-source MLLMs, it tends to reduce performance in open-source models. Although the cause remains unclear without direct access to training data or methods, we analyze error rate per skill category to propose a conjecture. Figure~\ref{fig:task diff} compares accuracy differences between CoT and Direct prompting for six models on two math tasks: 2D Transformation and Multi-Hop Object Counting. On 2D Transformation, all but one model shows reduced performance with CoT; on Multi-Hop Object Counting, while CoT generally helps closed-source models, it makes all but one open-source model perform worse. In fact, we notice that CoT prompting introduces more hallucinations for open-source models.

\begin{figure}[t]
    \centering 
    \includegraphics[width=\columnwidth]{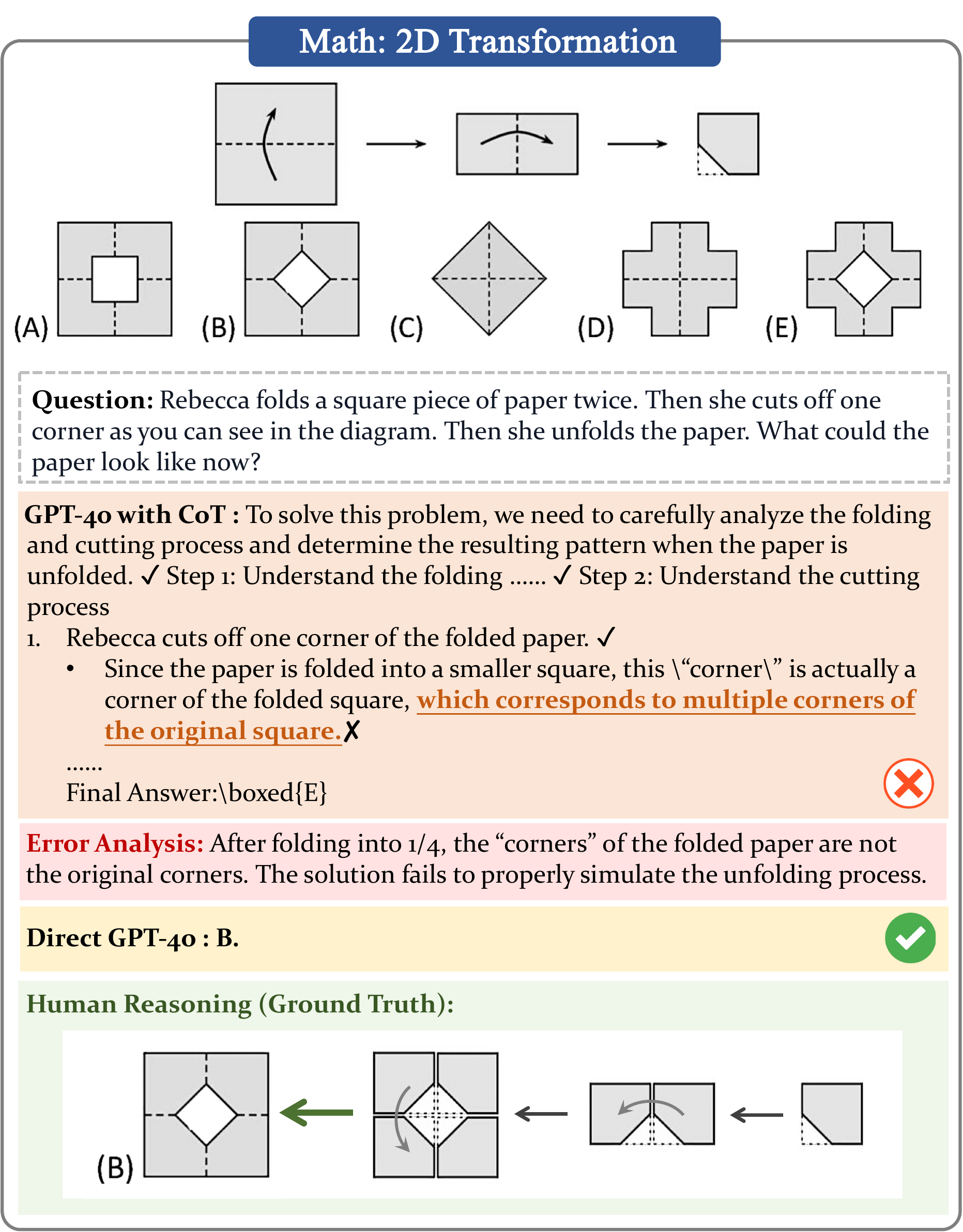}
    \caption{\textbf{An example where textual CoT is unhelpful for solving the problem with current models.} While GPT-4o correctly resolves the problem without CoT, it answers incorrectly with CoT. The thought process demonstrates a superficial association with spatial manipulations and relations rather than genuine visual reasoning.}
    \label{fig:cot unhelpful}
\end{figure}

2D Transformation questions primarily test for visual simulation and spatial imagination, which are difficult to verbalize. The question in Figure~\ref{fig:cot unhelpful}, for example, calls for spatial imagination beyond the power of language. In contrast, Multi-Hop Object Counting can leverage language to describe the relative positions of objects. Hence, we conjecture that visual-centric tasks, such as 2D Transformation, are poorly suited for textual CoT. In contrast, tasks that benefit from language-based reasoning, such as Multi-Hop Object Counting, allow models to achieve greater performance gains with textual CoT, as evidenced by the closed-source models.

\section{Conclusion}
We contribute EMMA, an Enhanced MultiModal reAsoning benchmark. EMMA features multimodal questions requiring advanced cross-modal reasoning, which cannot be solved by independently reasoning within each modality. Evaluation of nine state-of-the-art MLLMs reveals a substantial performance gap compared to human experts on EMMA, with techniques such as Chain-of-Thought prompting and test-time compute scaling offering only marginal gains. 
EMMA highlights the need for new architectures and training paradigms that can better integrate and reason over diverse modalities.
Like any benchmark, EMMA has its limitations, which can be improved in future works. 
For example, future iterations could enrich the currently underrepresented physics section or expand the chemistry section to incorporate a broader range of chemistry topics. Nonetheless, EMMA sets a new standard for assessing MLLMs on multimodal reasoning. 

\newpage
{\small
\bibliographystyle{ieee_fullname}
\bibliography{egbib}
}

\clearpage
\appendix
\section{Overview of the Appendix}

This Appendix is organized as follows:
\begin{itemize}
    \item Section~\ref{sec: EMMA Details} contains details about the composition of EMMA, the data curation process, and comparison with existing benchmarks;
    \item Section~\ref{sec: Experiment Details} contains experimental details, including the prompts used, models tested, hyperparameter settings, and the breakdown results on different categories;
    \item Section~\ref{sec:case} contains additional case studies for each subject.
\end{itemize}

\section{EMMA Details}
\label{sec: EMMA Details}

\subsection{Composition of EMMA}
EMMA comprises 2,788 questions across four subjects: math, physics, chemistry, and coding. We now provide a detailed breakdown of EMMA by subject:

\paragraph{Math} 
The math portion of EMMA consists of 892 questions, of which 562 are multiple-choice questions and 330 are free-form questions.
These questions can be categorized into five areas: \textbf{2D Transformation} (266 questions); \textbf{3D Spatial Simulation} (275 questions); \textbf{Path Tracing/Change of View simulation} (127 questions); \textbf{Multi-Hop Visual Object Counting} (124 questions); and \textbf{Pattern Inference} (100 questions). 

\begin{figure}[h]
    \centering 
    \includegraphics[width=\columnwidth]{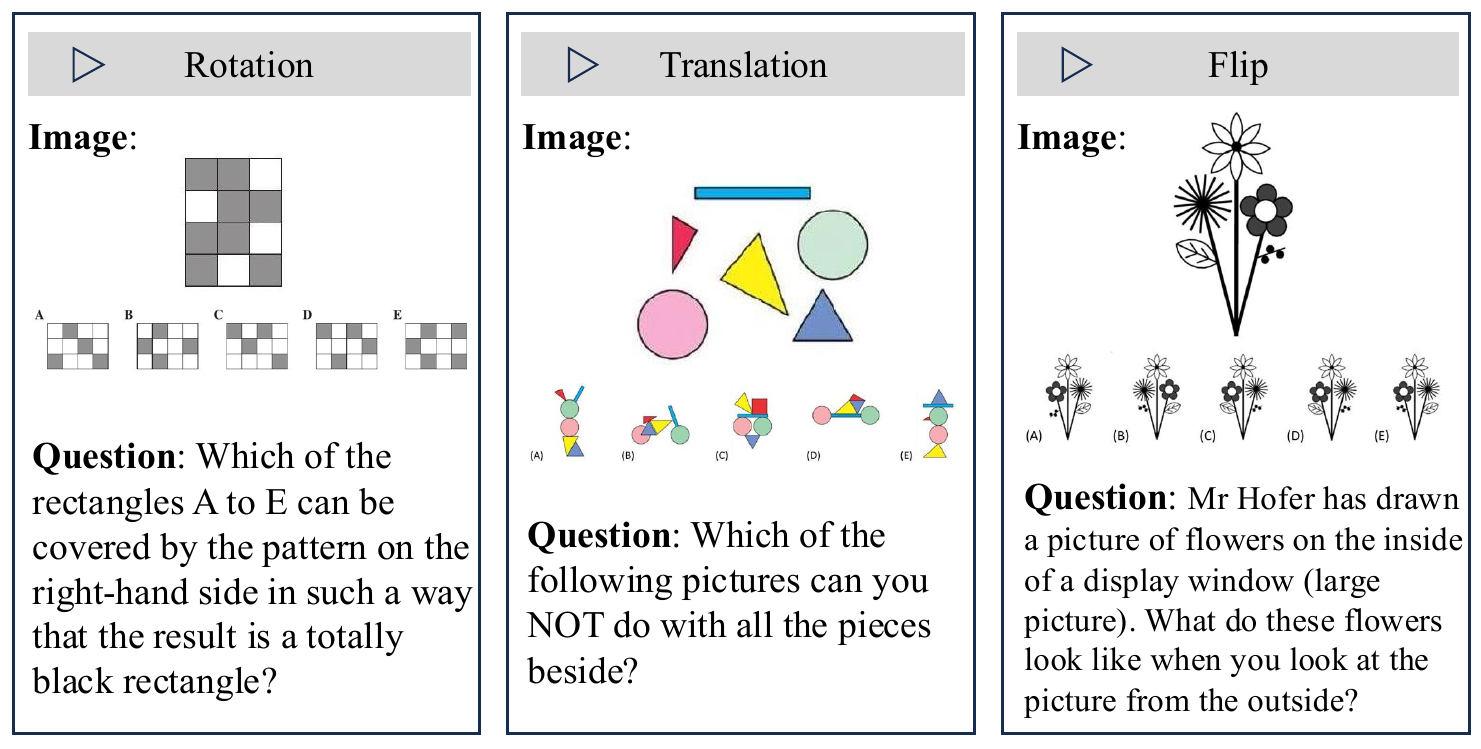}
    \caption{Three main types of questions belonging to the 2D Transformation category in math: Rotation, Translation, and Flipping.}
    \label{fig: math-case-2D}
\end{figure}

\begin{figure}[h]
    \centering 
    \includegraphics[width=\columnwidth]{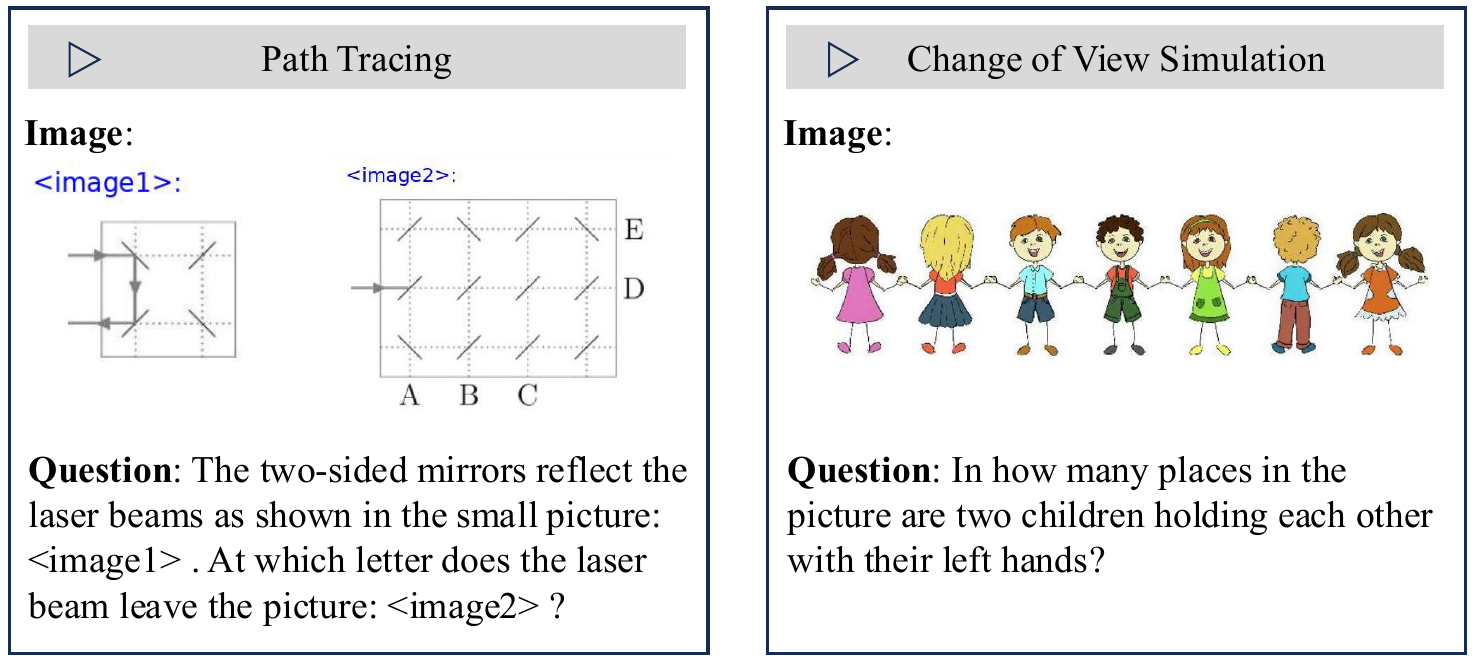}
    \caption{Two typical examples of the Path Tracing / Change of View Simulation category in math.}
    \label{fig: math-case-path}
\end{figure}

\begin{figure}[h]
    \centering 
    \includegraphics[width=\columnwidth]{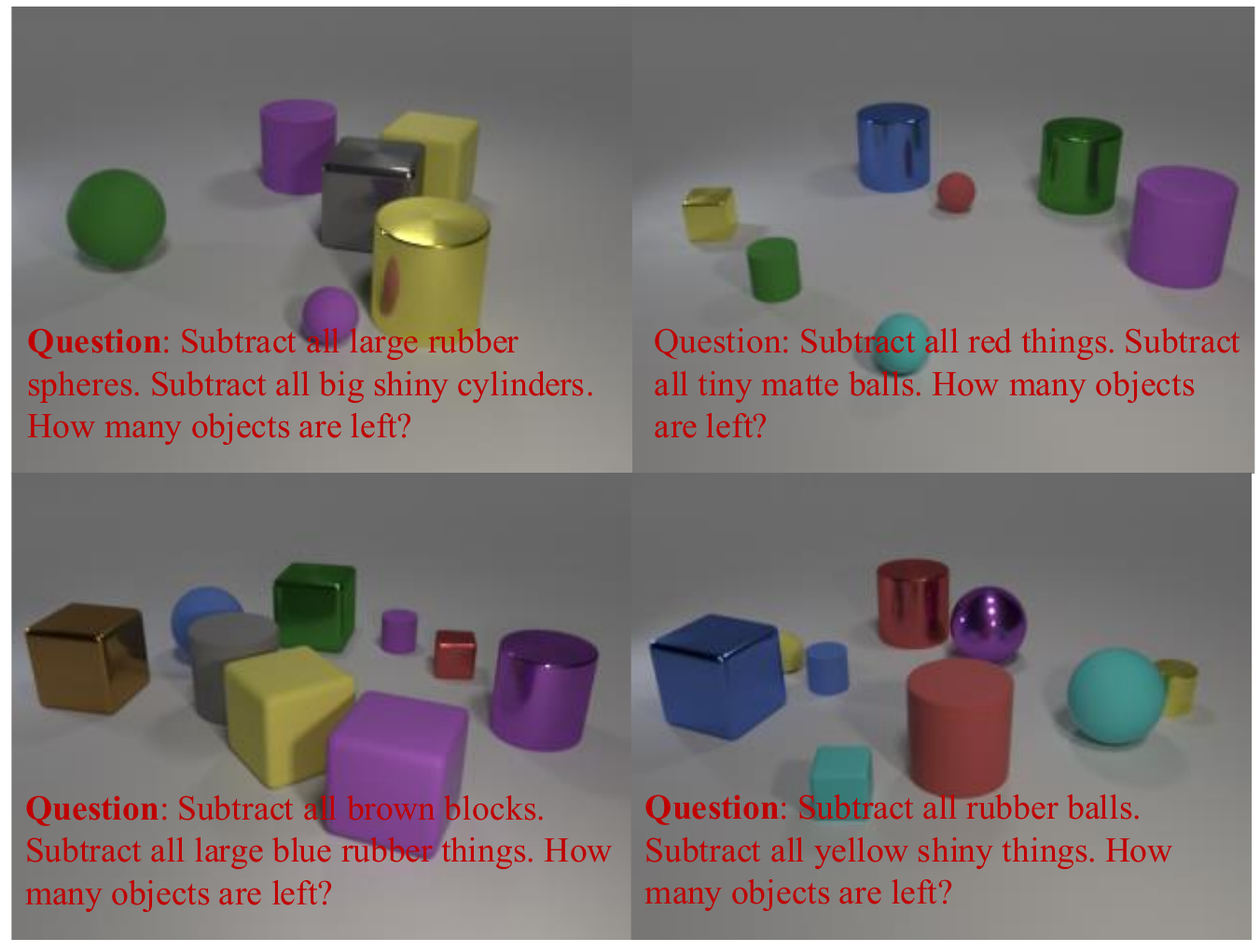}
    \caption{Four typical problems from the Multi-Hop Visual Object Counting category in math. }
    \label{fig: math-case-counting}
\end{figure}

\begin{figure}[h]
    \centering 
    \includegraphics[width=\columnwidth]{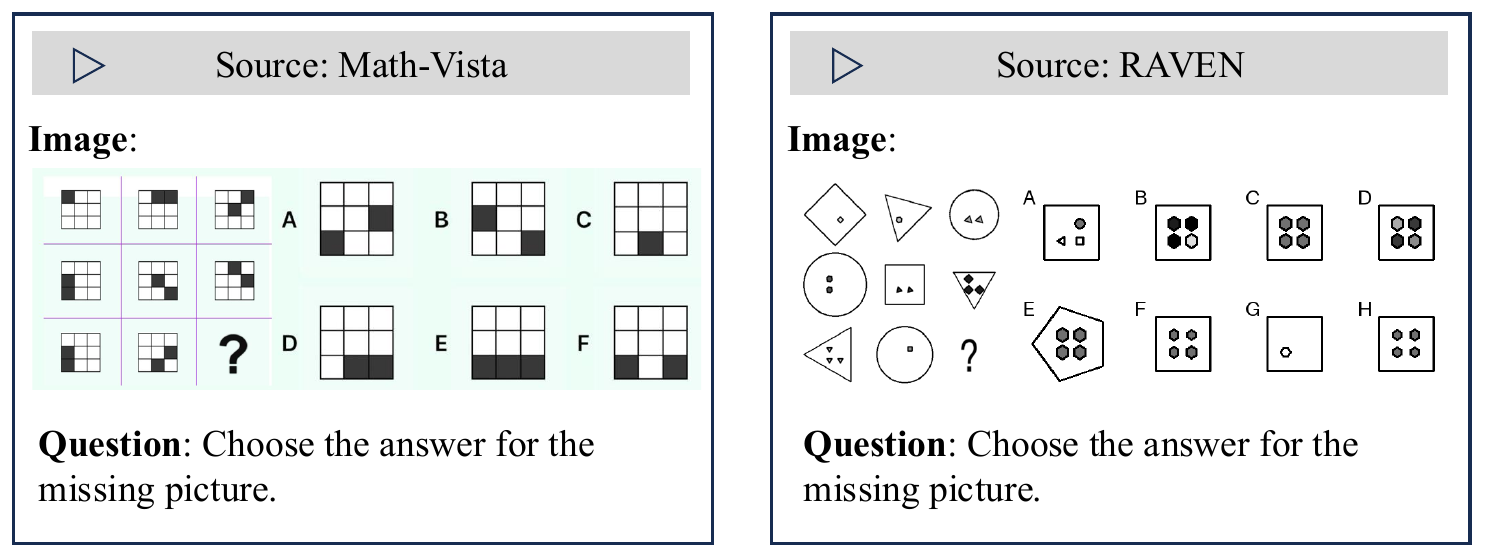}
    \caption{Questions in the Pattern Inference category for math are sourced from two datasets: Math-Vista~\cite{lu2024mathvista} and RAVEN~\cite{raven}. Two examples from these sources are presented for illustration.}
    \label{fig: math-case-pattern}
\end{figure}

Tasks in the \textbf{2D Transformation} category often involve operations such as rotating, translating, or flipping shapes. Examples are provided in Figure~\ref{fig: math-case-2D}. Humans typically solve these problems by leveraging their ability to ``see'' and mentally manipulate objects, simulating spatial transformations to arrive at a solution. During data filtering, we observe that models also rely heavily on visual information to solve these problems, as they often fail when provided only with textual descriptions of the accompanying images. Similarly, problems in the \textbf{3D Spatial Simulation} category require a similar visual reasoning approach, but with the key difference that the simulation must be performed in three-dimensional space. The \textbf{Path Tracing/Change of View Simulation} category involves solving problems akin to maze navigation, where the task requires tracing a path from a starting point to an endpoint while considering changes in perspective. We present two typical examples in Figure~\ref{fig: math-case-path}. 
Problems in the \textbf{Multi-Hop Visual Object Counting} category are sampled from Math-Vista~\cite{lu2024mathvista}, with some examples shown in Figure~\ref{fig: math-case-counting}. Unlike straightforward object counting, which might ask, ``How many objects are present?'', these questions require models to identify objects based on their attributes and perform subtraction operations grounded in visual properties. The \textbf{Pattern Inference} category involves identifying how shapes or colors evolve across a series of diagrams and predicting the next pattern in the sequence. Solving such problems draws on the ability to recognize visual regularities, which are challenging to describe accurately using text alone, necessitating strong visual reasoning. Typical examples are provided in Figure~\ref{fig: math-case-pattern}.

\paragraph{Physics}  

The physics portion of EMMA includes 156 questions divided into five categories: \textbf{3D Field Simulation}~(37 questions), \textbf{Graph Reasoning}~(26 questions), \textbf{Multi-Hop Visual Reasoning}~(33 questions), \textbf{Path Tracing}~(13 questions), and \textbf{Visual Decomposition Simulation}~(47 questions). All physics questions in EMMA are multiple-choice questions.

The \textbf{Visual Decomposition Simulation} category (47 questions) is often related to topics in physics such as Dynamics, Circular Motion, \& Gravitation. It typically involves the analysis of forces acting on static or dynamic objects, which requires visual decomposition of forces and visual simulation of current and future states.
The \textbf{3D Field Simulation} category (37 questions) addresses topics like Electric Force, Field, and Potential. These tasks emphasize visual simulations of properties and phenomena in three-dimensional electric and magnetic fields.
The \textbf{Graph Reasoning} category (26 questions) involves interpreting and reasoning about physics-related graphs, such as velocity-time (v-t) graphs, displacement-time (s-t) graphs, and trajectory graphs.
The \textbf{Multi-Hop Visual Reasoning} category (33 questions) is a mixed category, which includes some circuit analysis diagrams, as well as various types of problems that require multi-hop thinking.
In the \textbf{Path Tracing} category (13 questions), most problems involve light refraction. These tasks require analyzing the paths of particles by tracing their trajectories based on the perspectives depicted in the images.

\paragraph{Chemistry}  
The chemistry portion of EMMA includes 1,176 questions. 
Only 20 of them come from existing benchmarks, \textbf{the vast majority of 1,156 are created by us.}
These questions can be categorized into five areas: \textbf{Knowledge-based Counting} (456 questions); \textbf{Graph Reasoning} (9 questions); \textbf{Structure Recognition} (474 questions); \textbf{Reaction Simulation} (132 questions); and \textbf{Reaction Simulation Pro} (105 questions; Based on the test Reaction Simulation skill, the requirements for visual reasoning are more professional, where all options in multiple-choice questions are images). Among the chemistry questions, only the answer corresponding to the category Knowledge-based Counting is free-form, and all the remaining questions are multiple-choice.

EMMA prioritizes chemistry problems that require rich multimodal reasoning over simple fact recall or direct applications. To fill the gaps in current datasets, we manually developed a new test suite concentrating on organic chemistry because we found that this type of problem makes up the majority of the remaining data set filtered. In these newly developed sections, we focus on ``arrow-pushing'' diagrams, a common representation used to illustrate electron flow in mechanistic steps.
The dataset includes structural molecular identifiers of molecular images in chemical reaction mechanisms. The questions are categorized into three types in increasing reasoning difficulty: \textbf{Knowledge-based Counting}, \textbf{Structure Recognition}, and \textbf{Reaction Simulation}.

The \textbf{Knowledge-based Counting} category (456 questions) involves counting the number of chemical bonds of a chemical structure. The task requires domain-specific knowledge and multiple inference steps to accurately count different bonds.
\textbf{Structure Recognition} (474 questions) presents a more difficult task. It requires correctly identifying the number and type of atoms and chemical bonds, recognizing the molecular structure, and deriving the corresponding chemical expression or SMILES expression. Notably, each molecule in the diagram corresponds to a unique SMILES expression, encapsulating both compositional and structural information.

\textbf{Reaction Simulation} (132 questions) is the task requiring the most advanced visual reasoning. It involves inferring the post-reaction SMILES/Chemical expression based on the molecular composition and structural information available before the reaction, guided by the direction of electron flow indicated by arrows. Due to its complexity and the poor performance of models on the simpler open-ended tasks, this task is presented in the form of multiple-choice questions. 
\textbf{Reaction Simulation Pro} (105 questions) assess reaction simulation skills and visual reasoning with a higher level of expertise. Each question presents options in the form of images, requiring participants to not only simulate reactions but also engage in complex multi-step reasoning by comparing different choices.
\textbf{Graph Reasoning} (9 questions) involves graph-based reasoning problems related to chemistry knowledge, such as reaction rate changes.

\begin{figure*}[h]
    \centering 
    \includegraphics[width=\textwidth]{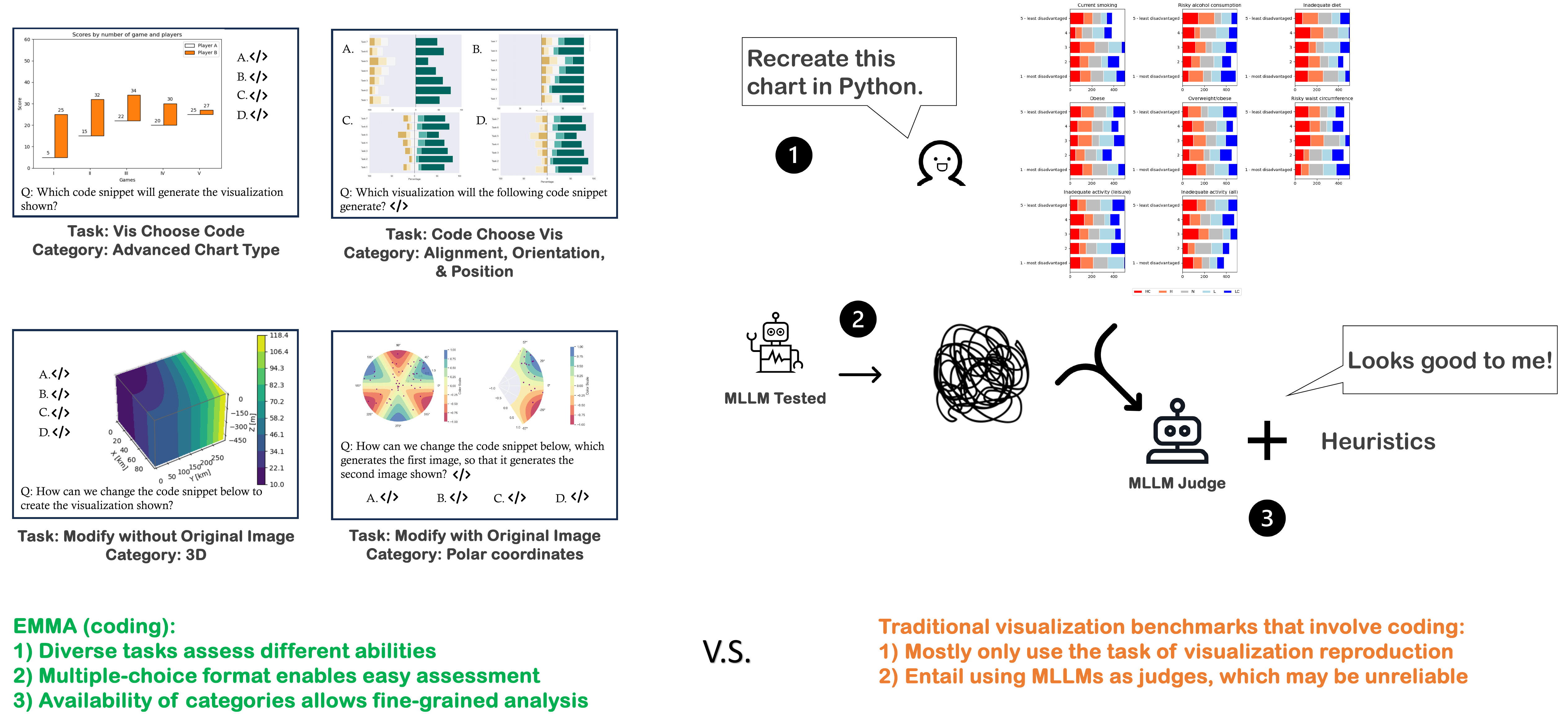}
    \caption{Traditional visualization benchmarks for coding often require MLLMs to recreate a target visualization, with evaluations conducted using a combination of MLLMs as judges and heuristic methods. In contrast, EMMA introduces four visualization-related coding tasks designed to assess multimodal coding abilities across multiple dimensions. By employing a multiple-choice format, EMMA eliminates the reliance on potentially unreliable MLLM-based judgment. Additionally, our fine-grained categories facilitate a detailed analysis of the limitations of multimodal coding skills.}
    \label{fig:coding advantage}
\end{figure*}

\paragraph{Coding}
\label{coding description}
Implementing user interfaces (UIs) is a fundamental task in software engineering. In this work, we focus on a critical aspect of UI development: data visualization. Creating data visualizations not only requires working knowledge of the charting toolkits, but also demands reasoning over how various visual elements coordinate to achieve the desired results. To stress-test MLLMs' visualization skills, we design four tasks: \textbf{Visualization Choose Code} (188 questions; given an image of a visualization, choosing which visualization program generates it), \textbf{Code Choose Visualization} (188 questions; given a visualization program, choosing which image it generates), \textbf{Modify without the Original Image} (94 questions; given a target visualization image and a visualization program that does not yet generate the target image, choosing what change should be applied to the program to create the target image), and \textbf{Modify with the Original Image} (94 questions; given a target visualization image, a visualization program that does not yet generate the target image, and the image that the current program generates, choosing what change should be applied to the program to create the target image). The four examples in Figure~\ref{fig:coding advantage} illustrate each task. All coding questions in EMMA are multiple-choice questions \textbf{created by us from scratch}. Notably, these tasks simulate various real-world applications of MLLMs, requiring skills essential for replicating a target visualization or redesigning an existing one. To ensure familiarity, all visualization code in our benchmark is generated in Python using matplotlib or seaborn. 

Further, similar to the other subjects, we provide fine-grained categorizations for each coding question based on the skills it measures. Since visualizations involve multiple design choices and each problem includes at least four visualizations (or visualization programs), each question may be assigned to multiple categories. On average, each question is associated with 2.11 categories. Through manual coding, we identify a total of ten categories: \textbf{3D} (108 questions; reasoning about visualizations in 3D), \textbf{Color \& Texture} (156 questions; reasoning about the colors and textures of marks), \textbf{Data Reasoning} (108 questions; reasoning about the data in visualizations); \textbf{Advanced Chart Type} (276 questions; involving advanced chart types, such as fishbone diagrams), \textbf{Alignment, Orientation, \& Position} (180 questions; reasoning about how visual elements should be arranged), \textbf{Gridline} (60 questions; reasoning about the use of gridlines), \textbf{Polar Coordinates} (48 questions; reasoning about charts in polar coordinates), \textbf{Axis \& Scale} (108 questions; reasoning about the use of axes and scales), \textbf{Legend} (96 questions; reasoning about the appearance, content, and position of legends), \textbf{Marker \& Line} (48 questions; reasoning about the style of markers and lines). Section~\ref{sec: Case Study} contains sample questions for some of the categories. 

\subsection{Additional Data Curation Details}
We now provide additional details on data curation for chemistry and coding. 

\paragraph{Chemistry} 
We generate ten responses with state-of-the-art LLMs for image-captioned versions of chemistry questions in existing multimodal reasoning datasets~\cite{yue2023mmmu,examsv,scienceqa}. Questions that are answered correctly in at least five out of ten rounds are filtered out. Analysis of the remaining questions shows that most involve molecular formulas, indicating that molecular-related tasks often require additional visual information for effective reasoning.

Based on this observation, we construct the chemistry section in EMMA from scratch, which features three tasks of increasing difficulty. The original data is sourced from SMiCRM~\cite{smicrm}. Notably, the correct answers for the Reaction Simulation task---the most challenging and the one that best reflects vision’s role in the reasoning process---are constructed and verified by a PhD candidate in chemical molecules.  The questions and answers for the Reaction Simulation task are sourced from Li et al.'s collection of chemical reactions~\cite{name_reactions}.

\paragraph{Coding} We manually curate all questions for coding. Our curation process consists of three stages. In the first stage, we identify ``seed visualizations'' that employ advanced visualization techniques or present a rich space for design variations. We source these seed visualizations through three channels: CharXiv~\cite{wang2024charxiv} (a benchmark consisting of diverse charts extracted from arXiv papers), the official matplotlib example gallery~\footnote{\url{https://matplotlib.org/stable/gallery/index.html}. This approach is inspired by Wu et al.~\cite{wu2024plot2code}.}, and our prior experience. For each source, we attempt to reproduce visualizations demonstrating advanced techniques (e.g., 3D bar charts) in Python using GPT-4o or Claude~3.5~Sonnet, retaining only those that MLLMs cannot reasonably replicate after multiple iterations of prompting. 
The first stage ultimately results in 47 seed visualizations.

In the second stage, we generate four variations for each seed visualization. When we prompt MLLMs to reconstruct visualizations during the first stage, MLLM-generated visualizations are often ill-formed, nonsensical, or otherwise fail to achieve the desired effects. However, since these visualizations are generated by MLLMs, they may be indistinguishable from the correct visualizations to the models. Yet, it is crucial for MLLMs to recognize such flaws, as early identification of errors is essential for efficient human-AI collaboration. As such, we include such ``buggy'' code snippets as variations of seed visualizations. In addition, we further enrich the set by introducing design variations (e.g., changes in spine configuration, line style, or axis scaling) either manually or through prompting MLLMs, with post-hoc manual verification. After the second stage, we are left with 188 visualizations, organized into 47 sets, each containing four visualizations.

In the third stage, we construct questions using these 188 visualizations. For Vis Choose Code, we iterate through each visualization within a set and construct questions asking models to select the code snippet used to generate the chart. For Code Choose Vis, we iteratively choose a code snippet from each set and ask models to identify the corresponding generated image. For Modification, we first introduce another design variation in each set, and then select two pairs of visualizations from the set, where each set contains a relatively well-formed chart and another random chart. 
We construct questions by comparing the code of the randomly selected snippet with others, asking what changes are needed to produce the target visualization. While the target visualization image is requisite, we vary whether the original visualization image is provided. In sum, this procedure generates four Code Choose Vis questions, four Vis Choose Code questions, and four Modification questions (two with the original image and two without) per set, resulting in a total of 564 questions evenly divided among the tasks.

\subsection{Comparison with Other Benchmarks}
EMMA stands out from existing multimodal benchmarks by emphasizing questions that truly demand multimodal reasoning capabilities. Through meticulous manual labeling or verification, we provide fine-grained labels for each question, categorizing them based on the specific skills they assess. This approach enables a more detailed analysis of the limitations of MLLMs.

\paragraph{Math}

\begin{figure*}[htbp]
    \centering 
    \includegraphics[width=\textwidth]{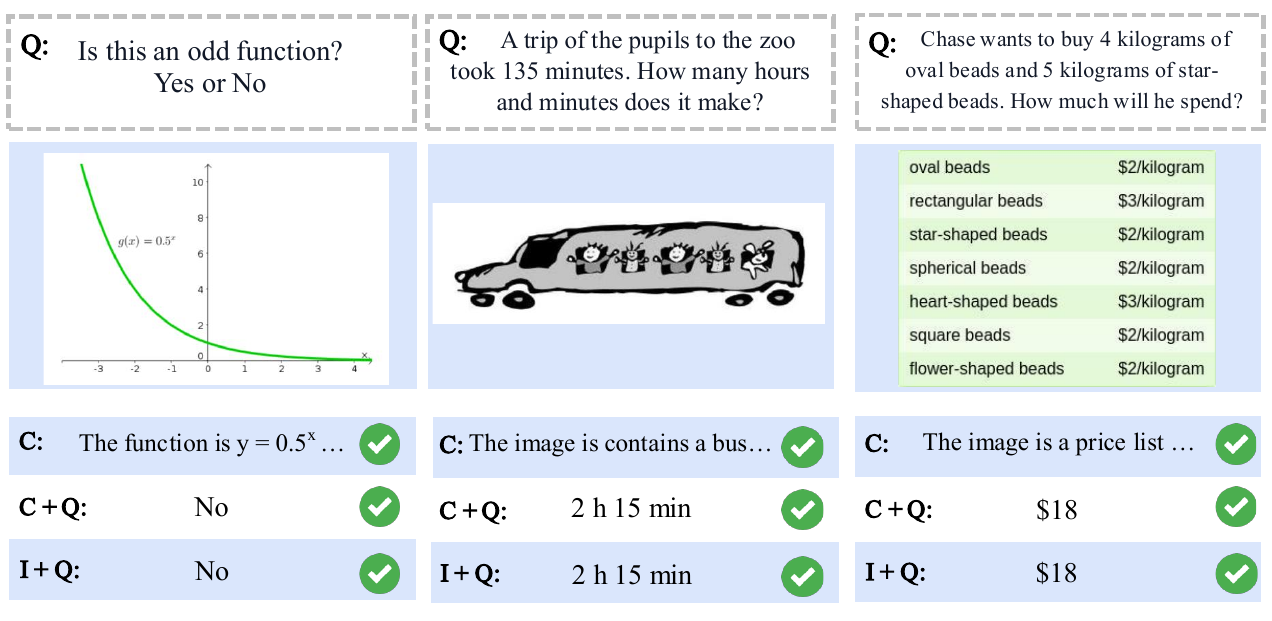}
    \caption{Three typical examples from other math datasets that do not really require the images or can be correctly answered using only the captions of images. The leftmost and rightmost examples are from Math-Vista~\cite{lu2024mathvista} and the middle example is from Math-Vision~\cite{wang2024measuring}. \textbf{Q} represents the question, \textbf{C} represents the caption of the image, and \textbf{I} represents the image. }
    \label{fig: Math-comparison}
\end{figure*}

Various benchmark datasets~\cite{lu2024mathvista, wang2024measuring, yue2023mmmu, yue2024mmmu} have been proposed to evaluate the mathematical reasoning capabilities of MLLMs. However, existing math benchmarks often emphasize shallow perceptual cues or rely heavily on text-dominant reasoning. In contrast, our dataset mainly focuses on assessing the performance of MLLMs on tasks that require integrated reasoning, particularly those that are highly dependent on visual information. Specifically, we employ an enhanced data filtering pipeline to separate questions that could be answered correctly using only the caption of images. Representative examples of such problems are illustrated in Figure~\ref{fig: Math-comparison}. 
In some cases, the images provide no additional information required to solve the question, and the answer can be derived entirely from the text of the question alone, as shown in the middle example in Figure~\ref{fig: Math-comparison}. In other instances, questions can be solved using image captions, leveraging the text reasoning capabilities of MLLMs. These images either consist solely of textual information, as illustrated in the rightmost example in Figure~\ref{fig: Math-comparison}, or can be fully described textually without necessitating further reasoning involving image transformations. For instance, in the leftmost example in Figure~\ref{fig: Math-comparison}, as long as the key textual information in the image, such as $y=0.5^x$, is identified, the question can be solved without employing a graphical approach.

\begin{figure}[h]
    \centering 
    \includegraphics[width=\columnwidth]{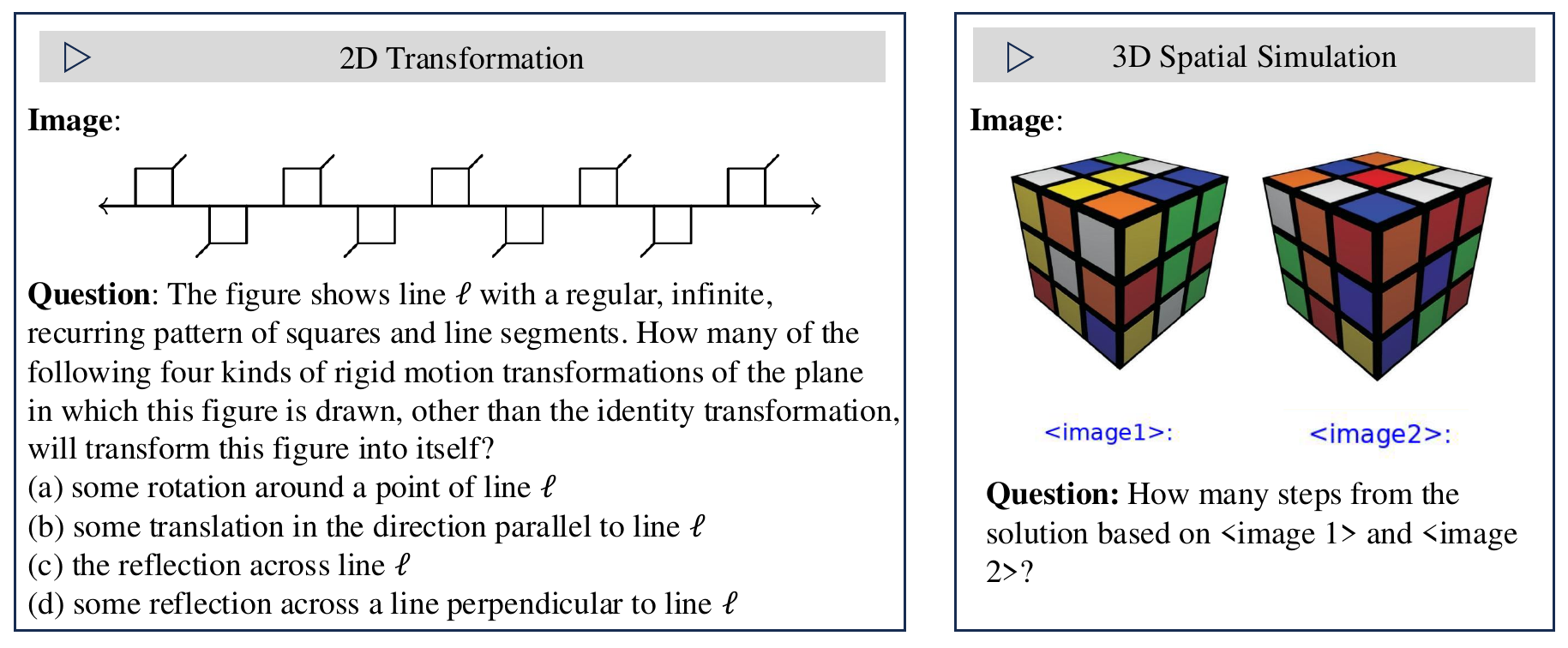}
    \caption{Two examples from MMMU~\cite{yue2023mmmu}, which belong to our defined categories of 2D Transformation and 3D Spatial Simulation respectively. Our classification is equally applicable to other datasets.}
    \label{fig: mmmu}
\end{figure}

In addition to the above, we provide a fine-grained taxonomy for math problems. By first categorizing the questions using GPT-4o and conducting expert-level manual verification of the classifications, we identify categories that are highly likely to require graphic transformation and spatial simulation. These categories are not only applicable to all math problems in our dataset but are also adaptable to other dataset, such as MMMU~\cite{yue2023mmmu}. In figure~\ref{fig: mmmu}, we present two examples from MMMU that fall under our defined categories. We hope these categories will inspire further exploration of the visual reasoning capability of MLLMs.

\paragraph{Science~(Physics \& Chemistry)}
The latest multimodal reasoning benchmarks in science, such as \cite{yue2023mmmu, examsv, scienceqa}, do not provide many multimodal physics and chemistry problems. In addition, they often focus on superficial visual cues or heavily rely on text-based reasoning. As pointed out by \citep{isobench}, text representation can address 90\% of physics and chemistry questions in ScienceQA~\cite{scienceqa}. As a result, our filtering pipeline leaves only 100 problems in total for these subjects from relevant benchmarks, which we expand to 1,332 with our newly constructed problems.

Our benchmark places greater emphasis on the role of vision in multimodal reasoning. Beyond filtering out problems solvable through text alone, we manually review and annotate the remaining questions to ensure a strong reliance on visual information.

In particular, we focus on two specific types of problems. First, some physics problems require visual imagination and simulation of physical processes. Auxiliary images can significantly enhance both the accuracy and efficiency of problem-solving. Second, in chemistry, tasks such as molecule counting, structure recognition, and reaction simulation demand effective utilization of visual information. These two types of problems and their related data have been largely overlooked in current science datasets. To address this gap, we emphasize these under-explored aspects in the science portion of EMMA by manually constructing test questions and carefully sourcing them from existing datasets.

\paragraph{Coding}
Most current visualization-related benchmarks assess visualization understanding~\cite{chartqa, chartx, deplot, plotqa}. In this work, we focus on evaluating how well MLLMs can reason in multimodality when \textit{generating} visualizations. To this end, past work~\cite{chartllama, wu2024plot2code, shi2024chartmimic, zhang2024gpt, novachart} has proposed the task of visualization reproduction---generating code to reproduce a target visualization. To evaluate the quality of generations, researchers have developed heuristic measures and employed MLLMs as judges.

EMMA enhances past work by introducing new task types. Our four tasks enable targeted assessments of \textit{Vis2Code}, \textit{Code2Vis}, and \textit{Visualization Modification}. In particular, while many users rely on MLLMs for visualization debugging, this task has not been addressed by existing benchmarks. We further provide fine-grained, expert-generated categorizations for each question based on the skills it measures (see Section~\ref{coding description}). Finally, all of our coding questions are posed as multiple-choice questions, which removes the need for using MLLMs as judges, which can be unreliable.

\section{Experimental Details}
\label{sec: Experiment Details}

\subsection{Prompts for Data Curation}
As previously discussed, we conduct much filtering to cull out questions from existing datasets that genuinely require visual reasoning, which involves using GPT-4o to generate captions for images and passing the captions along with textual questions to MLLMs to generate responses. The prompts used to generate captions and responses are shown in Table~\ref{tab: prompts for data curation}. Notably, when generating image captions, we also provide the corresponding questions to models to make sure the captions are as accurate as possible.

For math, after filtering the data, we conduct a detailed observation and analysis of the remaining problems and develop a taxonomy consisting of five categories. We then utilize GPT-4o to assist with the categorization and the prompts used during this process are shown in Table~\ref{tab: prompts for data curation}.

\begin{table*}[htbp]
    \centering
    \resizebox{\textwidth}{!}{
    \begin{tabular}{{p{0.2\textwidth} p{0.8\textwidth}}}
        \toprule
        \textbf{Setting} & \textbf{Prompt} \\
        \midrule
        Generate Captions & There is a question about the image or figure. Please describe the fine-grained content of the image or figure based on this question, including scenes, objects, relationships, and any text present. Please note that you do not need to answer this question directly, just describe the information of this picture. \\
        \midrule
        Generate Responses & Please first solve the problem step by step, then put your final answer or a single letter (if it is a multiple choice question) in one ``\textbackslash boxed\{\}''. Here is the natural description of the figure, please solve the following problem based on the description. \\
        \midrule
        Categorize & There are some math problems combining images and text, and existing large models cannot correctly reason through these problems. We have analyzed the reasons why large models fail to classify these problems and have categorized them based on the challenges present in the problems. The categories are as follows: A: To solve the problem, a 2D transformation is required, such as translation, rotation, scaling, shearing, reflection, etc. B: To solve the problem, 3D spatial imagination is needed. C: To solve the problem, path tracing/change of view simulation is needed, such as a math problem about a maze. D: None of the above. It belongs to another category. Here is a math problem, please give the math category that you think this problem belongs to and explain why. If you choose D, please additionally include the type of math problem you believe it to be. \\
        \bottomrule
    \end{tabular}
    }
    \caption{The prompts used to caption images and generate responses during data curation.}
    \label{tab: prompts for data curation}
\end{table*}

\subsection{Prompts for Response Generation}
We evaluate several state-of-the-art MLLMs on EMMA, considering two different prompting strategies: \textit{Direct} prompting and Chain-of-Thought (\textit{CoT}) prompting. Furthermore, our benchmark features two question types: \textit{Open-ended} and \textit{Multiple Choice}. The corresponding prompts vary according to the type of question and the prompting strategy, as shown in Table~\ref{tab:prompts for response generation}

\begin{table*}[htbp]
    \small
    \centering
    \resizebox{\textwidth}{!}{
    \begin{tabular}{{p{0.1\textwidth} p{0.07\textwidth} p{0.66\textwidth}}}
    \toprule
        \textbf{Type} & \textbf{Strategy} & \textbf{Prompt} \\
        \midrule
        \multirow{6}{*}{Open-ended} & \multirow{3}{*}{CoT} & \{context\}\{question\}\\
         & & Answer the question using a single word or phrase and put the answer in one ``\textbackslash boxed\{\}''. Please solve the problem step by step. \\
        \cmidrule{2-3}
        & \multirow{4}{*}{Direct} & \{context\} \{question\}\\
         & & Answer the question using a single word or phrase and put the answer in one ``\textbackslash boxed\{\}''. Please ensure that your output only contains the final answer without any additional content (such as intermediate reasoning steps). \\
        \midrule
        \multirow{7}{*}{\begin{tabular}[l]{@{}l@{}}Multiple \\ Choice\end{tabular}} & \multirow{3}{*}{CoT} & \{context\} \{question\} \{options\}\\
         & & Answer with the option's letter from the given choices and put the letter in one ``\textbackslash boxed\{\}''. Please solve the problem step by step. \\
        \cmidrule{2-3}
         & \multirow{4}{*}{Direct} & \{context\} \{question\} \{options\}\\
         & & Answer with the option's letter from the given choices and put the letter in one ``\textbackslash boxed\{\}''. Please ensure that your output only contains the final answer without any additional content (such as intermediate reasoning steps). \\
        \bottomrule
    \end{tabular}
    }
    \caption{The prompts used for evaluation across different question types and prompting strategies.}
    \label{tab:prompts for response generation}
\end{table*}

\subsection{Models and Settings}
During math data curation, we filter out questions that models can answer when provided only with the image caption and the question. In this process, each model generates ten candidate responses to ensure reliable and effective filtering. To expedite response generation, we use the vLLM~\cite{vllm} library, an open-source tool for fast LLM inference and serving. For all other cases, we load models directly using the Transformers~\cite{transformers} library. All model sources are official and listed in Table~\ref{tab: model parameters}. When evaluating different models, we use default hyperparameter values unless otherwise specified, with detailed parameter settings provided in Table~\ref{tab: model parameters}.

\begin{table*}[htbp]
    \centering
    \resizebox{\textwidth}{!}{
    \begin{tabular}{cccp{0.4\textwidth}}
        \toprule
        \textbf{Model} & \textbf{Parameter Setting} & \textbf{Source} & \textbf{URL} \\
        \midrule
        GPT-4o pass@1 & temperature = 0.0 & chatgpt-4o-latest & \url{https://platform.openai.com}\\
        \midrule
        GPT-4o pass@n & temperature = 0.7 & chatgpt-4o-latest & \url{https://platform.openai.com}\\
        \midrule
        Claude 3.5 Sonnet &  temperature = 0.0 & claude-3-5-sonnet & \url{https://www.anthropic.com/}\\
        \midrule
        \begin{tabular}[c]{@{}c@{}}Gemini 2.0 Flash \\ pass@1\end{tabular} & temperature = 0.0 & gemini-2.0-flash-exp & \url{https://ai.google.dev/} \\
        \midrule
        \begin{tabular}[c]{@{}c@{}}Gemini 2.0 Flash \\ pass@n\end{tabular} & temperature = 0.7 & gemini-2.0-flash-exp & \url{https://ai.google.dev/} \\
        \midrule
        \begin{tabular}[c]{@{}c@{}}Gemini 2.0 Flash \\ Thinking\end{tabular} & temperature = 0.0 & \begin{tabular}[c]{@{}c@{}}gemini-2.0-flash-\\ thinking-exp-1219\end{tabular} & \url{https://ai.google.dev/} \\
        \midrule
        OpenAI o1 & \begin{tabular}[c]{@{}c@{}} - \end{tabular} & interface & \url{https://chatgpt.com/} \\
        \midrule
        Qwen2-VL-72B-Instruct & temperature = 0.7 & local checkpoint & \url{https://huggingface.co/Qwen/Qwen2-VL-72B-Instruct}\\
        \midrule
        LLaVA-Onevision-72B & \begin{tabular}[c]{@{}c@{}}do\underline{~}sample=True,\\ temperature = 0.7\end{tabular} & local checkpoint & \url{https://huggingface.co/llava-hf/llava-onevision-qwen2-72b-ov-hf} \\
        \midrule
        InternVL2-Llama3-76B & \begin{tabular}[c]{@{}c@{}}do\underline{~}sample=True,\\ temperature = 0.7\end{tabular} & local checkpoint & \url{https://huggingface.co/OpenGVLab/InternVL2-Llama3-76B} \\
        \midrule
        InternVL2.5-78B & \begin{tabular}[c]{@{}c@{}}do\underline{~}sample=True,\\ temperature = 0.7\end{tabular} & local checkpoint & \url{https://huggingface.co/OpenGVLab/InternVL2_5-78B} \\
        \bottomrule
    \end{tabular}
    }
    \caption{The sources of models used in the experiments and the hyperparameters configuration. Pass@1 refers to scenarios where evaluation is performed only once, while pass@n refers to cases that require generating multiple candidate responses. }
    \label{tab: model parameters}
\end{table*}

\subsection{Breakdown of Experiment Results by Category}
In this section, we present a detailed breakdown of the results for each category across different subjects. Specifically, the results for math are shown in Table~\ref{tab:math detailed results}, for physics in Table~\ref{tab:physics detailed results}, for chemistry in Table~\ref{tab:chemistry detailed results}, and for coding in Table~\ref{tab:coding detailed results}.

\begin{table*}[htbp]
    \centering
    \small
    \begin{adjustbox}{max width=\textwidth}
    \begin{tabular}{@{}lccccccc@{}}
        \toprule
        \multirow{3}{*}{} & \textbf{EMMA-mini} & \multicolumn{6}{c}{\textbf{EMMA}} \\
        \cmidrule(l){2-2} \cmidrule(l){3-8}
        & \textbf{Overall} & \textbf{2D} & \textbf{3D} & \textbf{Path} & \textbf{MH} & \textbf{Pat} & \textbf{Overall} \\
          & (100) & (266) & (275) & (127) & (124) & (100)  & (892) \\

        \midrule
        Random choice & 13.00 & 15.04 & 12.73 & 10.24 & 22.58 & 9.00 & 14.01  \\
        Human Expert & 75.00 & - & - & - & - & - & - \\
        
        \midrule

        Direct Claude 3.5 Sonnet & 23.00 & 26.69 & 18.18 & 21.26 & 49.19 & 17.00 & 25.34 \\
        Direct Gemini 2.0 Flash & 20.00 & 25.19 & 20.73 & 19.69 & 37.90 & 17.00 & 23.88\\
        Direct GPT-4o & 30.00 & 27.44 & 19.64 & 17.32 & 58.87 & 21.00 & 27.24 \\

        \addlinespace[0.1em]\hdashline\addlinespace[0.1em]
        Direct Qwen2-VL-72B-Instruct & 38.00 & 24.81 & 20.00 & 18.90 & 78.23 & 53.00 & 33.07 \\
        Direct LLaVA-Onevision-72B & 25.00 & 24.81 & 22.18 & 20.47 & 69.35 & 8.00 & 27.69 \\
        Direct InternVL2-Llama3-76B & 31.00 & 22.18 & 14.55 & 22.83 & 65.32 & 15.00 & 25.11 \\
        Direct InternVL2.5-78B & 30.00 & 28.95 & 21.82 & 18.90 & 80.65 & 19.00 & 31.39 \\

        \midrule
        CoT Claude 3.5 Sonnet & 30.00 & 26.69 & 22.18 & 22.83 & 60.48 & 26.00 & 29.37 \\
        CoT Gemini 2.0 Flash & 24.00 & 23.31 & 26.55 & 15.75 & 37.90 & 29.00 & 25.90 \\
        CoT GPT-4o & 27.00 & 23.68 & 17.82 & 14.17 & 60.48 & 23.00 & 25.56 \\
        CoT Gemini 2.0 Flash Thinking & 35.00 & 30.83 & 27.64 & 20.47 & 60.48 & 23.00 & 31.61 \\
        CoT OpenAI o1 & 41.00 & - & - & - & - & - & - \\
        \addlinespace[0.1em]\hdashline\addlinespace[0.1em]
        CoT Qwen2-VL-72B-Instruct & 32.00 & 18.80 & 16.00 & 14.96 & 78.23 & 37.00 & 27.69 \\
        CoT LLaVA-Onevision-72B & 23.00 & 19.17 & 13.45 & 16.54 & 64.52 & 11.00 & 22.42 \\
        CoT InternVL2-Llama3-76B & 27.00 & 16.17 & 14.55 & 15.75 & 64.52 & 15.00 & 22.20 \\
        CoT InternVL2.5-78B & 31.00 & 22.18 & 13.09 & 16.54 & 75.81 & 18.00 & 25.56 \\
        \bottomrule
    \end{tabular}
    \end{adjustbox}
    \caption{Performance of state-of-the-art MLLMs on Math. Column abbreviations: 2D =  2D Transformation, 3D = 3D Spatial Simulation, Path = Path Tracing, Pat =  Pattern Inference, MH = Multi-Hop Visual Object Counting.}
    \label{tab:math detailed results}
\end{table*}

\begin{table*}[htbp]
    \centering
    \small
    \begin{adjustbox}{max width=\textwidth}
    \begin{tabular}{@{}lccccccc@{}}
        \toprule
        \multirow{3}{*}{} & \textbf{EMMA-mini} & \multicolumn{6}{c}{\textbf{EMMA}} \\
        \cmidrule(l){2-2} \cmidrule(l){3-8}
        & \textbf{Overall} & \textbf{Path} & \textbf{3D} & \textbf{MH} & \textbf{VD} & \textbf{GR} & \textbf{Overall} \\
          & (100) & (13) & (37) & (33) & (47) & (26)  & (156) \\

        \midrule
        Random choice & 23.00 & 38.46 & 21.62 & 27.27 & 31.91 & 11.54 & 25.64 \\
        Human Expert & 64.50 & - & - & - & - & - & - \\
        
        \midrule

        Direct Claude 3.5 Sonnet & 34.00 & 30.77 & 37.84 & 36.36 & 31.91 & 30.77 & 33.97 \\
        Direct Gemini 2.0 Flash & 40.00 & 38.46 & 29.73 & 42.42 & 38.30 & 46.15 & 38.46 \\
        Direct GPT-4o & 38.00 & 30.77 & 40.54 & 33.33 & 36.17 & 50.00 & 38.46 \\

        \addlinespace[0.1em]\hdashline\addlinespace[0.1em]
        Direct Qwen2-VL-72B-Instruct & 40.00 & 30.77 & 48.65 & 45.45 & 42.55 & 34.62 & 42.31 \\
        Direct LLaVA-Onevision-72B & 32.00 & 23.08 & 27.03 & 39.39 & 48.94 & 26.92 & 35.90  \\
        Direct InternVL2-Llama3-76B & 22.00 & 15.38 & 32.43 & 21.21 & 6.45 & 23.08 & 22.44 \\
        Direct InternVL2.5-78B & 40.00 & 38.46 & 43.24 & 42.42 & 38.30 & 26.92 & 38.46 \\

        \midrule
        CoT Claude 3.5 Sonnet & 38.00 & 53.85 & 37.84 & 36.36 & 42.55  & 42.31 & 41.03 \\
        CoT Gemini 2.0 Flash & 41.00 & 30.77 & 29.73 & 36.36 & 44.68 & 46.15 & 38.46 \\
        CoT GPT-4o & 44.00 & 69.23 & 35.14 & 39.39 & 44.68 & 46.15 & 43.59 \\
        CoT Gemini 2.0 Flash Thinking & 57.00 & 61.54 & 43.24 & 57.58 & 61.70 & 61.54 & 56.41\\
        CoT OpenAI o1 & 49.00 & - & - & - & - & - & - \\
        \addlinespace[0.1em]\hdashline\addlinespace[0.1em]
        CoT Qwen2-VL-72B-Instruct & 34.00 & 15.38 & 35.14 & 33.33 & 34.04 & 46.15 & 34.62 \\
        CoT LLaVA-Onevision-72B & 26.00 & 15.38 & 10.81 & 18.18 & 6.38 & 15.38 & 12.18 \\
        CoT InternVL2-Llama3-76B & 33.00 & 53.85 & 27.03 & 21.21 & 12.90 & 38.46 & 32.05 \\
        CoT InternVL2.5-78B & 36.00 & 46.15 & 35.14 & 30.30 & 46.81 & 42.31 & 39.74 \\
        \bottomrule
    \end{tabular}
    \end{adjustbox}
    \caption{Performance of state-of-the-art MLLMs on Physics. Column abbreviations: Path = Path Tracing, 3D = 3D Field Simulation, MH = Multi-Hop Visual Reasoning, VD = Visual Decomposition Simulation, GR = Graph Reasoning. }
    \label{tab:physics detailed results}
\end{table*}

\begin{table*}[htbp]
    \centering
    \small
    \begin{adjustbox}{max width=\textwidth}
    \begin{tabular}{@{}lccccccc@{}}
        \toprule

         \multirow{3}{*}{} & \textbf{EMMA-mini} & \multicolumn{6}{c}{\textbf{EMMA}} \\
        \cmidrule(l){2-2} \cmidrule(l){3-8}
        & \textbf{Overall} & \textbf{SR} & \textbf{GR} & \textbf{RS} & \textbf{RS-pro} & \textbf{KS} & \textbf{Overall} \\
          & (100) & (474) & (9) & (132) & (105) & (456)  & (1176) \\
          
        \midrule
        Random choice & 27.00 & 24.47 & 33.33 & 27.27 & 35.24 & 0.44 & 16.50 \\
        Human Expert & 86.00 & - & - & - & - & - & -  \\
        
        \midrule

        Direct Claude 3.5 Sonnet & 44.00 & 66.88 & 22.22 & 55.30 & 55.24 & 6.80 & 40.90 \\
        Direct Gemini 2.0 Flash & 36.00 & 54.01 & 11.11 & 53.79 & 58.10 & 8.33 & 36.31 \\
        Direct GPT-4o & 33.00 & 47.05 & 11.11 & 51.52 & 42.86 & 8.33 & 31.89 \\

        \addlinespace[0.1em]\hdashline\addlinespace[0.1em]
        Direct Qwen2-VL-72B-Instruct & 34.00 & 45.99 & 33.33 & 48.48 & 45.71 & 9.65 & 32.06 \\
        Direct LLaVA-Onevision-72B & 24.00 & 38.19 & 33.33 & 39.39 & 26.67 & 7.24 & 25.26  \\
        Direct InternVL2-Llama3-76B & 21.00 & 37.34 & 22.22 & 31.45 & 24.76 & 8.55 & 24.06 \\
        Direct InternVL2.5-78B & 38.00 & 55.06 & 33.33 & 47.73 & 43.81 & 8.99 & 35.20  \\

        \midrule
        CoT Claude 3.5 Sonnet & 41.00 & 57.17 & 33.33 & 58.33 & 58.10 & 15.57 & 41.07 \\
        CoT Gemini 2.0 Flash & 36.00 & 22.15 & 33.33 & 50.00 & 59.05 & 11.84 & 24.66 \\
        CoT GPT-4o & 35.00 & 42.41 & 33.33 & 51.52 & 45.71 & 16.67 & 33.67  \\
        CoT Gemini 2.0 Flash Thinking & 41.00 & 48.31 & 33.33 & 45.45 & 69.52 & 17.76 & 37.93\\
        CoT OpenAI o1 & 40.00 & - & - & - & - & - & -  \\
        \addlinespace[0.1em]\hdashline\addlinespace[0.1em]
        CoT Qwen2-VL-72B-Instruct & 32.00 & 33.33 & 11.11 & 37.12 & 42.86 & 7.89 & 24.57 \\
        CoT LLaVA-Onevision-72B & 23.00 & 33.76 & 0.00 & 37.88 & 20.95 & 7.24 & 22.53 \\
        CoT InternVL2-Llama3-76B & 21.00 & 29.11 & 22.22 & 30.65 & 22.86 & 6.58 & 19.73  \\
        CoT InternVL2.5-78B & 24.00 & 37.13 & 33.33 & 37.12 & 33.33 & 13.16 & 27.47 \\
        \bottomrule
    \end{tabular}
    \end{adjustbox}
    \caption{Performance of state-of-the-art MLLMs on Chemistry. Column abbreviations: SR = Structure Recognition, GR = Graph Reasoning, RS = Reaction Simulation, RS-pro = Reaction Simulation-Pro, KS = Knowledge-based Counting. }
    \label{tab:chemistry detailed results}
\end{table*}

\begin{table*}[htbp]
    \centering
    \small
    \begin{adjustbox}{max width=\textwidth}
    \begin{tabular}{@{}lcccccc@{}}
        \toprule
        \multirow{3}{*}{} & \textbf{EMMA-mini} & \multicolumn{5}{c}{\textbf{EMMA}} \\
        \cmidrule(l){2-2} \cmidrule(l){3-7}
        & \textbf{Overall} & \textbf{CCV} & \textbf{VCC} & \textbf{MwoI} & \textbf{MwI} & \textbf{Overall} \\
          & (100) & (188) & (188) & (94) & (94) & (564)  \\

        \midrule
        Random choice & 28.00 & 22.87 & 23.94 & 29.79 & 30.85 & 25.71 \\
        Human Expert & 85.50 & - & - & - & - & - \\
        
        \midrule

        Direct Claude 3.5 Sonnet & 35.00 & 32.98 & 41.49 & 40.43 & 42.55 & 38.65  \\
        Direct Gemini 2.0 Flash & 41.00 & 39.36 & 42.02 & 43.62 & 45.74 & 42.02  \\
        Direct GPT-4o & 40.00 & 43.09 & 35.11 & 40.43 & 43.62 & 40.07  \\

        \addlinespace[0.1em]\hdashline\addlinespace[0.1em]
        Direct Qwen2-VL-72B-Instruct & 37.00 & 35.11 & 30.85 & 36.17 & 39.36 & 34.57   \\
        Direct LLaVA-Onevision-72B & 28.00 & 22.34 & 28.19 & 38.30 & 32.98 & 28.72   \\
        Direct InternVL2-Llama3-76B & 28.00 & 27.66 & 39.52  & 30.85 & 28.72 & 27.84   \\
        Direct InternVL2.5-78B & 33.00 & 30.85 & 31.38 & 35.11 & 31.91 & 31.91   \\

        \midrule
        CoT Claude 3.5 Sonnet & 39.00 & 39.36 & 38.83 & 43.62 & 43.62 & 40.60  \\
        CoT Gemini 2.0 Flash & 44.00 & 38.30 & 46.28 & 37.23 & 39.36 & 40.96   \\
        CoT GPT-4o & 38.00 & 40.43 & 35.11 & 44.68 & 38.30 & 39.01   \\
        CoT Gemini 2.0 Flash Thinking & 41.00 & 43.62 & 46.81 & 39.36 & 40.43 & 43.44   \\
        CoT OpenAI o1 & 53.00 & - & - & - & - & -   \\
        \addlinespace[0.1em]\hdashline\addlinespace[0.1em]
        CoT Qwen2-VL-72B-Instruct & 23.00 & 31.38 & 28.72 & 25.53 & 30.85 & 29.43   \\
        CoT LLaVA-Onevision-72B & 29.00 & 24.47 & 32.98 & 35.11 & 34.04 & 30.67   \\
        CoT InternVL2-Llama3-76B & 32.00 & 24.47 & 29.79 & 38.30 & 35.11 & 30.32  \\
        CoT InternVL2.5-78B & 19.00 & 25.53 & 25.00 & 25.53 & 24.47 & 25.18  \\
        \bottomrule
    \end{tabular}
    \end{adjustbox}
    \caption{Performance of state-of-the-art MLLMs on Coding. Column abbreviations: CCV = Code Choose Vis, VCC = Vis Choose Code, MwoI = Modify without Original Image, MwI = Modify with Original Image. }
    \label{tab:coding detailed results}
\end{table*}

\begin{table*}[t]
    \centering
    \small
    \begin{adjustbox}{max width=\textwidth}
    \begin{tabular}{ccccccc}
    \toprule
    \textbf{Model} & \textbf{Method} & \textbf{N=1} & \textbf{N=2} & \textbf{N=4} & \textbf{N=8} & \textbf{N=16} \\
    \midrule
    \multirow[c]{3}{*}{GPT-4o} & BoN w. Self-RM & \multirow{3}{*}{27.00} & 29.00 & 27.00 & 25.00 & $-$ \\
     & BoN w. Gemini 2.0 Flash Thinking as RM &  & \second{30.00} & 28.00 & \best{31.00} & $-$ \\
     & BoN w. Qwen2.5-Math-RM-72B as RM &  & 26.00 & 24.00 & 25.00 & 29.00  \\
    \midrule
    \multirow[c]{3}{*}{Gemini 2.0 Flash} & BoN w. Self-RM & \multirow{3}{*}{24.00} & \best{33.00} & 24.00 & 27.00 & $-$ \\
     & BoN w. Gemini 2.0 Flash Thinking as RM &  & \best{33.00} & 24.00 & 25.00 & $-$ \\
     & BoN w. Qwen2.5-Math-RM-72B as RM &  & 27.00 & 28.00 & 23.00 & 23.00  \\
    \midrule
    Gemini 2.0 Flash Thinking & $-$ & \best{35.00} & $-$ & $-$ & $-$ & $-$  \\
    \midrule
    o1 & $-$ & \best{41.00} & $-$ & $-$ & $-$ & $-$ \\
    \bottomrule
    \end{tabular}
    \end{adjustbox}
    \caption{Test-time scaling results on the math portion of EMMA-mini using Qwen2.5-Math-RM-72B, a specialized reward model for math, and some generalist models as reward models. Since Qwen2.5-Math-RM-72B does not take images as input, we provide GPT-4o-generated captions of images in the problems to it. Overall, generalist reward models provide better reward signals than Qwen2.5-Math-RM-72B. 
    } 
    \label{tab:math scaling}
\end{table*}

\subsection{Best-of-N With a Specialized Math Reward Model} 
While we want to evaluate specialized reward models in addition to generalist reward models, they are currently only available for math. Qwen2.5-Math-RM-72B~\cite{yang2024qwen2} is one such specialized LLM for evaluating the quality of responses to math problems. For the math portion of EMMA-mini, we first generate 16 responses using each of GPT-4o and Gemini 2.0 Flash. We then score each response using Qwen2.5-Math-RM-72B and select the highest-scoring answer. As in previous experiments, we test with \textit{N} values of 2, 4, 8, and 16. Since Qwen2.5-Math-RM-72B is not multimodal, we supply GPT-4o-generated captions for all images in the questions to enable evaluation. 

Table~\ref{tab:math scaling} compares scaling results on the math portion of EMMA-mini using Qwen2.5-Math-RM-72B, a specialized reward model for math, against other generalist reward models. Overall, generalist reward models generate better rewards than Qwen2.5-Math-RM-72B. We note, however, that Qwen2.5-Math-RM-72B is not a text-only model, which likely affects its performance on a multimodal benchmark like EMMA.

\section{Case Study}
\label{sec:case}
We now present additional case studies, showcasing both correct and incorrect responses by MLLMs, organized by subject. We provide the original questions, MLLM responses, ground truth solutions, and our error analyses. Some questions also feature o1 responses to the captioned version of the original questions. Overall, we find that a recurring pattern in error cases is that MLLMs fail to fully engage their multimodal reasoning skills. When questions require multiple visual passes or extended visual simulation or manipulation, MLLMs often skip over these critical steps in their thought processes or produce completely incorrect solutions.

\label{sec: Case Study}

\begin{figure*}[htbp]
    \centering 
    \includegraphics[width=\textwidth]{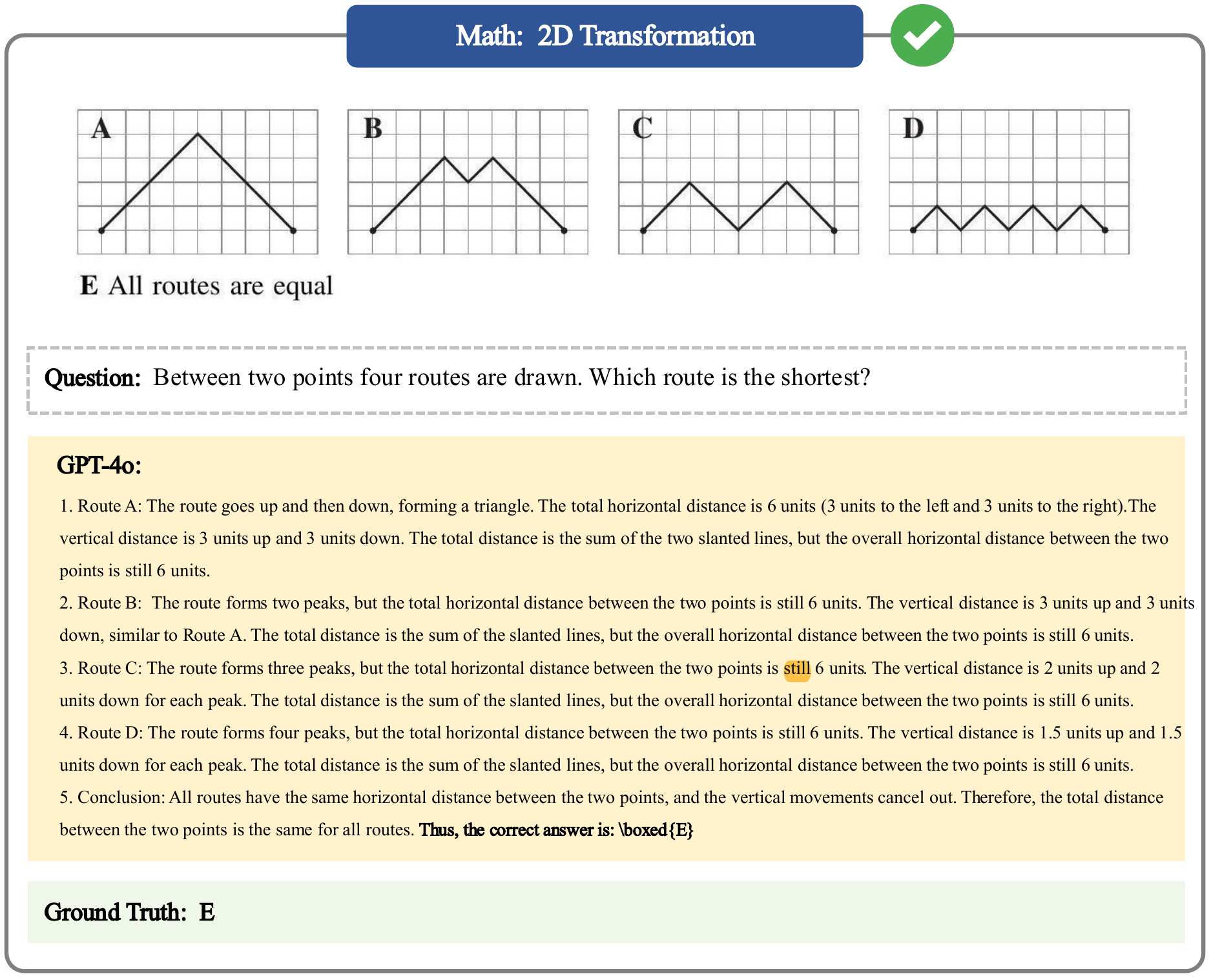}
    \caption{A sample correct case in math. }
    \label{fig: math-case-correct-1}
\end{figure*}

\begin{figure*}[htbp]
    \centering 
    \includegraphics[width=\textwidth]{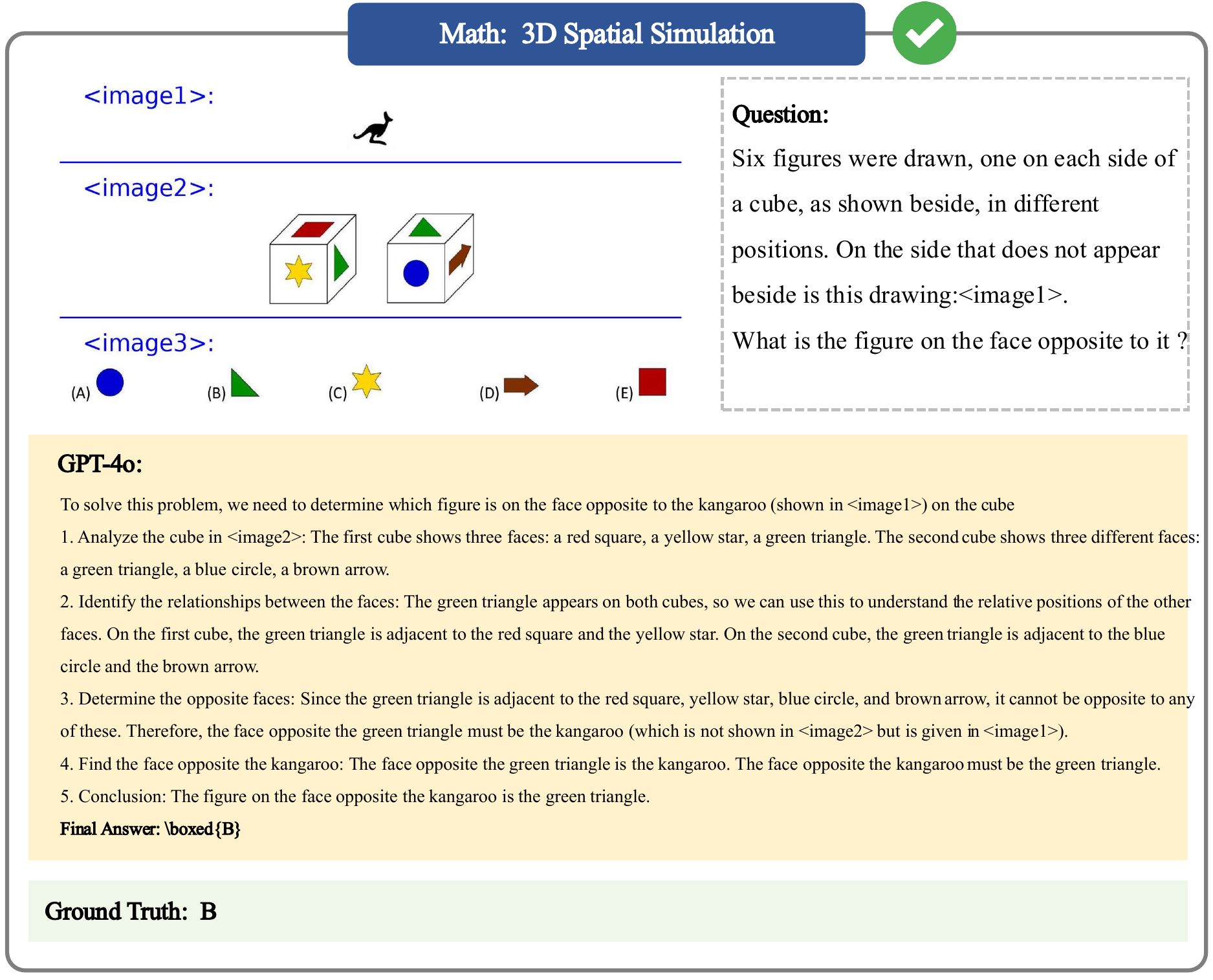}
    \caption{A sample correct case in math. }
    \label{fig: math-case-correct-2}
\end{figure*}

\begin{figure*}[htbp]
    \centering 
    \includegraphics[width=\textwidth]{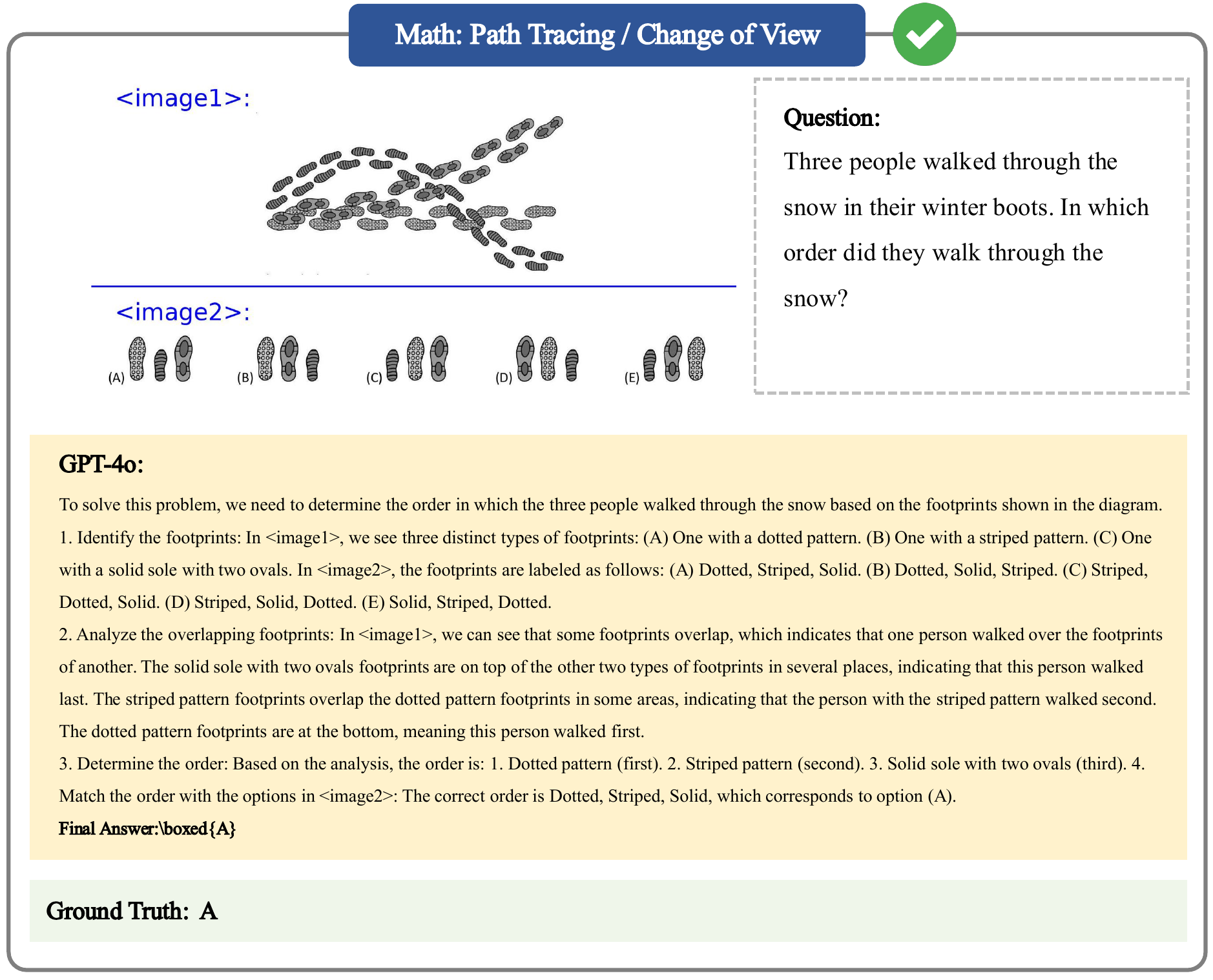}
    \caption{A sample correct case in math. }
    \label{fig: math-case-correct-3}
\end{figure*}

\begin{figure*}[htbp]
    \centering 
    \includegraphics[width=\textwidth]{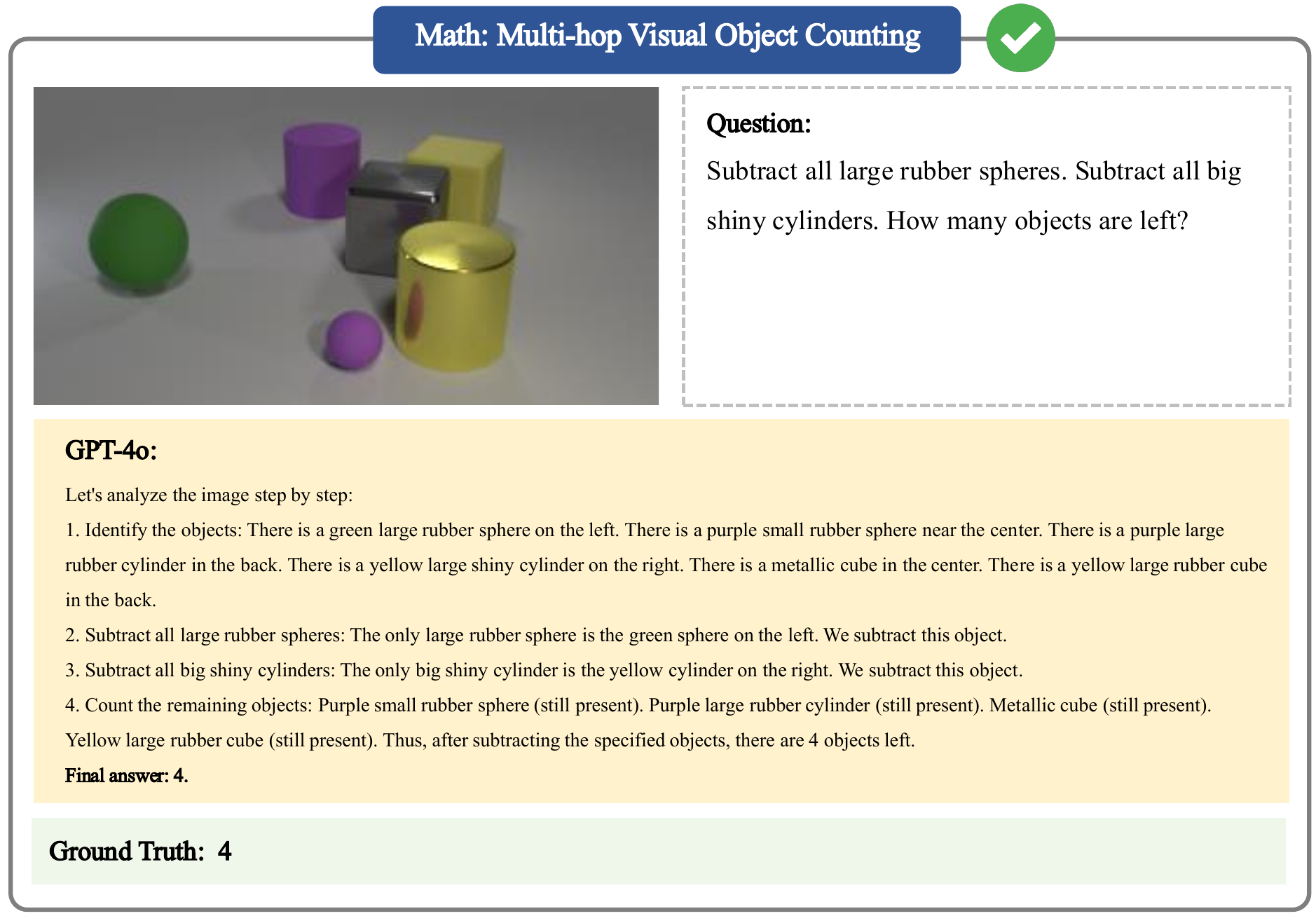}
    \caption{A sample correct case in math. }
    \label{fig: math-case-correct-4}
\end{figure*}

\begin{figure*}[htbp]
    \centering 
    \includegraphics[width=\textwidth]{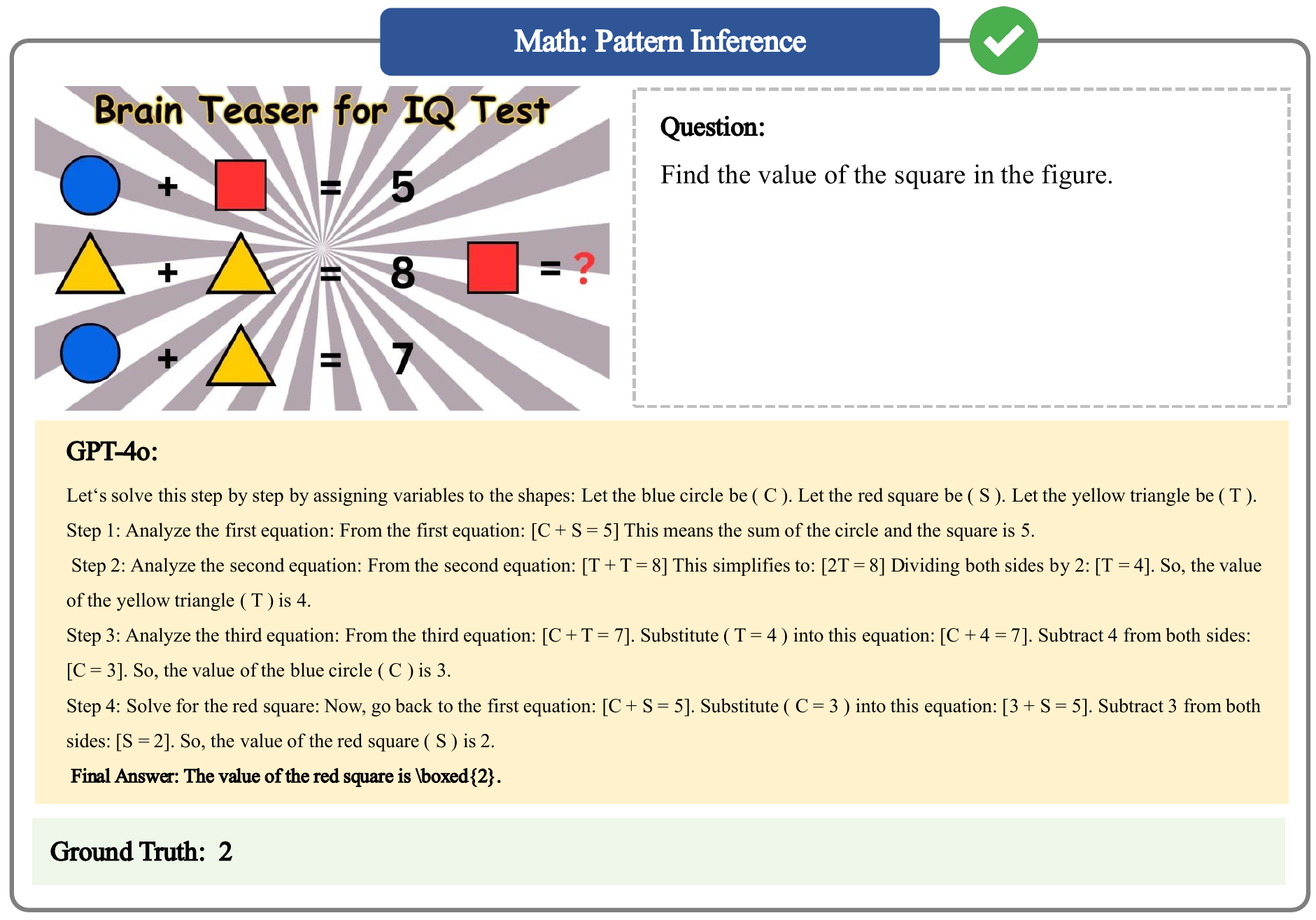}
    \caption{A sample correct case in math. }
    \label{fig: math-case-correct-5}
\end{figure*}

\begin{figure*}[h]
    \centering 
    \includegraphics[height=0.8\textheight]{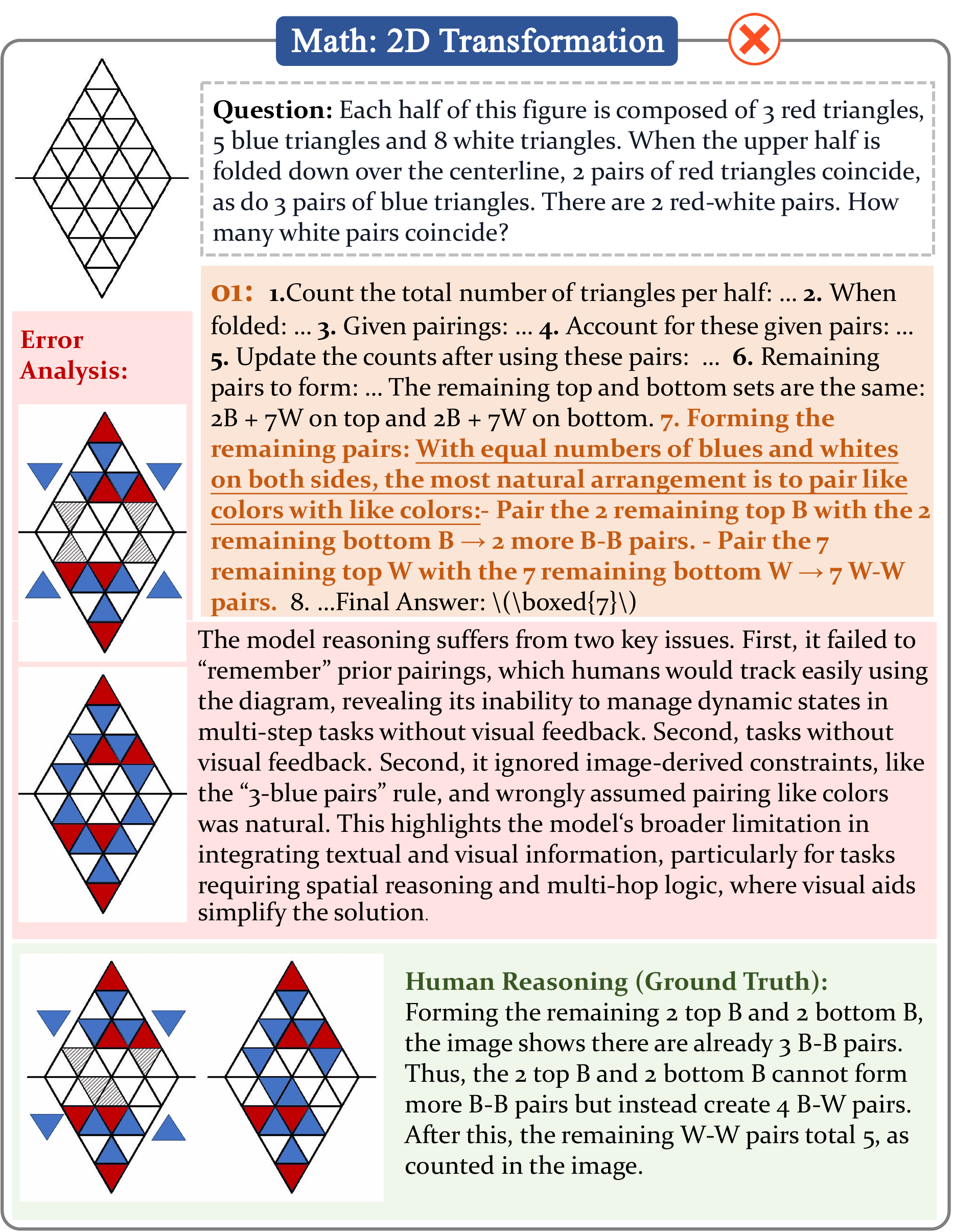}
    \caption{A sample error case in math. }
    \label{fig: math-case-wrong-4}
\end{figure*}

\begin{figure*}[h]
    \centering 
    \includegraphics[width=\textwidth]{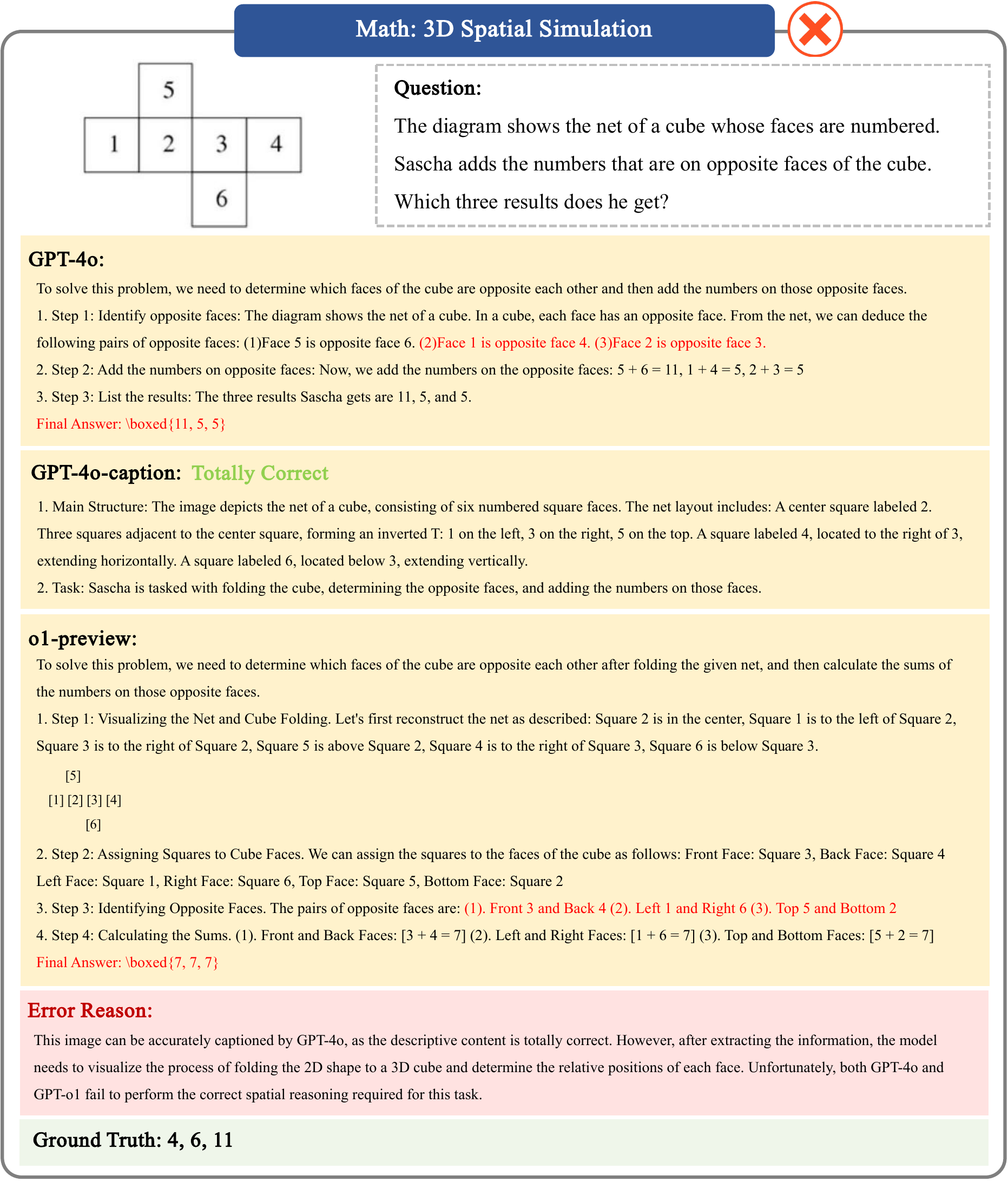}
    \caption{A sample error case in math. }
    \label{fig: math-case-wrong-1}
\end{figure*}

\begin{figure*}[h]
    \centering 
    \includegraphics[height=\textheight]{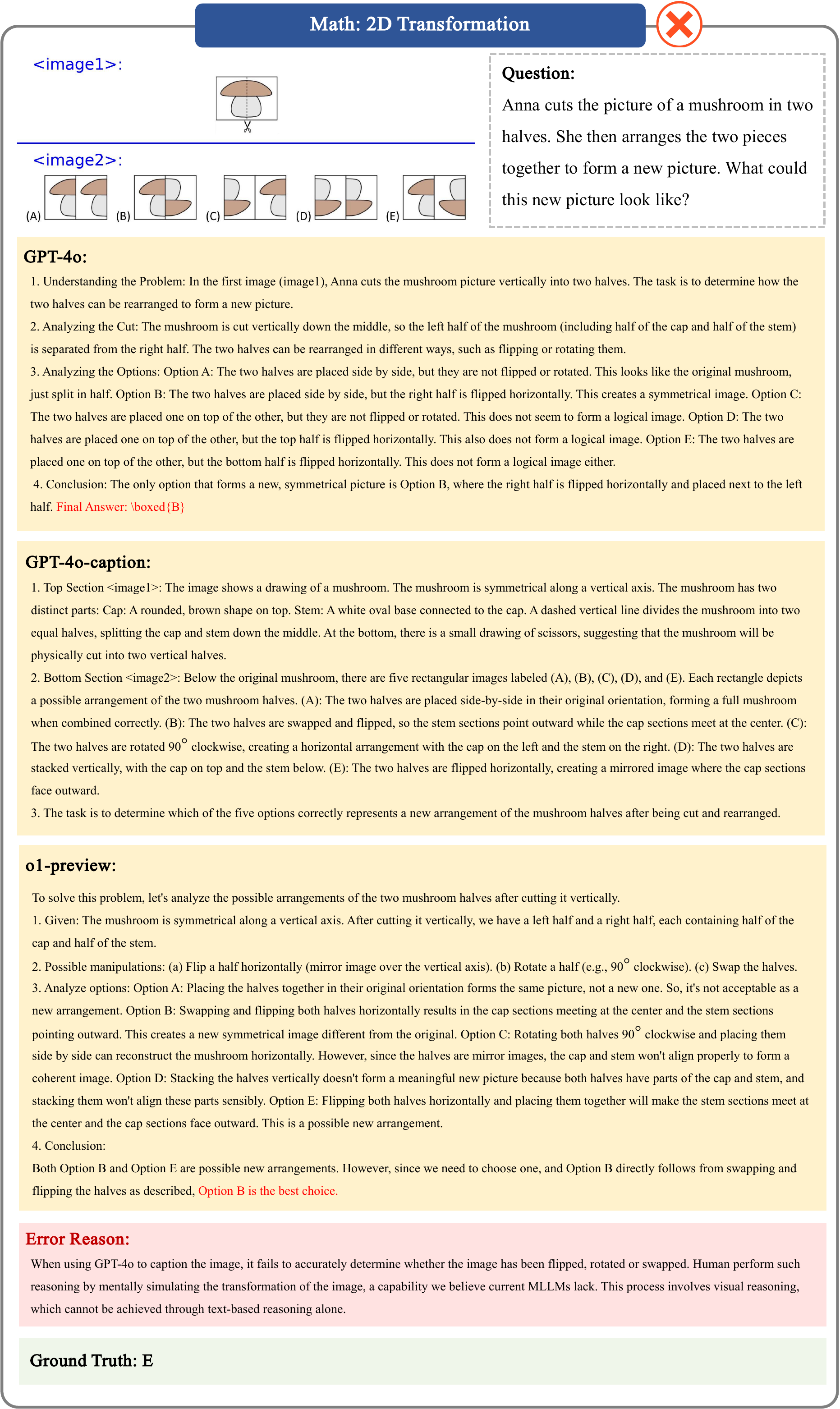}
    \caption{A sample error case in math. }
    \label{fig: math-case-wrong-2}
\end{figure*}

\begin{figure*}[h]
    \centering 
    \includegraphics[height=\textheight]{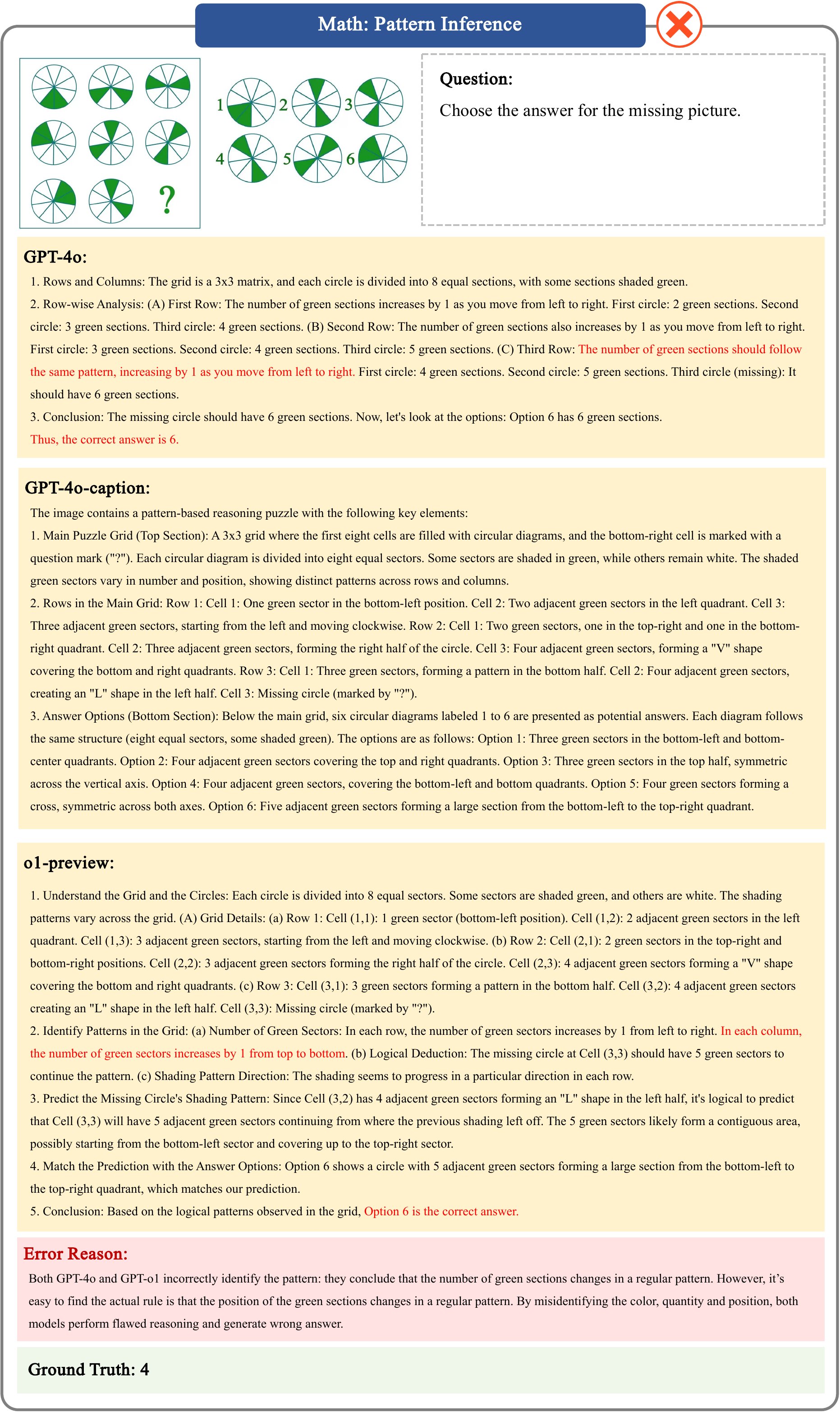}
    \caption{A sample error case in math. }
    \label{fig: math-case-wrong-3}
\end{figure*}


\begin{figure*}[h]
    \centering 
    \includegraphics[width=\textwidth]{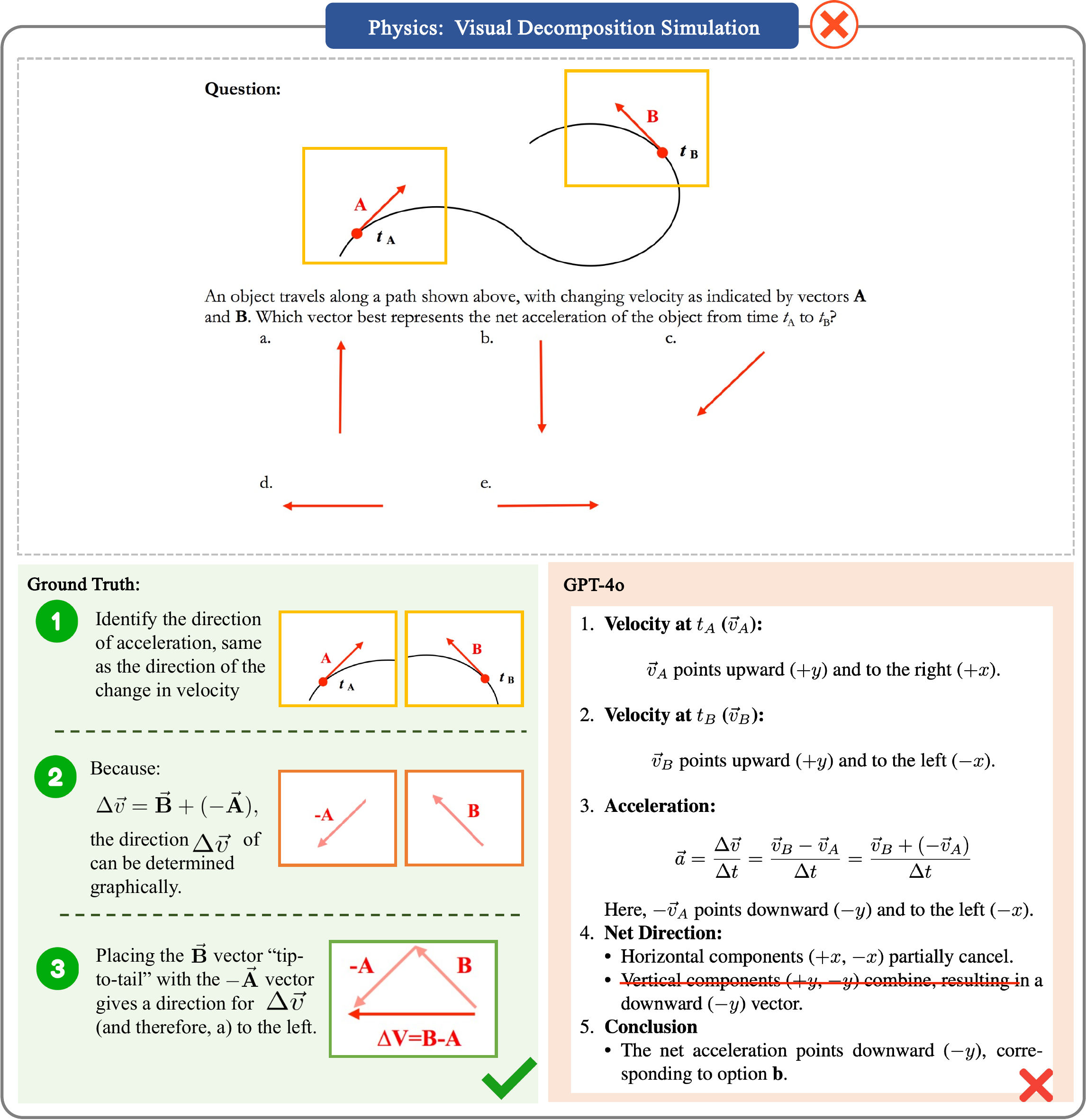}
    \caption{A sample error case in physics.}
    \label{fig:physics_wrong_4}
\end{figure*}

\begin{figure*}[h]
    \centering 
    \includegraphics[width=0.8\textwidth]{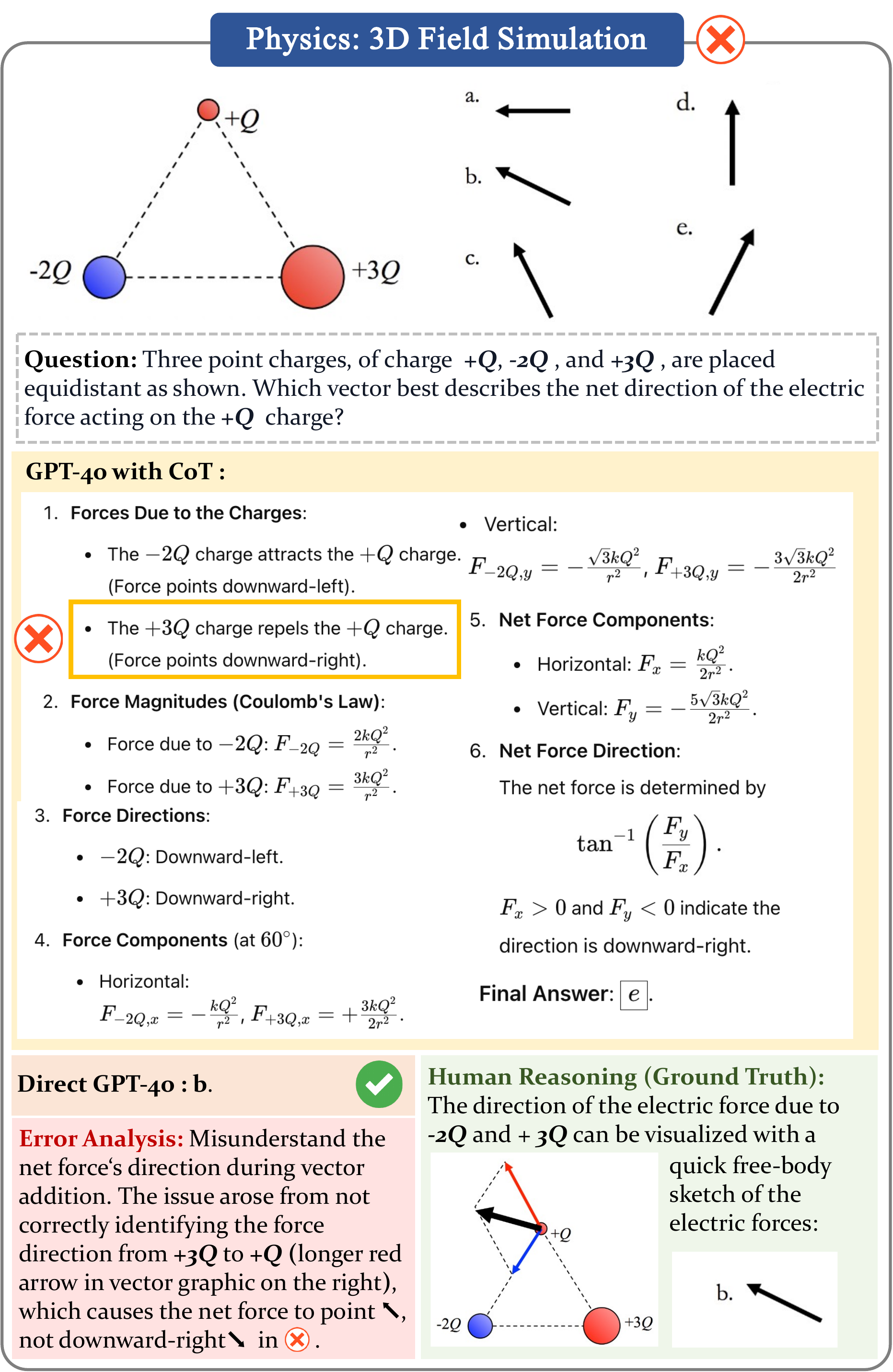}
    \caption{A sample error case in physics.}
    \label{fig:physics_wrong_5}
\end{figure*}

\begin{figure*}[h]
    \centering 
    \includegraphics[width=\textwidth]{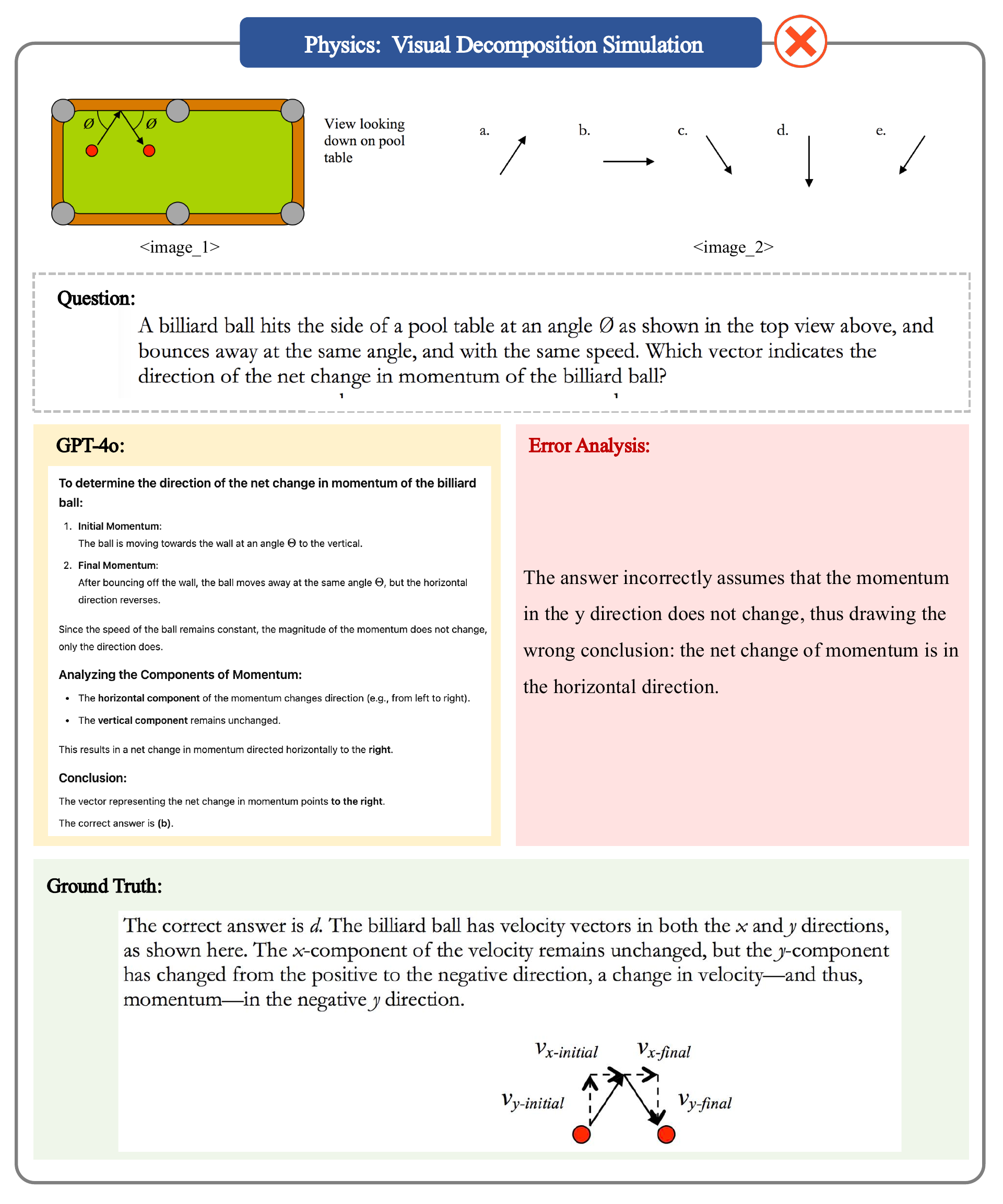}
    \caption{A sample error case in physics.}
    \label{fig:physics_wrong_1}
\end{figure*}

\begin{figure*}[h]
    \centering 
    \includegraphics[width=\textwidth]{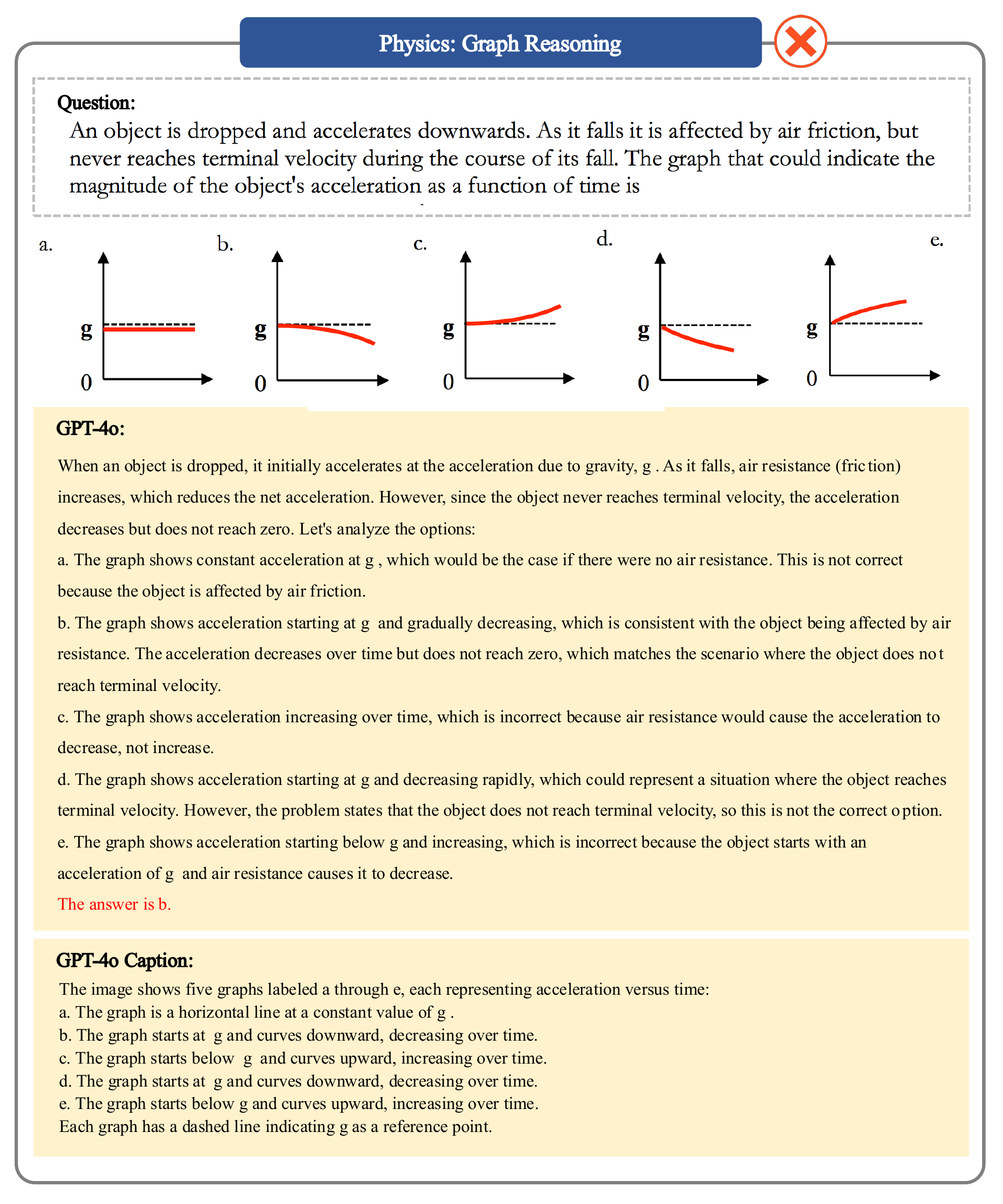}
    \caption{A sample error case in physics.}
    \label{fig:physics_wrong_2}
\end{figure*}

\begin{figure*}[h]
    \centering 
    \includegraphics[width=\textwidth]{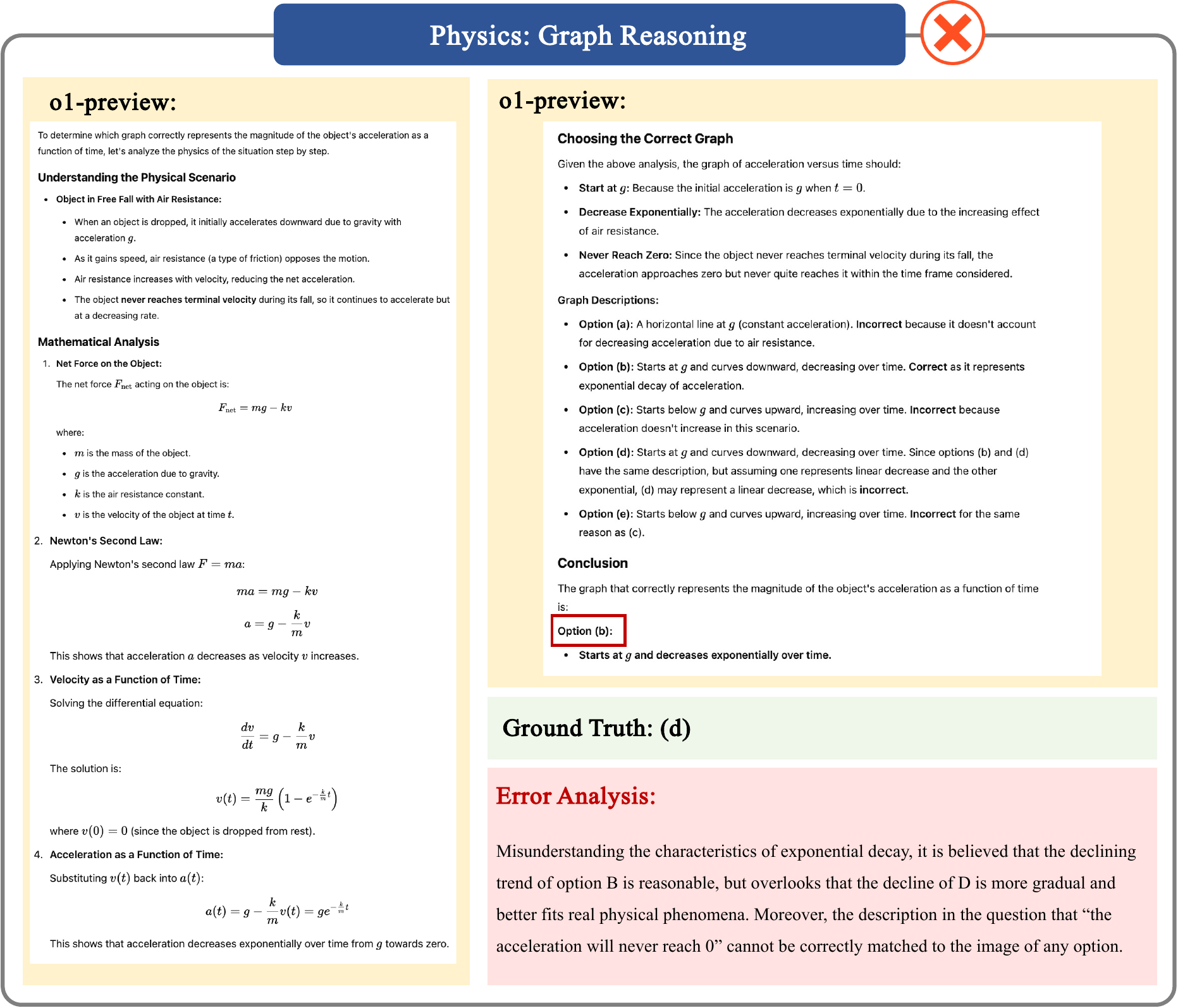}
    \caption{A sample error case in physics.}
    \label{fig:physics_wrong_3}
\end{figure*}

\begin{figure*}[h]
    \centering 
    \includegraphics[width=\textwidth]{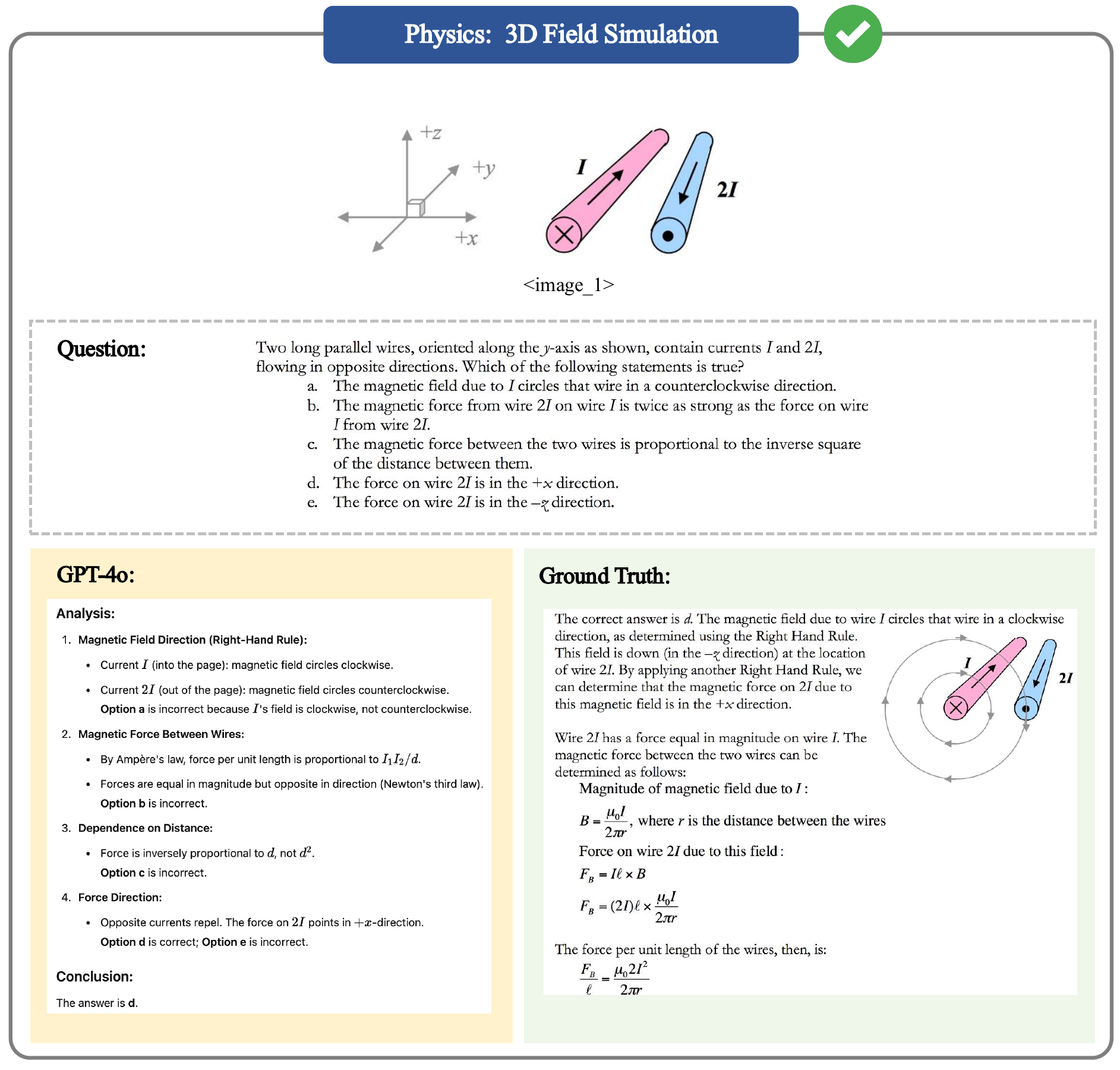}
    \caption{A sample correct case in physics.}
    \label{fig:physics_correct_1}
\end{figure*}

\begin{figure*}[h]
    \centering 
    \includegraphics[width=\textwidth]{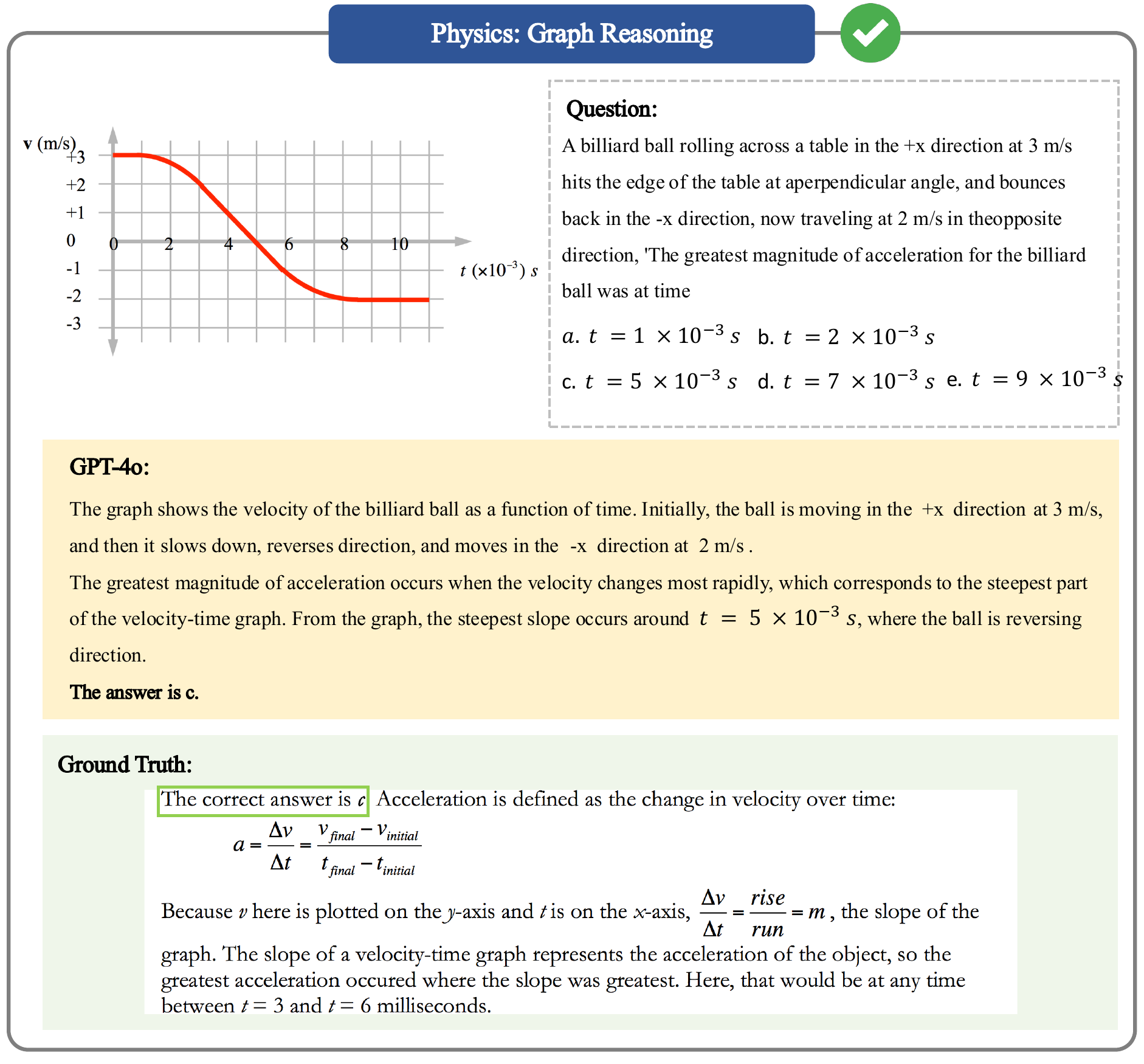}
    \caption{A sample correct case in physics.}
    \label{fig:physics_correct_2}
\end{figure*}


\begin{figure*}[h]
    \centering 
    \includegraphics[width=\textwidth]{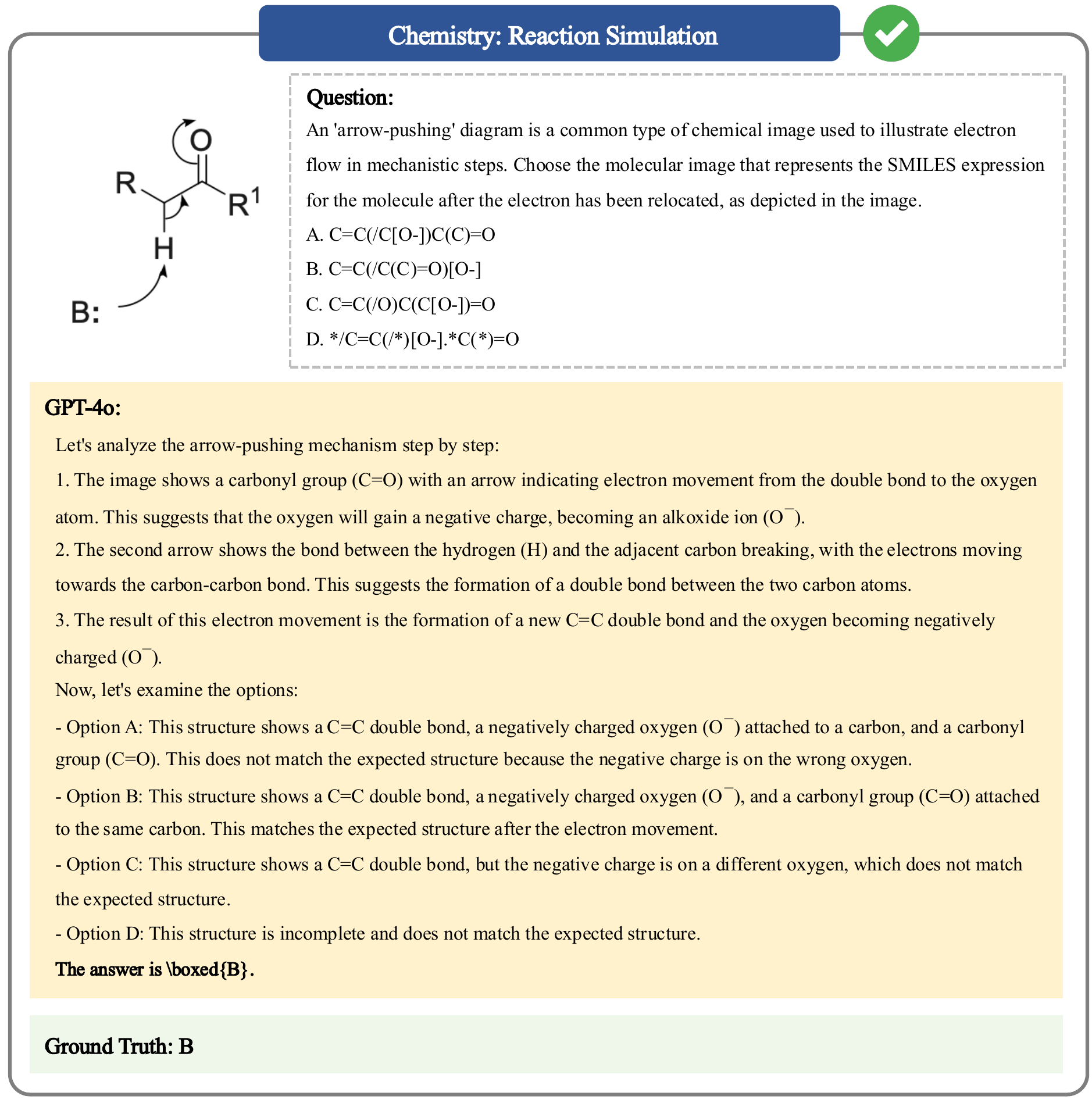}
    \caption{A sample correct case in chemistry.}
    \label{fig:chemistry_correct_1}
\end{figure*}

\begin{figure*}[h]
    \centering 
    \includegraphics[width=\textwidth]{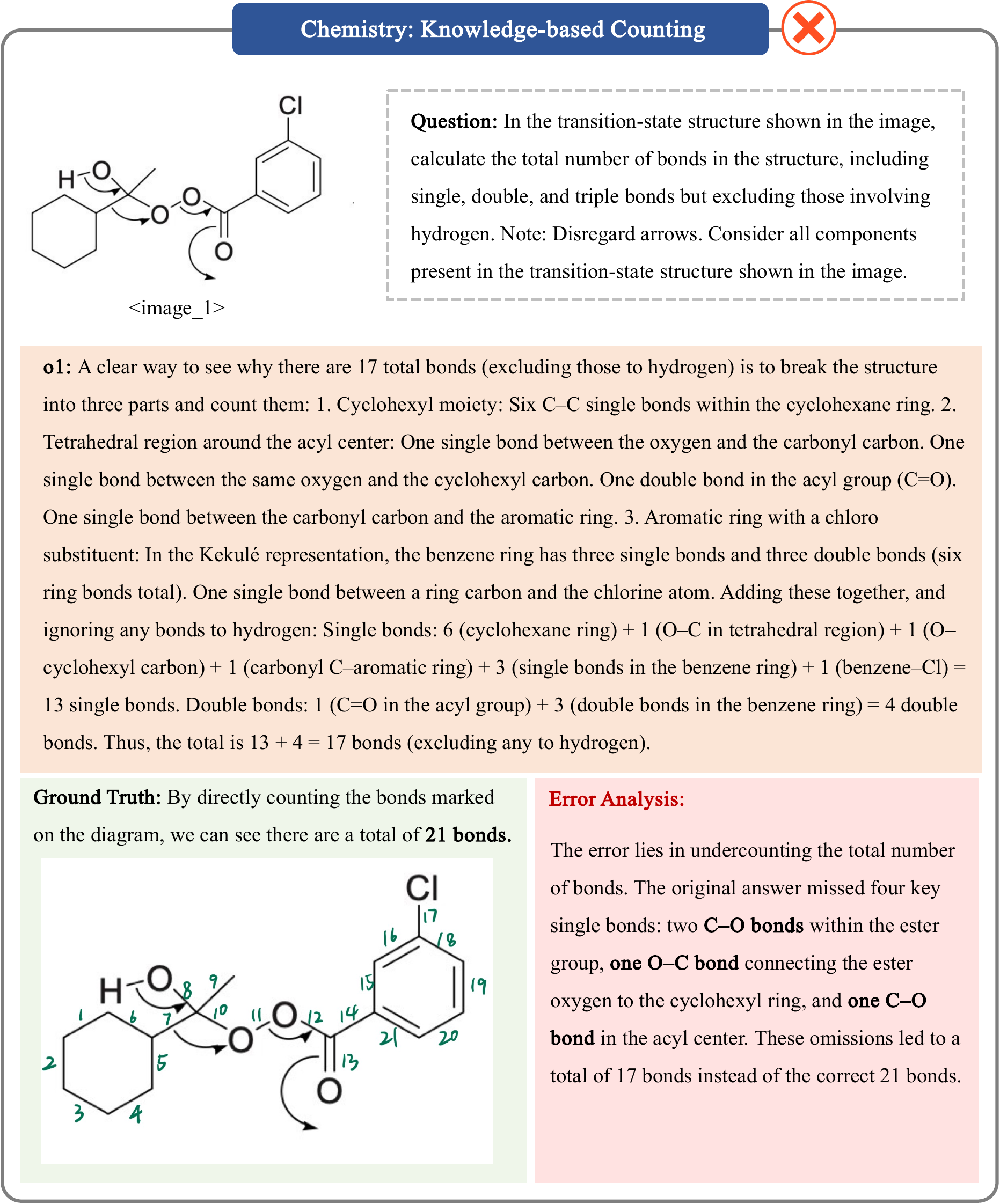}
    \caption{A sample error case in chemistry.}
    \label{fig:chemistry_wrong_3}
\end{figure*}


\begin{figure*}[h]
    \centering 
    \includegraphics[height=\textheight]{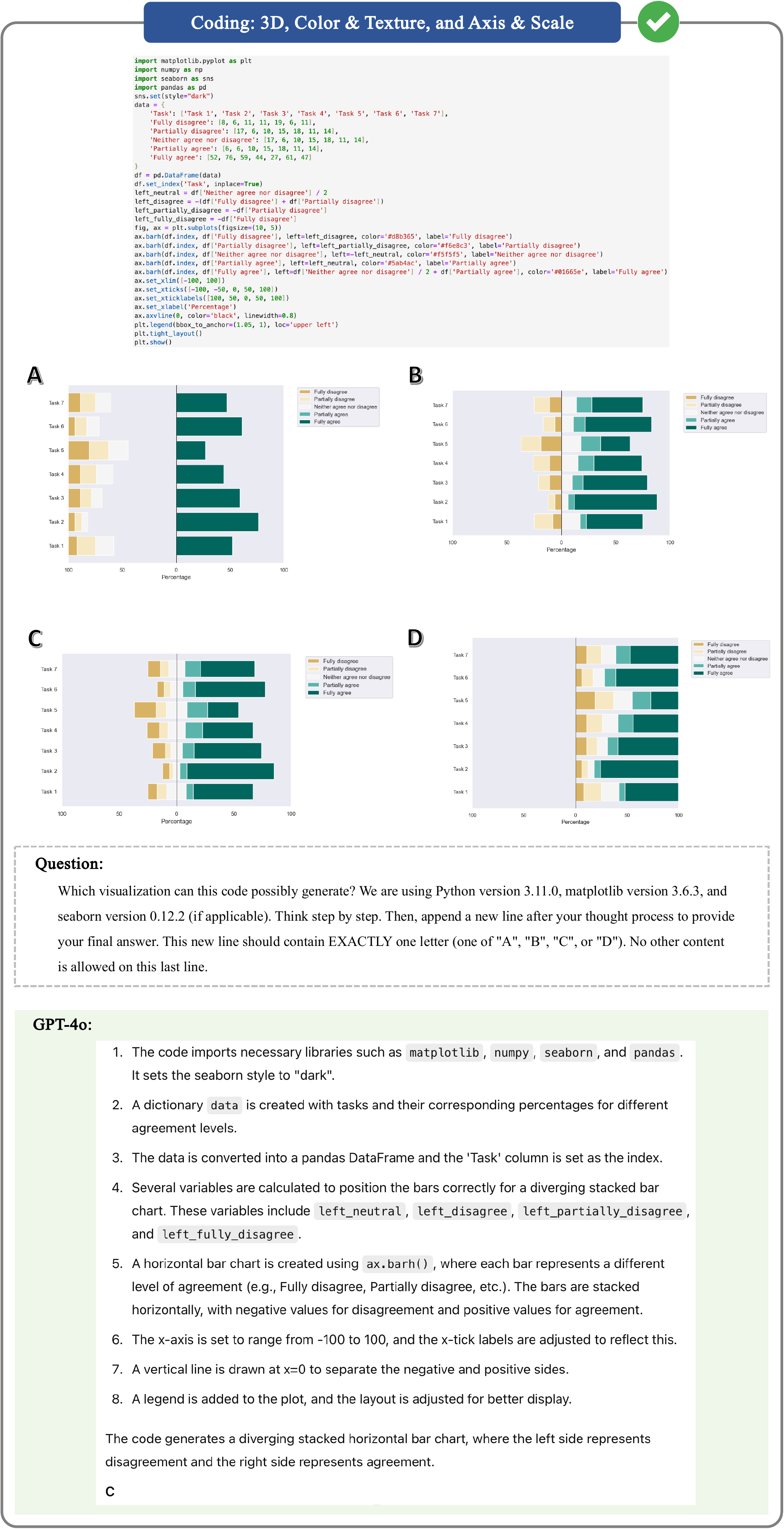}
    \caption{A sample correct case in coding.}
    \label{fig:coding-correct}
\end{figure*}

\begin{figure*}[h]
    \centering 
    \includegraphics[height=\textheight]{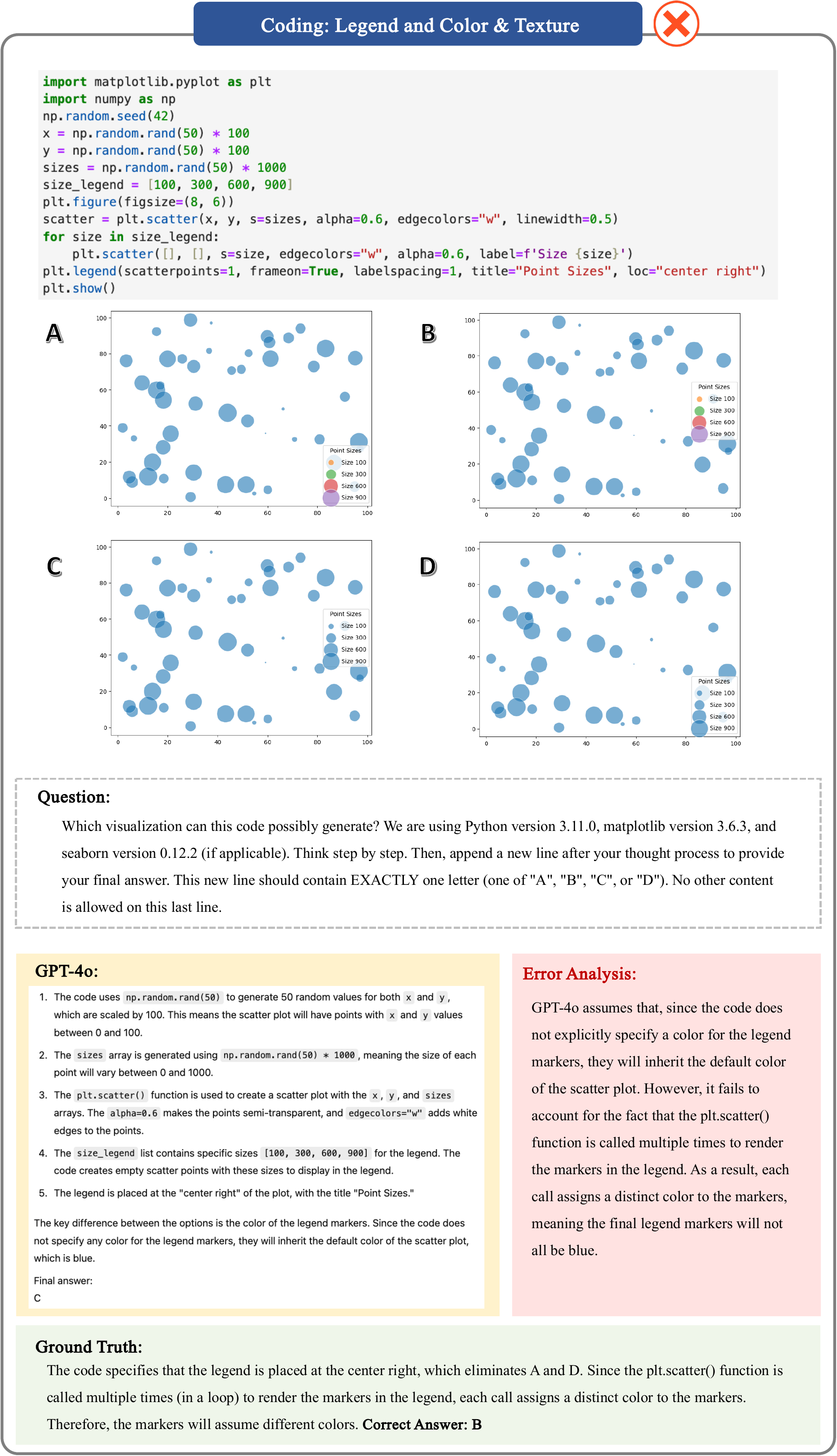}
    \caption{A sample error case in coding.}
    \label{fig:coding-case-wrong-1}
\end{figure*}

\begin{figure*}[h]
    \centering 
    \includegraphics[height=\textheight]{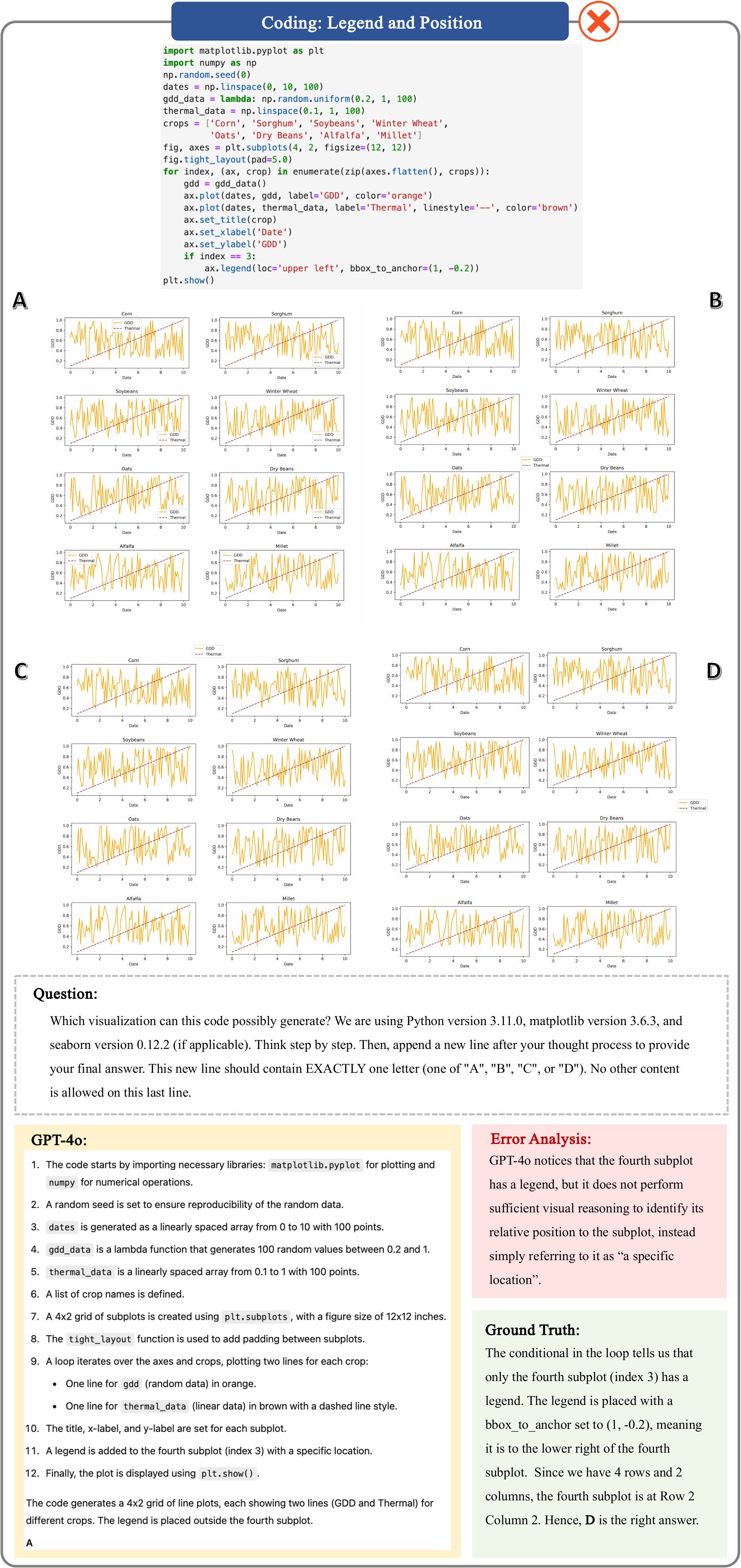}
    \caption{A sample error case in coding.}
    \label{fig:coding-case-wrong-2}
\end{figure*}

\begin{figure*}[h]
    \centering 
    \includegraphics[height=\textheight]{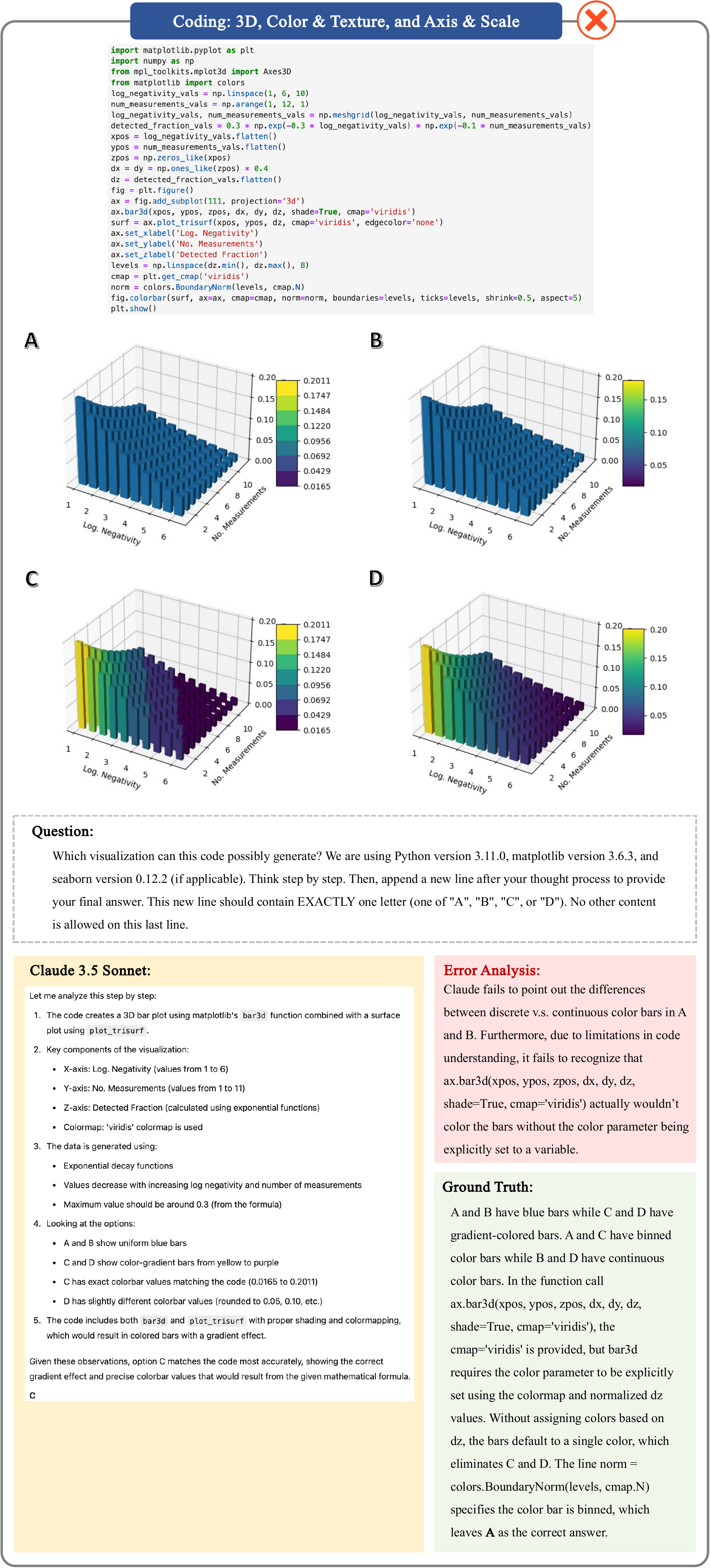}
    \caption{A sample error case in coding.}
    \label{fig:coding-case-wrong-3}
\end{figure*}

\begin{figure*}[h]
    \centering 
    \includegraphics[height=\textheight]{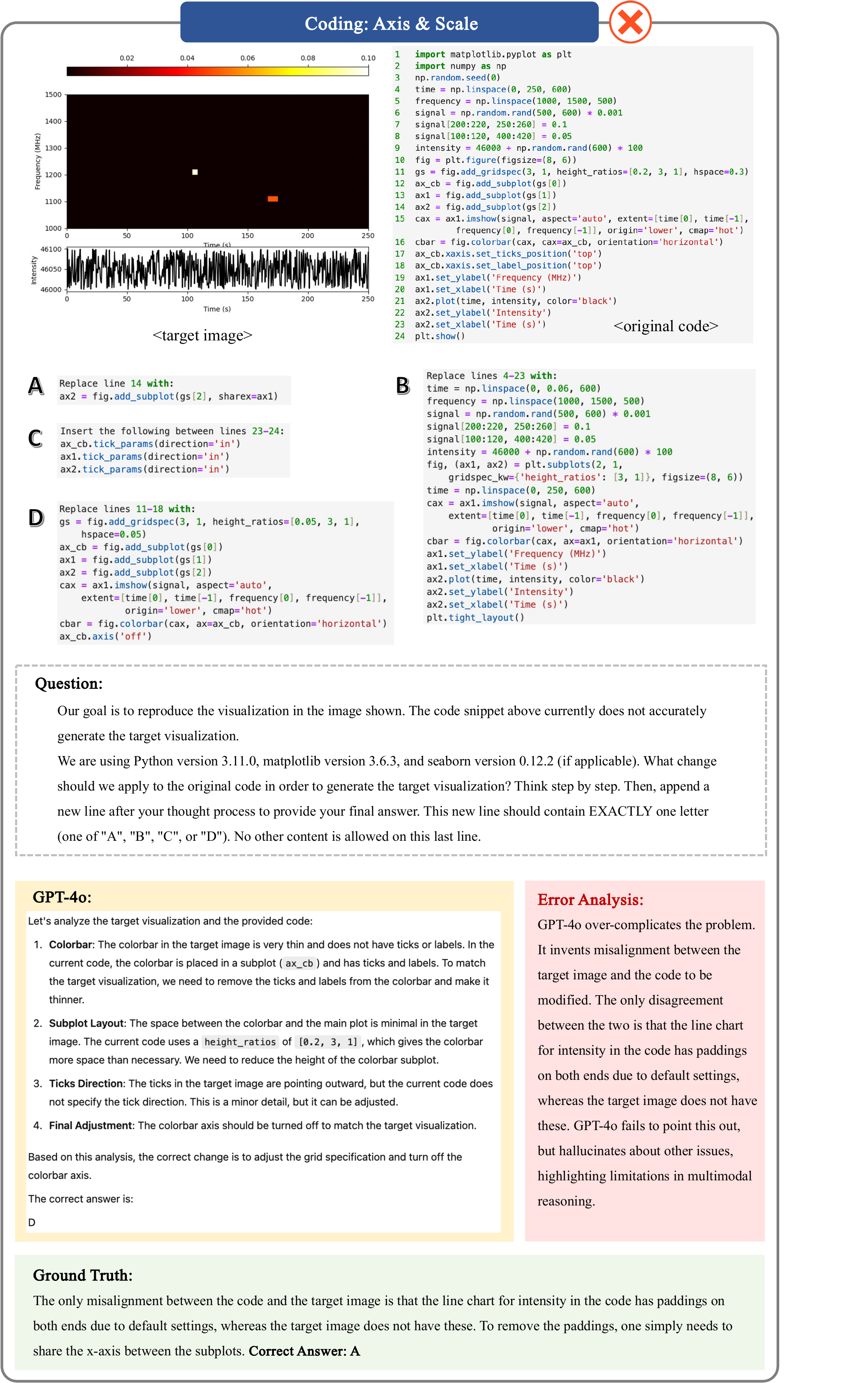}
    \caption{A sample error case in coding.}
    \label{fig:coding-case-wrong-4}
\end{figure*}

\begin{figure*}[h]
    \centering 
    \includegraphics[height=\textheight]{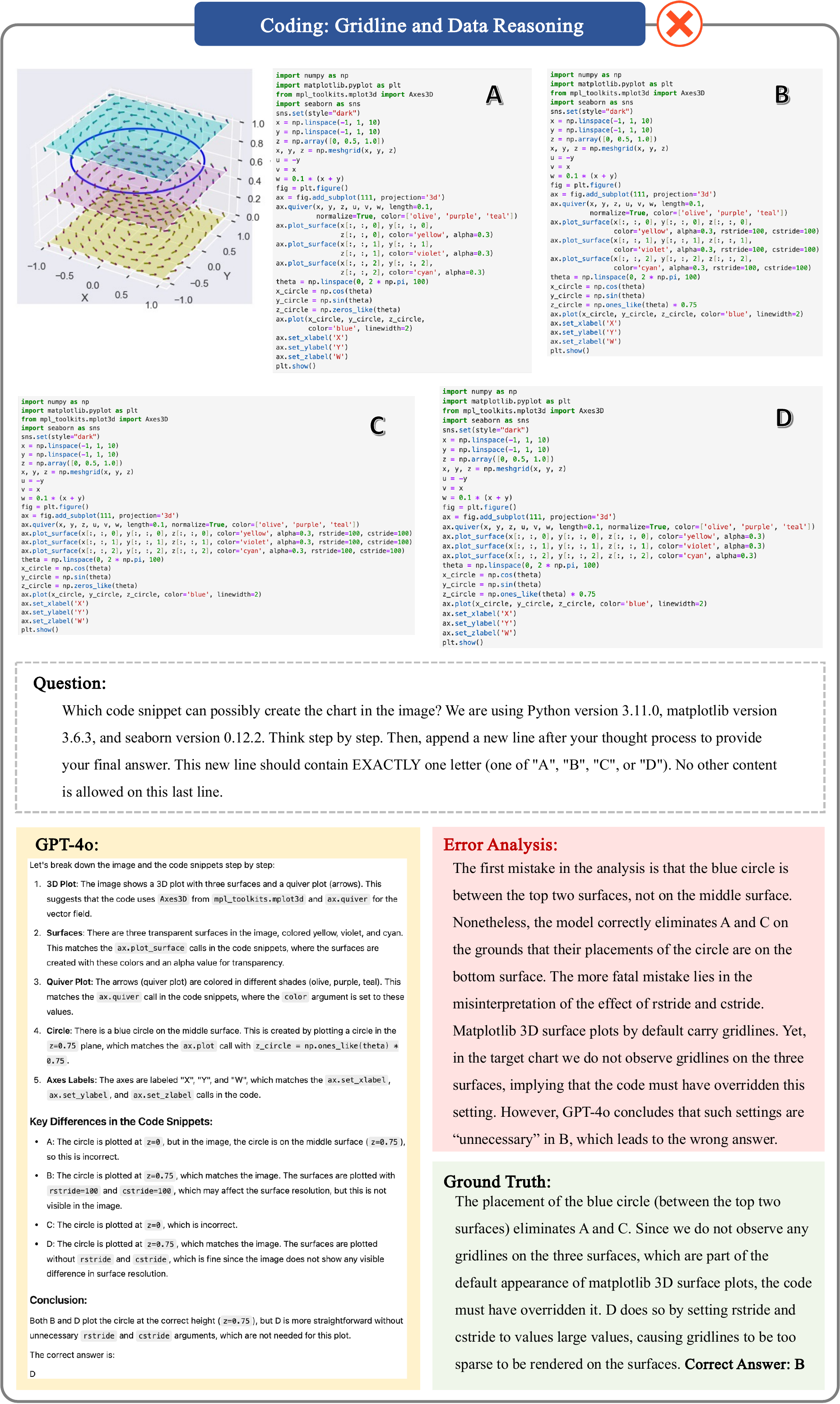}
    \caption{A sample error case in coding.}
    \label{fig:coding-case-wrong-5}
\end{figure*}

\end{document}